\documentclass[10pt,twocolumn,letterpaper]{article}

\usepackage{iccv}              
\usepackage{graphicx}
\graphicspath{{figures/}}
\usepackage[table,xcdraw]{xcolor}
\usepackage[normalem]{ulem}
\useunder{\uline}{\ul}{}
\usepackage{pifont}
\usepackage{multirow}
\usepackage[accsupp]{axessibility} 
\definecolor{hotpink}{RGB}{255,105,180}
\definecolor{darkgreen}{RGB}{0,150,0}  
\definecolor{darkred}{RGB}{150,0,0}   
\definecolor{darkblue}{RGB}{0,0,150} 

\definecolor{iccvblue}{rgb}{0.21,0.49,0.74}
\usepackage[pagebackref,breaklinks,colorlinks,allcolors=iccvblue]{hyperref}
\def\paperID{2159} 
\def\confName{ICCV}
\def\confYear{2025}

\title{From Easy to Hard: Progressive Active Learning Framework for Infrared Small Target Detection with Single Point Supervision}

\author{
\begin{minipage}{\textwidth}
\centering
\fontsize{10pt}{12pt}\selectfont
Chuang Yu$^{1,2,3}$, \, Jinmiao Zhao$^{1,2,3}$, \, Yunpeng Liu$^{2}$\textsuperscript{\dag}, \, Sicheng Zhao$^4$, \, Yimian Dai$^5$, \, Xiangyu Yue$^{6,7}$\textsuperscript{\dag} \\
\vspace{5pt}
$^1$Key Laboratory of Opto-Electronic Information Processing, Chinese Academy of Sciences\\
$^2$Shenyang Institute of Automation, Chinese Academy of Sciences  \,\,\,
$^3$University of Chinese Academy of Sciences\\
$^4$Tsinghua University \,\,\,
$^5$Nankai University \,\,\,
$^6$MMLab, The Chinese University of Hong Kong \,\,\,
$^7$CPII under InnoHK
\end{minipage}
}

\begin{document}
\maketitle
\renewcommand{\thefootnote}{\fnsymbol{footnote}}
\footnotetext{\textsuperscript{\dag} Corresponding authors.}
\begin{abstract}
Recently, single-frame infrared small target (SIRST) detection with single point supervision has drawn wide-spread attention. However, the latest label evolution with single point supervision (LESPS) framework suffers from instability, excessive label evolution, and difficulty in exerting embedded network performance. Inspired by organisms gradually adapting to their environment and continuously accumulating knowledge, we construct an innovative \textbf{\underline{P}}rogressive \textbf{\underline{A}}ctive \textbf{\underline{L}}earning (\textbf{PAL}) framework, which drives the existing SIRST detection networks progressively and actively recognizes and learns harder samples. Specifically, to avoid the early low-performance model leading to the wrong selection of hard samples, we propose a model pre-start concept, which focuses on automatically selecting a portion of easy samples and helping the model have basic task-specific learning capabilities. Meanwhile, we propose a refined dual-update strategy, which can promote reasonable learning of harder samples and continuous refinement of pseudo-labels. In addition, to alleviate the risk of excessive label evolution, a decay factor is reasonably introduced, which helps to achieve a dynamic balance between the expansion and contraction of target annotations. Extensive experiments show that existing SIRST detection networks equipped with our PAL framework have achieved state-of-the-art (SOTA) results on multiple public datasets. Furthermore, our PAL framework can build an efficient and stable bridge between full supervision and single point supervision tasks. Our code is available at {\href{https://github.com/YuChuang1205/PAL}{\textcolor{hotpink}{https://github.com/YuChuang1205/PAL}}}
\end{abstract} 

\begin{figure}[t]
  \centering
  \captionsetup{skip=4pt}
   \includegraphics[width=\columnwidth]{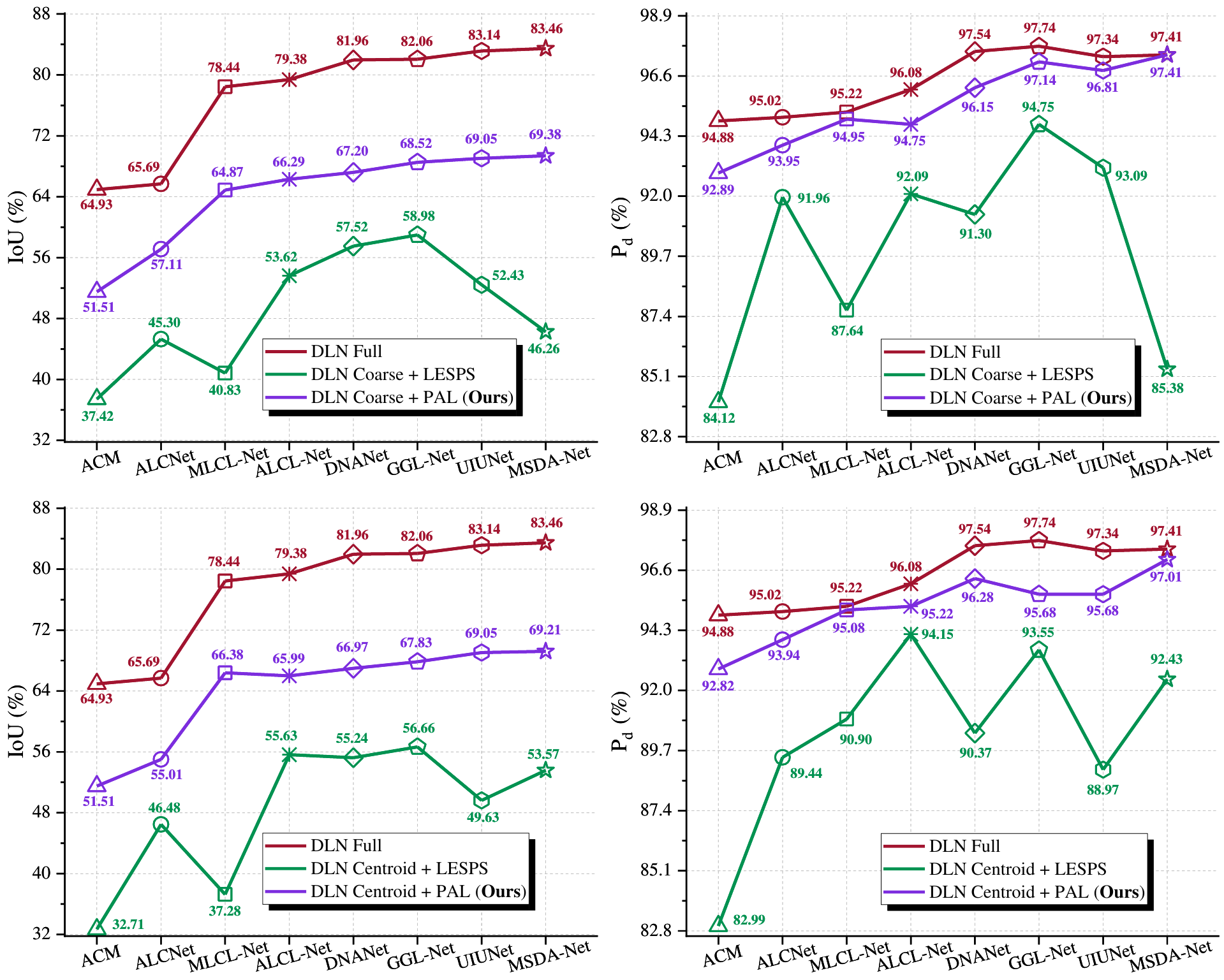}
   \caption{Comparison of different methods on the SIRST3 dataset. \textit{DLN Full}, \textit{DLN Coarse}, and \textit{DLN Centroid} denote DLN-based methods with full supervision, coarse and centroid point supervision. The curve trend of DLNs equipped with the PAL framework is \textbf{basically consistent} with that of full supervision, whereas DLNs with the LESPS framework~\cite{ying2023mapping} is not. In addition, compared with the LESPS, using our PAL framework can improve by \textbf{8.53\%-29.1\%} and \textbf{1.07\%-12.03\%} in the $IoU$ and $P_d$.}
   \label{fig:fig01}
   \vspace{-7pt}
\end{figure}

\vspace{-10pt}
\section{Introduction}
\label{sec:intro}
Single-frame infrared small target (SIRST) detection is one of the key technologies in infrared imaging systems~\cite{ying2023mapping,cui2025weakly,ni2025point}, which has been widely used in traffic analysis~\cite{zhang2022isnet,li2022dense}, environment observation~\cite{li2018robust,xu2023multiscale}, and maritime assistance~\cite{zhang2022exploring,zhao2024infrared}. However, this task faces many challenges, such as complex backgrounds, scarcity of intrinsic features and insufficient annotated data~\cite{ying2023mapping,zhang2022isnet,li2022dense,yu2022infrared,yu2022pay,zhao2023gradient}. Existing research mainly focuses on full supervision, but the cost of pixel-level dense annotation is expensive. Therefore, achieving high-performance SIRST detection with single point supervision is both challenging and significant.

Early SIRST detection focus on model-driven methods~\cite{arce1987theoretical,bai2010analysis,tomasi1998bilateral,qin2019infrared,chen2014novel,qi2012robust,chen2013local,wei2016multiscale,deng2016small,bai2018derivative,gao2013infrared,he2015small,zhang2018infrared,dai2017non}, which are prone to instability due to static background assumptions and hyperparameter sensitivity. Compared with model-driven methods, deep learning networks (DLNs) are data-driven and can automatically learn the nonlinear relationship between input images and labels~\cite{lecun2015deep}. Different from general targets, infrared small targets are small in size and lack of intrinsic features. It is difficult to achieve satisfactory detection results~\cite{ying2023mapping} by directly using existing general target detection methods such as the YOLO series~\cite{redmon2016you,redmon2017yolo9000,wang2025yolov10} and SSD~\cite{liu2016ssd}. Therefore, existing methods focus on building proprietary SIRST detection networks by leveraging domain knowledge~\cite{zhang2022isnet,yu2022infrared,yu2022pay,zhao2023gradient,dai2021attentional,zhao2025multi}. However, existing SIRST detection methods focus on full supervision, which requires expensive pixel-level dense annotations. To reduce the annotation cost, Ying \etal recently propose a LESPS framework~\cite{ying2023mapping} that combines the existing excellent SIRST detection network with continuous pseudo-label evolution to achieve SIRST detection with single point supervision. However, from the Sec. A of Supplementary Materials and \cref{fig:fig01}, the LESPS framework has problems such as instability, excessive label evolution, and difficulty in exerting embedded network performance.

To address these problems, we attempt to build a new, efficient and stable SIRST detection framework. Inspired by organisms gradually adapting to their environment and continuously accumulating knowledge, we came up with an idea: \textbf{\textit{The learning of network models should also be from easy to hard?}} \textit{We believe that an excellent learning process should be from easy to hard and take into account the learning ability of the current learner (model) rather than directly treating all tasks (samples) equally.} On this basis, we construct an innovative progressive active learning framework, which drives the existing SIRST detection networks progressively and actively recognizes and learns harder samples. Our PAL can be classified as automatic curriculum learning (CL)~\cite{wang2021survey}. However, unlike previous automatic CL that focuses on full supervision or other weakly supervised tasks~\cite{kumar2010self,kumar2011learning,wang2024efficienttrain++,zhou2023active,tang2012self}, this paper is the first attempt to explore its derivation to single point supervision tasks and construct an effective learning framework for SIRST detection with single point supervision. From \cref{fig:fig01}, our PAL achieves significant performance improvements and builds an efficient and stable bridge between full supervision and single point supervision tasks.

Specifically, \textbf{first}, to enable the network model to have basic task-specific learning capabilities in the initial training phase, we propose a model pre-start concept. Different from the existing methods that directly use all samples with point labels for training, it focuses on automatically selecting a portion of easy samples and put them into the training pool for initial training. For this task, an easy-sample pseudo-label generation (EPG) strategy that does not require manual annotation is constructed to achieve the initial selection of easy samples and the generation of their corresponding pseudo-labels. \textbf{Second}, considering that the model after model pre-start phase has the ability to recognize some hard samples and the pseudo-labels of samples in the training pool need to be refined, we construct a fine dual-update strategy that includes coarse outer updates (COU) and fine inner updates (FIN). Unlike the existing methods that perform only label evolution, the fine dual-update strategy not only dynamically selects relatively easy samples from hard samples into the training pool based on the capability of the current model, but also performs fine label updates on samples that have entered the training pool. \textbf{Finally}, we observe an interesting phenomenon: the existing framework has the problem that the evolution area does not shrink after over-expansion. To solve this problem, a decay factor is rationally introduced to achieve a dynamic balance between the expansion and contraction of target annotations. The contributions can be summarized as follows:

\begin{itemize}
    \item We construct an innovative PAL framework that can transfer existing SIRST detection networks to single point supervision tasks and drive them to progressively and actively learn harder samples, thereby achieving continuous performance improvements.
    \item A model pre-start concept is proposed, which focuses on automatically selecting a portion of easy samples and can help models have basic task-specific learning capabilities. For this task, we construct an EPG strategy to achieve the initial selection of easy samples and the generation of corresponding pseudo-labels.
    \item A fine dual-update strategy is proposed, which can promote reasonable learning of harder samples and continuous refinement of pseudo-labels.
    \item To alleviate the risk of excessive label evolution, a decay factor is reasonably introduced, which helps achieve a dynamic balance between the expansion and contraction of target annotations.
\end{itemize}

\begin{figure*}[htbp]
    \captionsetup{skip=5pt}
    \centering
    \includegraphics[width=\textwidth]{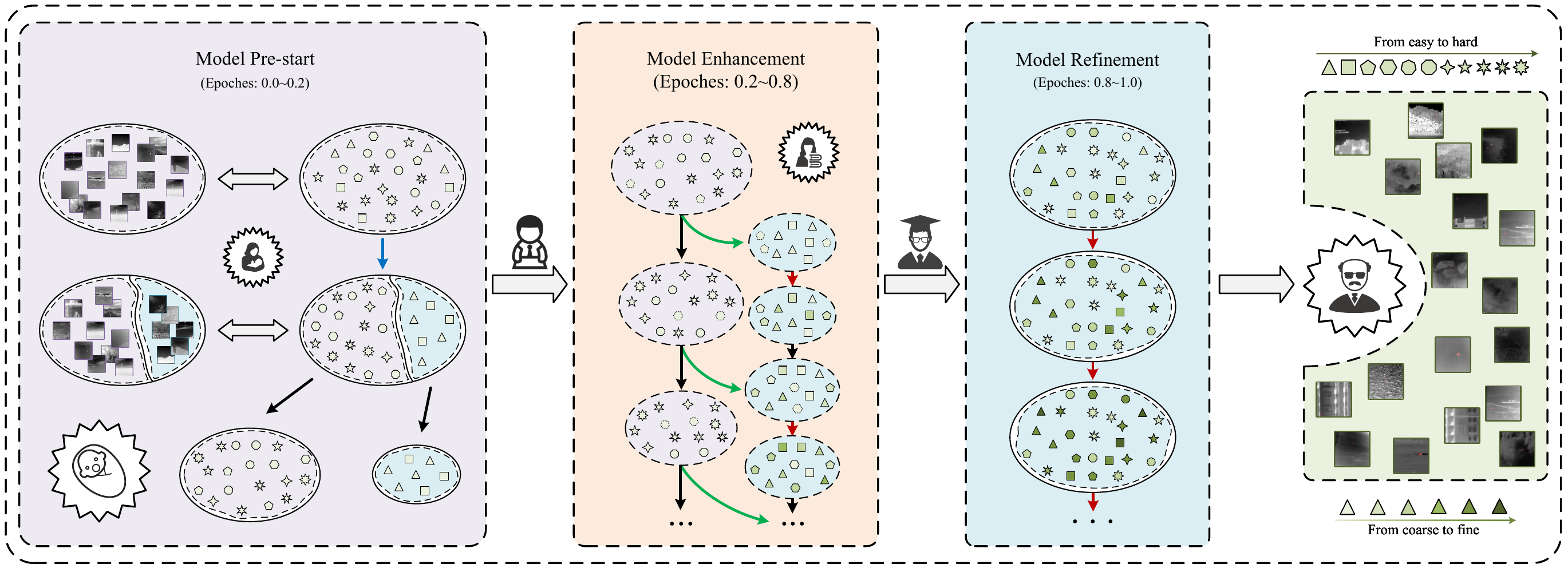}
    \caption{The PAL framework. The purple ellipse area is the preparation pool, which contains samples that cannot be trained temporarily. The blue ellipse area is the training pool, which contains samples that can be trained. Each shape refers to a different level of difficulty, and the color depth refers to the refinement of the corresponding pseudo-label. The \textcolor{darkblue}{\textit{blue}} arrow denotes the EPG strategy. The \textcolor{darkgreen}{\textit{green}} arrows and \textcolor{darkred}{\textit{red}} arrows denote the \textit{coarse outer updates} and \textit{fine inner updates}. 0.0, 0.2, 0.8 and 1.0 denote the division of the total training epochs.}
    \label{fig:fig02}
    \vspace{-6pt}
\end{figure*}

\section{Related Work}
\label{sec:related_work}
\textbf{SIRST detection} can be divided into model-driven non-deep learning methods~\cite{arce1987theoretical,bai2010analysis,tomasi1998bilateral,qin2019infrared,chen2014novel,qi2012robust,chen2013local,wei2016multiscale,deng2016small,bai2018derivative,gao2013infrared,he2015small,zhang2018infrared,dai2017non} and data-driven deep learning methods~\cite{zhang2022isnet,li2022dense,yu2022infrared,yu2022pay,zhao2023gradient,dai2021attentional,zhao2025multi,dai2021asymmetric,wu2022uiu}. Early research focus on model-driven methods, including background suppression-based methods~\cite{arce1987theoretical,bai2010analysis,tomasi1998bilateral,qin2019infrared,chen2014novel}, human visual system-based methods~\cite{qi2012robust,chen2013local,wei2016multiscale,deng2016small,bai2018derivative}, and image data structure-based methods~\cite{gao2013infrared,he2015small,zhang2018infrared,dai2017non}. However, these methods require careful hyperparameter tuning. Subsequently, data-driven methods gradually replaced model-driven methods. Its research focuses on incorporating domain knowledge~\cite{dai2021attentional,zhang2022isnet,yu2022infrared,yu2022pay,zhao2023gradient,zhao2025multi} and optimizing network architectures~\cite{dai2021asymmetric,li2022dense,wu2022uiu}. These methods perform well in SIRST detection with full supervision. However, pixel-wise dense annotations are expensive, which motivates us to exploit the more cost-effective point-level weak supervision~\cite{bearman2016s}.

\noindent\textbf{Point-based supervision} is of great significance and has been widely studied in object detection~\cite{papadopoulos2017training,zhang2022group,chen2021points}, object counting~\cite{laradji2018blobs,akiva2020finding,zand2022multiscale}, and semantic segmentation~\cite{bearman2016s,zhang2020interactive,cheng2022pointly,fan2022pointly,maninis2018deep,qian2019weakly,du2024pcl,zhang2021perturbed,wei2020multi,xu2022consistency}. We focus mainly on image segmentation in this paper. For weakly-supervised semantic segmentation with point labels, existing methods focus on general object segmentation~\cite{bearman2016s,zhang2020interactive,cheng2022pointly} and panoptic segmentation~\cite{fan2022pointly,maninis2018deep,qian2019weakly}, where targets exhibit rich colors, fine textures and often require multiple annotated points. Given the target characteristics and the single-point label of this task, the above method is difficult to apply directly. Currently, there is little research on this task, which can be mainly divided into static pseudo-label generation methods~\cite{li2023monte,kou2024mcgc,yuan2025beyond} and dynamic pseudo-label evolution methods~\cite{yang2024label,ying2023mapping}. The former often rely on additional prior or auxiliary information, while the latter (such as the LESPS framework) is more adaptable and has received wide-spread attention. However, existing dynamic methods suffer from instability, excessive label evolution, and difficulty in exerting embedded network performance. Therefore, we attempt to construct a new, efficient and stable framework.

\noindent\textbf{Curriculum Learning (CL)} is a training paradigm inspired by the structured learning progression in human curricula, where models are trained from easier to more difficult data~\cite{bengio2009curriculum}. It consists of two components:  a Difficulty Measurer and a Training Scheduler~\cite{wang2021survey}, and is categorized into predefined CL~\cite{tudor2016hard,el2020student,guo2018curriculumnet,ranjan2017curriculum,liu2018curriculum,tsvetkov2016learning,jimenez2019medical,penha2020curriculum,liu2020norm} and automatic CL~\cite{kumar2010self,kumar2011learning,wang2024efficienttrain++,zhou2023active,tang2012self,tang2012shifting,graves2017automated,zhou2018deep,zhao2015self,weinshall2018curriculum,hacohen2019power,matiisen2019teacher}. Predefined CL follows a manually designed training sequence, while automatic CL dynamically adjusts learning based on model feedback, offering greater flexibility~\cite{kumar2010self}. Unlike existing automatic CL methods that focus on full or other weakly supervision, this work is the first to explore its derivation to single point supervision, specifically for SIRST detection. Meanwhile, conventional automatic CL suffers from early-stage misjudgments in sample difficulty~\cite{wang2021survey}. The negative impact on our task will become more significant. Additionally, the single-point label format necessitates a rethinking of Difficulty Measurer and Training Scheduler design.  Therefore, we attempt to address these challenges and construct an innovative framework for infrared small target detection with single point supervision.

\section{Methods}
\label{sec:methods}
\subsection{Overview}
\label{sec:overview}
\begin{figure}[t]
  \captionsetup{skip=7pt}
  \centering
   \includegraphics[width=\columnwidth]{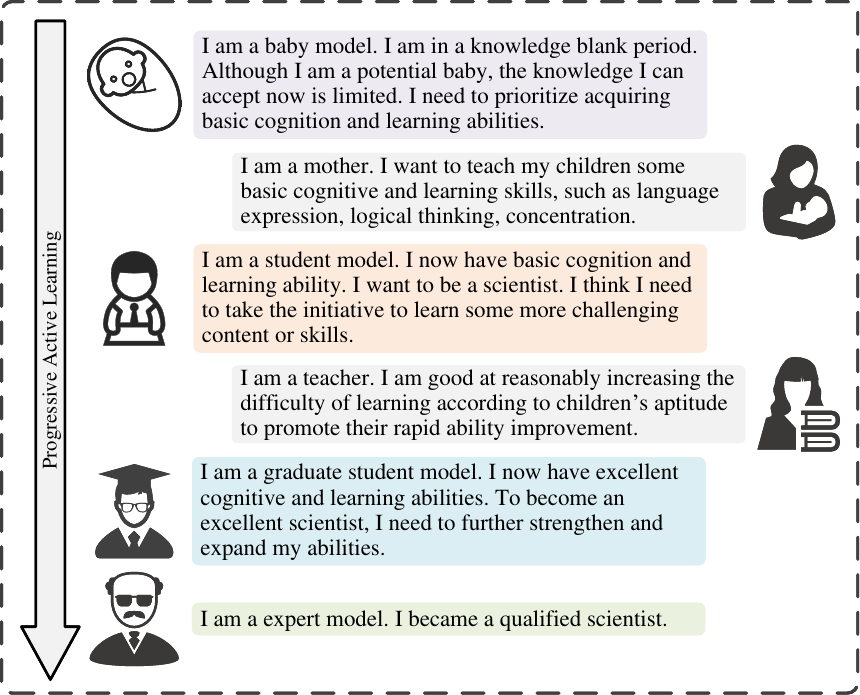}
   \caption{Schematic representation of progressive active learning.}
   \label{fig:fig03}
   \vspace{-6pt}
\end{figure}

For fine-grained SIRST detection with single point supervision, we propose an end-to-end PAL framework adaptable to all existing SIRST detection networks. From \cref{fig:fig02} and \cref{fig:fig03}, the training process consists of three phases: model pre-start, model enhancement, and model refinement.

The model pre-start phase aims to give model the initial task-specific learning ability. Like a newborn, an untrained model has potential but requires simple knowledge first. This is why existing methods that directly use all samples with point labels for training have problems such as unstable training and poor accuracy. Additionally, low-performance models may misidentify hard samples, thereby compromising final performance. Inspired by early maternal guidance, we construct an EPG strategy to automatically generate pseudo-labels for all samples, and divide them into easy samples and hard samples according to the quality of pseudo-labels. Only easy samples enter the training pool for training with a fixed epochs (0.2 × total epochs). For details, please see \cref{sec:model pre-start}. 

The model enhancement phase aims to make the model progressively and actively recognize and learn harder samples to ensure the accumulation of new knowledge, and to update the pseudo-labels of samples in the training pool. Specifically, The model evaluates hard samples in the preparation pool, transferring those with reliable pseudo-labels to the training pool through periodic coarse outer updates.  Simultaneously, all pseudo-labels in the training pool are iteratively refined via fine inner updates. It is like students adjusting their misperceptions of acquired information as their abilities grow. For details, please see \cref{sec:fine dual-update strategy}. In addition, inspired by teacher-guided learning, we introduce additional guidance to dynamically adjust the difficulty threshold based on the model's capabilities, accelerating learning and ensuring that the to-be-learned content is learned in a timely manner.

The model refinement phase aims to allow the model to fully learn hard samples and further refine pseudo-labels. Specifically, the hard samples are periodically and gradually added to the training pool in the model enhancement phase. For hard samples that enter later, it is difficult to be fully learned in a few periods. At the same time, as the model continues to learn hard samples, its stronger detection ability can help further refine the pseudo-labels. Therefore, we perform the fine inner updates periodically. This phase resembles graduates refining their expertise through repeated review and practice to achieve proficiency.

\subsection{Model pre-start}
\label{sec:model pre-start}
\begin{figure}[t]
  \captionsetup{skip=5pt}
  \centering
   \includegraphics[width=\columnwidth]{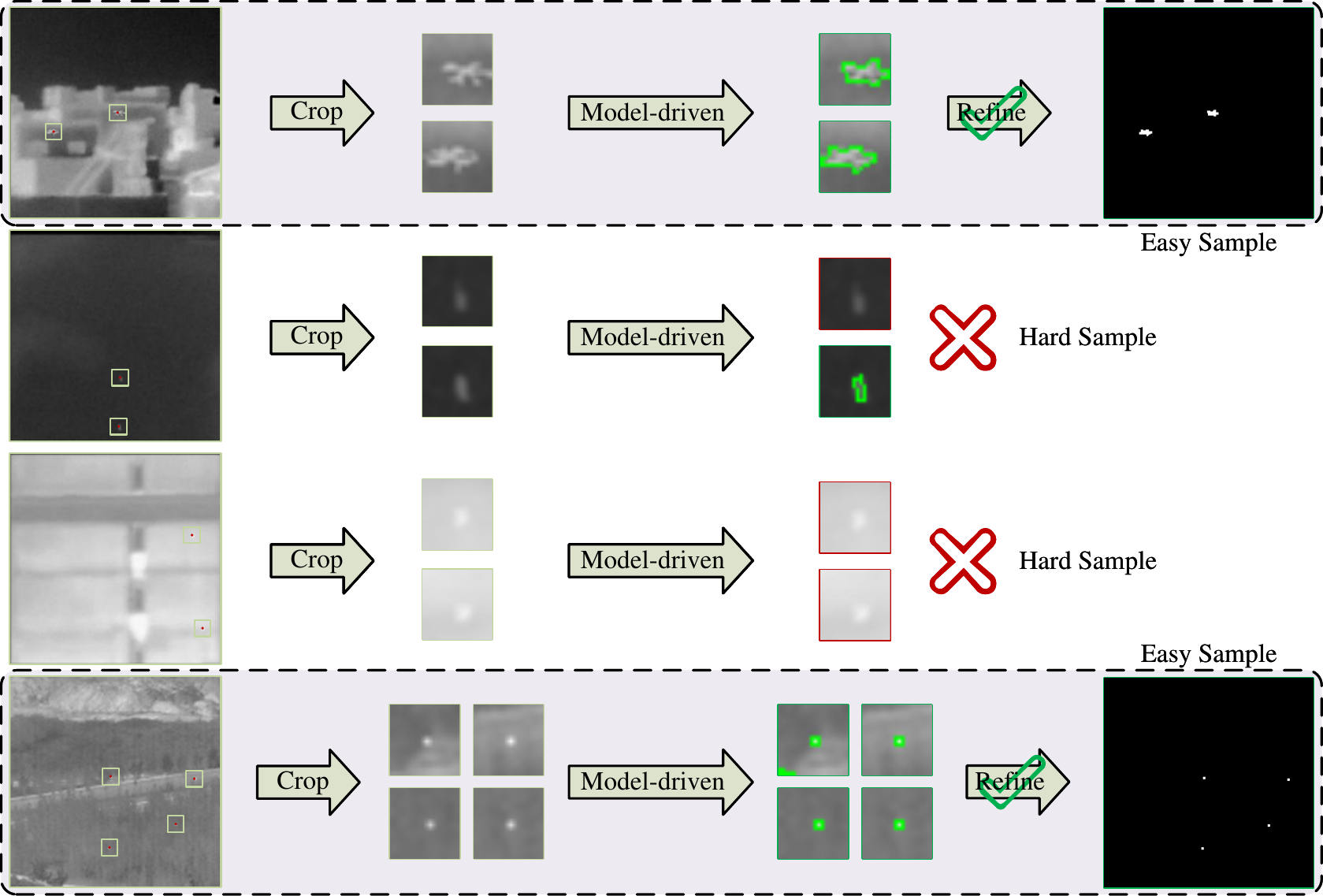}
   \vspace{-13pt}
   \caption{The EPG strategy. The \textcolor{darkgreen}{\textit{green}} boxes denote correct detections. The \textcolor{darkred}{\textit{red}} boxes denote error detections.}
   \label{fig:fig04}
   \vspace{-10pt}
\end{figure}

Compared to directly training with point labels, selecting easy samples that can generate high-quality pseudo-labels for initial training can help the network establish basic SIRST detection capabilities. Therefore, different from existing methods~\cite{ying2023mapping,zhao2024refined}, we propose a model pre-start concept. Considering the detection scene, imaging characteristics and target properties, infrared small targets are characterized by small size, high brightness and a lack of strong semantic contextual relationships with the surrounding environment~\cite{li2022dense,yu2022pay,dai2021attentional,zhao2025multi}. Some target areas can be detected by model-driven methods. However, these methods often lack robustness and are prone to false detections~\cite{zhao2025multi}. Therefore, we propose an EPG strategy to achieve the selection of easy samples, as shown in \cref{fig:fig04}.

To reduce background interference, we process local image patches centered on the given point label instead of the entire image. To accurately detect the target area, we first use the Gaussian filtering to appropriately suppress the noise. Second, the Canny operator~\cite{canny1986computational} is used to extract the target’s contour. Finally, a morphological closing operation mitigates contour fragmentation and fills the enclosed area to obtain the segmentation result.

To select easy samples and refine pseudo-labels, we make full use of the point labels and the segmentation results. \textbf{First}, the given point label is used to assess connected areas in the segmentation result. Areas containing the label point and below a set threshold are identified as true targets, while others are removed as false detections. \textbf{Second}, we calculate the target-level recall rate, which is the proportion of real targets that are correctly detected. If it is greater than and equal to the set threshold (0.8), it is regarded as an easy sample. Otherwise, it is a hard sample. \textbf{Third}, each segmented image patch of the easy sample is added to a pure black background with the original image size to generate a pseudo-label. \textbf{Finally}, point labels are added to compensate for missed targets, thereby refining the pseudo-labels.


The proposed EPG strategy selects easy samples with high-quality pseudo-labels for initial training, enabling the model to develop basic task-specific capabilities. This also prevents hard samples from introducing excessive noise or misleading information during the early training phase.

\subsection{Fine dual-update strategy}
\label{sec:fine dual-update strategy}
The quality of pseudo-labels directly impacts the final model's performance. Unlike existing frameworks that directly iteratively evolve all point labels~\cite{ying2023mapping,zhao2024refined}, we propose a fine dual-update strategy, consisting of a \textit{Coarse outer updates} and a \textit{Fine inner updates}.

\textbf{\textit{1) Coarse outer updates (COU):}} After the model pre-start phase, the model has basic SIRST detection capabilities. Based on it, we assess the remaining hard samples in the preparation pool to select ``easy samples'' whose generated pseudo-labels meet the selection rules into the training pool. The COU can be divided into three steps: determining recognizable samples, eliminating false targets, and supplementing missed targets.

First, we need to assess hard samples in the preparation pool that the current model can recognize:
\setlength{\abovedisplayskip}{5pt}
\setlength{\belowdisplayskip}{5pt}
\begin{gather}
S = \left\{ 
\begin{aligned}
& I \in Easy\,Sample \,\, \text{ if } R_m \leq T_m \,\&\, R_f \leq T_f \\
& I \in Hard\,Sample \,\, \text{ otherwise}
\end{aligned} 
\right.
\end{gather}
where $R_m$ and $R_f$ denote the target-level missed detection rate (the ratio of missed real targets) and false detection rate (the ratio of incorrectly detected targets to real targets). $T_m$ and $T_f$ denote the corresponding thresholds. When there is an intersection between the true point label and the predicted target area, the target is detected correctly.

Second, the selected easy samples undergo false target elimination. Specifically, predicted target areas that do not intersect with any true point label are directly removed. The formula expression is as follows:
\begin{gather}
{\hat P_b} = {P_b}\backslash \{ {A_f} \in {P_b}|Instersection({A_f},{L_{true}}) = \phi \} 
\end{gather}
where $P_b$ and ${\hat P_b}$ denote the detection results before and after false target elimination. $A_f$ denotes the predicted target area. $\backslash$ denotes the difference of sets.

Finally, the ground-truth point labels are incorporated to mitigate missed targets and ensure effective learning.

Considering that the model will have stronger cognitive and learning abilities after learning hard samples, we periodically perform COU during the model enhancement phase. This drives the model to progressively and actively recognize and learn more harder samples without catastrophically forgetting easy samples.

\begin{table*}[]
\caption{$IoU$ (\%), $nIoU$ (\%), $P_d$ (\%) and $F_a$ ($\times 10^{-6}$) values of different methods achieved on the SIRST3 dataset. \textit{NUAA-SIRST-Test}, \textit{NUDT-SIRST-Test} and \textit{IRSTD-1K-Test} denote the decompositions of \textit{SIRST3-Test} to verify the robustness of the model.}
\vspace{-5pt}
\label{tab:tab01}
\setlength{\tabcolsep}{2mm}
\resizebox{\textwidth}{!}{
\begin{tabular}{c|c|cccc|cccc|cccc|cccc}
\hline
                                    &                                                 & \multicolumn{4}{c|}{SIRST3-Test}                                                                                                                                                                                                                              & \multicolumn{4}{c|}{NUAA-SIRST-Test}                                                                                                                                                                                                                          & \multicolumn{4}{c|}{NUDT-SIRST-Test}                                                                                                                                                                                                                          & \multicolumn{4}{c}{IRSTD-1K-Test}                                                                                                                                                                                                                             \\
\hline
\end{tabular}
}
\vspace{-7pt}
\end{table*}

\textbf{\textit{2) Fine inner updates (FIU):}} Whether it is the easy samples that enter the training pool during the model pre-start phase or the hard samples that continuously enter the training pool during the model enhancement phase, the pseudo-labels they initially generate need to be refined. To solve this problem, we construct a FIU. It is inspired by~\cite{ying2023mapping} and effectively solves the potential problem of excessive label evolution as shown in the Supplementary Materials. The FIU is divided into three steps: candidate area extraction, false area elimination and pseudo-label update.

For candidate area extraction, we first obtain the binary label for the $n_{th}$ pseudo-label iteration. Second, the centroid of each target connected area in the pseudo-label is extracted and the corresponding $d \times d$ local area in the pseudo-label and prediction result is cropped. Finally, to highlight the target area and reduce accumulated errors, we use adaptive thresholding to extract the local candidate areas from the prediction results:
\begin{gather}
C_n^i = (P_n^i > {T_{adapt}}) \\
{T_{adapt}} = max(P_n^i)({T_b} + k(1 - {T_b})L_n^i/(hwr))
\end{gather}
where $C_n^i$ denotes candidate areas. $P_n^i$ and $L_n^i$ denote the prediction result and pseudo-label of the image patch. $h$ and $w$ denote the height and width of the image. $T_b$ and $k$ are both set to 0.5~\cite{ying2023mapping}. $r$ is set to 0.15\%~\cite{li2022dense,dai2021asymmetric}.

For false area elimination, similar to Eq.(2), we use the centroid of the target area in the pseudo-label to check for intersections with the candidate areas. Candidate areas without intersections are eliminated.

For pseudo-label update, we first restore the positions of all refined candidate areas and use the maximum method to combine them into complete candidate areas. Then, the pseudo-label is updated:
\begin{gather}
{N_n} = \max (\{ {N_n},\hat C_n^i\} ) \\
{L_{n + 1}} = \lambda {L_n} \odot (1 - {N_n}) + \frac{{{L_n} + {P_n}}}{2} \odot {N_n}
\end{gather}
where $\lambda$ is the decay factor. $N_n$ is initialized to a zero matrix with the same size as the label. $\hat C_n^i$ denotes the refined candidate areas of the image patch. $P_n$ and $L_n$ denote the prediction result and pseudo-label.

With the improvement of model capabilities and various constraints, the evolution of the target annotation area achieves a dynamic balance between expansion and contraction and generates more refined pseudo-labels.

\subsection{Loss Function}
\label{sec:loss function}
For the SIRST detection task, the lack of intrinsic features makes it difficult to accurately locate the target area~\cite{yu2022pay,dai2021attentional,zhao2025multi}. Therefore, we introduce an edge-enhanced difficulty-mining (EEDM) loss~\cite{zhao2025towards} to guide network optimization. This helps alleviate the imbalance problem of positive and negative pixels, and directs the model’s focus to edges and difficult pixels during training. Specifically, EEDM loss emphasizes pixel-level fine-grained detection, which consists of two parts: edge pixel enhancement and difficult pixel mining. On the one hand, the edge pixel enhancement increases the focus on edge details by weighting the loss of edge pixels, thereby enhancing the model's learning of important features. On the other hand, the difficult pixel mining helps the model better focus on difficult-to-detect target areas by filtering pixels with larger loss, thereby promoting the model's learning in complex areas and preventing small targets from being submerged by the background. More details can be found in the Supplementary Materials.

\section{Experiments}
\label{sec:experiments}
\subsection{Datasets}
\label{sec:datasets}
Four public SIRST detection datasets are used in the experiments, namely SIRST3~\cite{ying2023mapping}, NUAA-SIRST~\cite{dai2021asymmetric}, NUDT-SIRST~\cite{li2022dense}, and IRSTD-1K~\cite{zhang2022isnet}. They contain 2755, 427, 1327 and 1001 samples, respectively. We follow the partitioning in~\cite{ying2023mapping,li2022dense} for NUAA-SIRST and NUDT-SIRST and follow the partitioning in~\cite{ying2023mapping,zhang2022isnet} for IRSTD-1K. The stability under limited samples can be evaluated on these three datasets. SIRST3 is a multi-sample, multi-scene, and multi-target dataset that combines NUAA-SIRST, NUDT-SIRST, and IRSTD-1K. It can more effectively evaluate the comprehensive detection capabilities. Consistent with~\cite{ying2023mapping}, we process the labels of the training set into two types: \textit{Coarse} point labels and \textit{Centroid} point labels.

\subsection{Implementation Details}
\label{sec:implementation details}
\hspace*{1em} \textbf{\textit{Experimental Settings:}} We use the AdamW~\cite{loshchilov2017decoupled} optimizer with initial learning rate of $1e^{\text{\scriptsize -3}}$. The batch size is set to 16. The total epochs is 400, which is divided into 3 parts: \textit{0-0.2}, \textit{0-0.8} and \textit{0.8-1}. The cropped image patch in EPG is 20×20 pixels. We interestingly found that DLN-based models exhibit minimal false detections but significant missed detections for this task. Since our strategy eliminates false positive areas, we set $T_f$ to a larger value to reduce its constraint. To ensure all samples enter the training pool during model enhancement phase, the initial threshold of $T_m$ is set to 0.2 and increases linearly to 1 as the epoch increases. The epoch period of COU and FIU is set to 5. During training, all images are normalized and randomly cropped into 256×256 pixel patches as network inputs.

\textbf{\textit{Evaluation Metrics:}} We used two pixel-level evaluation metrics of intersection over union (IoU) and normalized IoU (nIoU)~\cite{dai2021asymmetric} and two target-level evaluation metrics of detection probability ($P_d$) and false alarm rate ($F_a$)~\cite{li2022dense}. In addition, since the magnitude of $F_a$ is $10^{\text{\scriptsize -6}}$, its value is small and prone to fluctuations. Therefore, we also evaluate the model using a threshold-based approach, and the model is judged invalid when $F_a > 1e^{\text{\scriptsize -4}}$~\cite{zhao2024refined,yu2024lr}. Invalid results are highlighted in red.

\begin{table*}[]
\caption{$IoU$ (\%), $nIoU$ (\%), $P_d$ (\%) and $F_a$ ($\times 10^{-6}$) values of different methods achieved on the separate NUAA-SIRST, NUDT-SIRST, and IRSTD-1k datasets. \textit{(213:214)}, \textit{(663:664)} and \textit{(800:201)} denote the division of training samples and test samples.}
\vspace{-5pt}
\label{tab:tab02}
\setlength{\tabcolsep}{3mm}
\resizebox{\textwidth}{!}{
\begin{tabular}{c|c|cccc|cccc|cccc}
\hline
                                    &                                                 & \multicolumn{4}{c|}{NUAA-SIRST (213:214)}                                                                                                                                                                                                                     & \multicolumn{4}{c|}{NUDT-SIRST (663:664)}                                                                                                                                                                                                                     & \multicolumn{4}{c}{IRSTD-1K (800:201)}                                                                                                                                                                                                                        \\ \cline{3-14} 
\multirow{-2}{*}{Scheme}            & \multirow{-2}{*}{Description}                   & $IoU$                                                           & $nIoU$                                                          & $P_d$                                                            & $F_a$                                                            & $IoU$                                                           & $nIoU$                                                          & $P_d$                                                            & $F_a$                                                            & $IoU$                                                           & $nIoU$                                                          & $P_d$                                                            & $F_a$                                                           \\ \hline
ACM\textsuperscript{~\cite{dai2021asymmetric}}                        & DLN Full                                        & 65.67                                                         & 63.74                                                         & 90.11                                                         & 24.01                                                         & 65.33                                                         & 65.12                                                         & 95.87                                                         & 12.92                                                         & 60.45                                                         & 53.70                                                         & 92.26                                                         & 46.06                                                         \\ \hline
ALCNet\textsuperscript{~\cite{dai2021attentional}}                     & DLN Full                                        & 66.41                                                         & 65.18                                                         & 91.63                                                         & 35.26                                                         & 69.74                                                         & 70.67                                                         & 97.46                                                         & 11.15                                                         & 62.47                                                         & 55.25                                                         & 88.89                                                         & 36.79                                                         \\ \hline
                                    & DLN Full                                        & 74.68                                                         & 76.50                                                         & 95.82                                                         & 28.74                                                         & 94.03                                                         & 93.97                                                         & 98.73                                                         & 7.721                                                         & 64.86                                                         & 63.35                                                         & 91.25                                                         & 23.23                                                         \\
                                    & \cellcolor[HTML]{CFCECE}DLN Coarse + LESPS      & \cellcolor[HTML]{CFCECE}41.31                                 & \cellcolor[HTML]{CFCECE}41.13                                 & \cellcolor[HTML]{CFCECE}90.87                                 & \cellcolor[HTML]{CFCECE}43.22                                 & \cellcolor[HTML]{CFCECE}53.55                                 & \cellcolor[HTML]{CFCECE}48.70                                 & \cellcolor[HTML]{CFCECE}90.05                                 & \cellcolor[HTML]{CFCECE}21.92                                 & \cellcolor[HTML]{CFCECE}44.64                                 & \cellcolor[HTML]{CFCECE}42.04                                 & \cellcolor[HTML]{CFCECE}88.22                                 & \cellcolor[HTML]{CFCECE}36.42                                 \\
\multirow{-3}{*}{MLCL-Net\textsuperscript{~\cite{yu2022infrared}}}  & \cellcolor[HTML]{CFCECE}DLN Coarse + PAL (Ours) & \cellcolor[HTML]{CFCECE}{\color[HTML]{066F29} \textbf{68.76}} & \cellcolor[HTML]{CFCECE}{\color[HTML]{066F29} \textbf{70.75}} & \cellcolor[HTML]{CFCECE}{\color[HTML]{066F29} \textbf{93.92}} & \cellcolor[HTML]{CFCECE}{\color[HTML]{066F29} \textbf{39.10}} & \cellcolor[HTML]{CFCECE}{\color[HTML]{066F29} \textbf{74.00}} & \cellcolor[HTML]{CFCECE}{\color[HTML]{066F29} \textbf{73.71}} & \cellcolor[HTML]{CFCECE}{\color[HTML]{066F29} \textbf{95.45}} & \cellcolor[HTML]{CFCECE}{\color[HTML]{066F29} \textbf{9.40}}  & \cellcolor[HTML]{CFCECE}{\color[HTML]{066F29} \textbf{58.90}} & \cellcolor[HTML]{CFCECE}{\color[HTML]{066F29} \textbf{58.79}} & \cellcolor[HTML]{CFCECE}{\color[HTML]{066F29} \textbf{90.91}} & \cellcolor[HTML]{CFCECE}{\color[HTML]{066F29} \textbf{24.63}} \\ \hline
                                    & DLN Full                                        & 72.84                                                         & 72.59                                                         & 95.06                                                         & 17.01                                                         & 92.80                                                         & 93.01                                                         & 99.05                                                         & 2.21                                                          & 65.56                                                         & 65.03                                                         & 91.58                                                         & 9.68                                                          \\
                                    & \cellcolor[HTML]{CFCECE}DLN Coarse + LESPS      & \cellcolor[HTML]{CFCECE}37.99                                 & \cellcolor[HTML]{CFCECE}41.91                                 & \cellcolor[HTML]{CFCECE}90.87                                 & \cellcolor[HTML]{CFCECE}35.54                                 & \cellcolor[HTML]{CFCECE}45.89                                 & \cellcolor[HTML]{CFCECE}43.66                                 & \cellcolor[HTML]{CFCECE}90.48                                 & \cellcolor[HTML]{CFCECE}34.45                                 & \cellcolor[HTML]{CFCECE}49.46                                 & \cellcolor[HTML]{CFCECE}45.00                                 & \cellcolor[HTML]{CFCECE}82.49                                 & \cellcolor[HTML]{CFCECE}26.72                                 \\
\multirow{-3}{*}{ALCL-Net\textsuperscript{~\cite{yu2022pay}}}  & \cellcolor[HTML]{CFCECE}DLN Coarse + PAL (Ours) & \cellcolor[HTML]{CFCECE}{\color[HTML]{066F29} \textbf{67.49}} & \cellcolor[HTML]{CFCECE}{\color[HTML]{066F29} \textbf{67.67}} & \cellcolor[HTML]{CFCECE}{\color[HTML]{066F29} \textbf{93.54}} & \cellcolor[HTML]{CFCECE}{\color[HTML]{066F29} \textbf{24.01}} & \cellcolor[HTML]{CFCECE}{\color[HTML]{066F29} \textbf{70.62}} & \cellcolor[HTML]{CFCECE}{\color[HTML]{066F29} \textbf{72.56}} & \cellcolor[HTML]{CFCECE}{\color[HTML]{066F29} \textbf{98.31}} & \cellcolor[HTML]{CFCECE}{\color[HTML]{066F29} \textbf{7.93}}  & \cellcolor[HTML]{CFCECE}{\color[HTML]{066F29} \textbf{57.98}} & \cellcolor[HTML]{CFCECE}{\color[HTML]{066F29} \textbf{53.36}} & \cellcolor[HTML]{CFCECE}{\color[HTML]{066F29} \textbf{88.22}} & \cellcolor[HTML]{CFCECE}{\color[HTML]{066F29} \textbf{20.61}} \\ \hline
                                    & DLN Full                                        & 76.40                                                         & 78.32                                                         & 96.20                                                         & 20.72                                                         & 95.17                                                         & 95.19                                                         & 98.94                                                         & 2.00                                                          & 69.06                                                         & 65.22                                                         & 91.58                                                         & 11.56                                                         \\
                                    & \cellcolor[HTML]{CFCECE}DLN Coarse + LESPS      & \cellcolor[HTML]{CFCECE}9.29                                  & \cellcolor[HTML]{CFCECE}7.36                                  & \cellcolor[HTML]{CFCECE}19.77                                 & \cellcolor[HTML]{CFCECE}{\color[HTML]{C00000} 415.72}         & \cellcolor[HTML]{CFCECE}48.10                                 & \cellcolor[HTML]{CFCECE}48.84                                 & \cellcolor[HTML]{CFCECE}81.80                                 & \cellcolor[HTML]{CFCECE}85.19                                 & \cellcolor[HTML]{CFCECE}50.99                                 & \cellcolor[HTML]{CFCECE}47.61                                 & \cellcolor[HTML]{CFCECE}85.52                                 & \cellcolor[HTML]{CFCECE}25.49                                 \\
\multirow{-3}{*}{DNANet\textsuperscript{~\cite{li2022dense}}}    & \cellcolor[HTML]{CFCECE}DLN Coarse + PAL (Ours) & \cellcolor[HTML]{CFCECE}{\color[HTML]{066F29} \textbf{66.57}} & \cellcolor[HTML]{CFCECE}{\color[HTML]{066F29} \textbf{69.73}} & \cellcolor[HTML]{CFCECE}{\color[HTML]{066F29} \textbf{90.49}} & \cellcolor[HTML]{CFCECE}{\color[HTML]{066F29} \textbf{27.44}} & \cellcolor[HTML]{CFCECE}{\color[HTML]{066F29} \textbf{73.29}} & \cellcolor[HTML]{CFCECE}{\color[HTML]{066F29} \textbf{74.70}} & \cellcolor[HTML]{CFCECE}{\color[HTML]{066F29} \textbf{98.10}} & \cellcolor[HTML]{CFCECE}{\color[HTML]{066F29} \textbf{22.08}} & \cellcolor[HTML]{CFCECE}{\color[HTML]{066F29} \textbf{58.40}} & \cellcolor[HTML]{CFCECE}{\color[HTML]{066F29} \textbf{60.29}} & \cellcolor[HTML]{CFCECE}{\color[HTML]{066F29} \textbf{91.25}} & \cellcolor[HTML]{CFCECE}{\color[HTML]{066F29} \textbf{24.52}} \\ \hline
                                    & DLN Full                                        & 75.47                                                         & 75.93                                                         & 97.34                                                         & 20.65                                                         & 94.86                                                         & 94.93                                                         & 99.47                                                         & 1.03                                                          & 69.09                                                         & 65.52                                                         & 92.59                                                         & 13.68                                                         \\
                                    & \cellcolor[HTML]{CFCECE}DLN Coarse + LESPS      & \cellcolor[HTML]{CFCECE}55.17                                 & \cellcolor[HTML]{CFCECE}54.51                                 & \cellcolor[HTML]{CFCECE}89.35                                 & \cellcolor[HTML]{CFCECE}26.27                                 & \cellcolor[HTML]{CFCECE}57.39                                 & \cellcolor[HTML]{CFCECE}56.67                                 & \cellcolor[HTML]{CFCECE}91.75                                 & \cellcolor[HTML]{CFCECE}23.92                                 & \cellcolor[HTML]{CFCECE}50.90                                 & \cellcolor[HTML]{CFCECE}48.12                                 & \cellcolor[HTML]{CFCECE}86.20                                 & \cellcolor[HTML]{CFCECE}{\color[HTML]{066F29} \textbf{27.73}} \\
\multirow{-3}{*}{GGL-Net\textsuperscript{~\cite{zhao2023gradient}}}  & \cellcolor[HTML]{CFCECE}DLN Coarse + PAL (Ours) & \cellcolor[HTML]{CFCECE}{\color[HTML]{066F29} \textbf{63.46}} & \cellcolor[HTML]{CFCECE}{\color[HTML]{066F29} \textbf{64.31}} & \cellcolor[HTML]{CFCECE}{\color[HTML]{066F29} \textbf{90.11}} & \cellcolor[HTML]{CFCECE}{\color[HTML]{066F29} \textbf{25.73}} & \cellcolor[HTML]{CFCECE}{\color[HTML]{066F29} \textbf{73.10}} & \cellcolor[HTML]{CFCECE}{\color[HTML]{066F29} \textbf{74.83}} & \cellcolor[HTML]{CFCECE}{\color[HTML]{066F29} \textbf{98.20}} & \cellcolor[HTML]{CFCECE}{\color[HTML]{066F29} \textbf{13.12}} & \cellcolor[HTML]{CFCECE}{\color[HTML]{066F29} \textbf{60.72}} & \cellcolor[HTML]{CFCECE}{\color[HTML]{066F29} \textbf{57.05}} & \cellcolor[HTML]{CFCECE}{\color[HTML]{066F29} \textbf{90.24}} & \cellcolor[HTML]{CFCECE}40.20                                 \\ \hline
                                    & DLN Full                                        & 78.02                                                         & 76.85                                                         & 96.58                                                         & 14.68                                                         & 95.07                                                         & 95.10                                                         & 98.73                                                         & 0.21                                                          & 70.94                                                         & 64.32                                                         & 91.25                                                         & 10.57                                                         \\
                                    & \cellcolor[HTML]{CFCECE}DLN Coarse + LESPS      & \cellcolor[HTML]{CFCECE}7.11                                  & \cellcolor[HTML]{CFCECE}8.02                                  & \cellcolor[HTML]{CFCECE}72.62                                 & \cellcolor[HTML]{CFCECE}{\color[HTML]{C00000} 550.80}         & \cellcolor[HTML]{CFCECE}35.89                                 & \cellcolor[HTML]{CFCECE}33.12                                 & \cellcolor[HTML]{CFCECE}78.41                                 & \cellcolor[HTML]{CFCECE}89.05                                 & \cellcolor[HTML]{CFCECE}21.49                                 & \cellcolor[HTML]{CFCECE}18.29                                 & \cellcolor[HTML]{CFCECE}73.74                                 & \cellcolor[HTML]{CFCECE}{\color[HTML]{C00000} 227.06}         \\
\multirow{-3}{*}{UIUNet\textsuperscript{~\cite{wu2022uiu}}}   & \cellcolor[HTML]{CFCECE}DLN Coarse + PAL (Ours) & \cellcolor[HTML]{CFCECE}{\color[HTML]{066F29} \textbf{62.25}} & \cellcolor[HTML]{CFCECE}{\color[HTML]{066F29} \textbf{61.68}} & \cellcolor[HTML]{CFCECE}{\color[HTML]{066F29} \textbf{91.25}} & \cellcolor[HTML]{CFCECE}{\color[HTML]{066F29} \textbf{34.99}} & \cellcolor[HTML]{CFCECE}{\color[HTML]{066F29} \textbf{74.89}} & \cellcolor[HTML]{CFCECE}{\color[HTML]{066F29} \textbf{76.11}} & \cellcolor[HTML]{CFCECE}{\color[HTML]{066F29} \textbf{98.62}} & \cellcolor[HTML]{CFCECE}{\color[HTML]{066F29} \textbf{7.10}}  & \cellcolor[HTML]{CFCECE}{\color[HTML]{066F29} \textbf{61.70}} & \cellcolor[HTML]{CFCECE}{\color[HTML]{066F29} \textbf{59.99}} & \cellcolor[HTML]{CFCECE}{\color[HTML]{066F29} \textbf{92.57}} & \cellcolor[HTML]{CFCECE}{\color[HTML]{066F29} \textbf{21.60}} \\ \hline
                                    & DLN Full                                        & 76.73                                                         & 77.78                                                         & 96.20                                                         & 21.75                                                         & 95.27                                                         & 95.18                                                         & 99.15                                                         & 1.72                                                          & 70.98                                                         & 65.70                                                         & 93.94                                                         & 33.95                                                         \\
                                    & \cellcolor[HTML]{CFCECE}DLN Coarse + LESPS      & \cellcolor[HTML]{CFCECE}43.39                                 & \cellcolor[HTML]{CFCECE}49.33                                 & \cellcolor[HTML]{CFCECE}88.59                                 & \cellcolor[HTML]{CFCECE}27.37                                 & \cellcolor[HTML]{CFCECE}41.60                                 & \cellcolor[HTML]{CFCECE}38.25                                 & \cellcolor[HTML]{CFCECE}79.79                                 & \cellcolor[HTML]{CFCECE}34.42                                 & \cellcolor[HTML]{CFCECE}43.29                                 & \cellcolor[HTML]{CFCECE}38.16                                 & \cellcolor[HTML]{CFCECE}83.16                                 & \cellcolor[HTML]{CFCECE}54.98                                 \\
\multirow{-3}{*}{MSDA-Net\textsuperscript{~\cite{zhao2025multi}}} & \cellcolor[HTML]{CFCECE}DLN Coarse + PAL (Ours) & \cellcolor[HTML]{CFCECE}{\color[HTML]{066F29} \textbf{66.78}} & \cellcolor[HTML]{CFCECE}{\color[HTML]{066F29} \textbf{65.92}} & \cellcolor[HTML]{CFCECE}{\color[HTML]{066F29} \textbf{92.78}} & \cellcolor[HTML]{CFCECE}{\color[HTML]{066F29} \textbf{22.77}} & \cellcolor[HTML]{CFCECE}{\color[HTML]{066F29} \textbf{74.48}} & \cellcolor[HTML]{CFCECE}{\color[HTML]{066F29} \textbf{75.37}} & \cellcolor[HTML]{CFCECE}{\color[HTML]{066F29} \textbf{97.46}} & \cellcolor[HTML]{CFCECE}{\color[HTML]{066F29} \textbf{16.80}} & \cellcolor[HTML]{CFCECE}{\color[HTML]{066F29} \textbf{62.72}} & \cellcolor[HTML]{CFCECE}{\color[HTML]{066F29} \textbf{56.71}} & \cellcolor[HTML]{CFCECE}{\color[HTML]{066F29} \textbf{89.23}} & \cellcolor[HTML]{CFCECE}{\color[HTML]{066F29} \textbf{28.64}} \\ \hline
\end{tabular}
}
\end{table*}

\begin{figure*}[htbp]
    \captionsetup{skip=5pt}
    \centering
    \includegraphics[width=\textwidth]{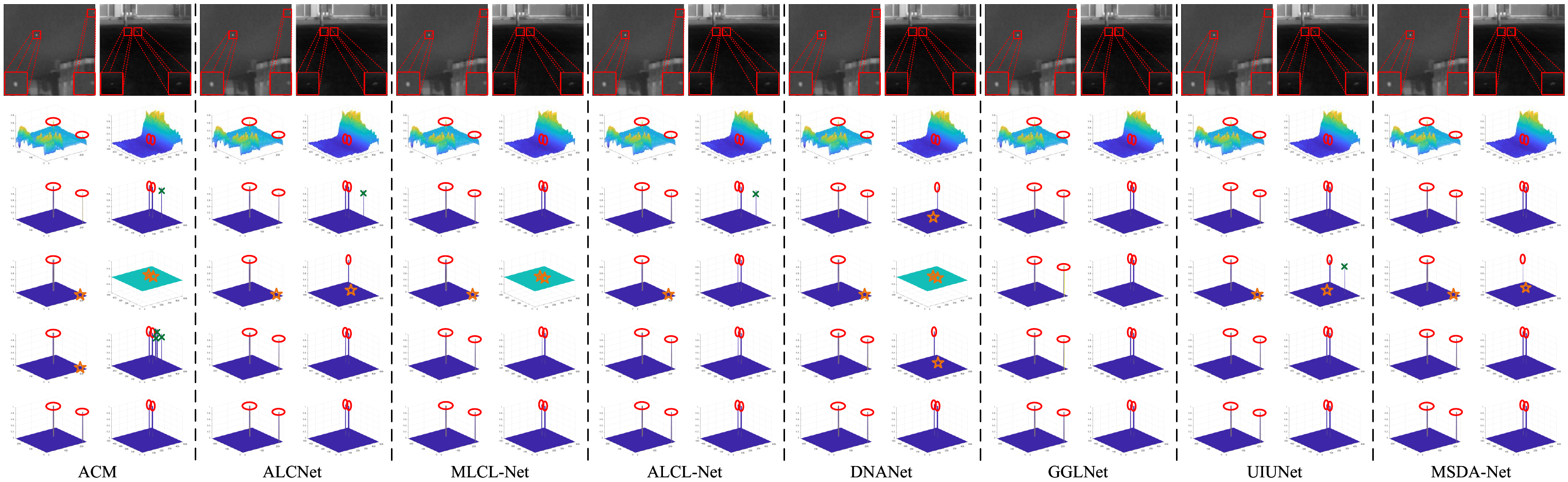}
    \caption{3D visualization on the SIRST3 dataset. The circles denote correct detections, the crosses denote false detections, and the stars denote missed detections. From top to bottom: \textit{2D image}, \textit{3D image}, \textit{DLN Full}, \textit{DLN Coarse + LESPS}, \textit{DLN Coarse + PAL}, \textit{True label}.}
    \label{fig:fig05}
    \vspace{-6pt}
\end{figure*}

\subsection{Comparison with SOTA Methods}
\label{sec:comparison with sota methods}
We embed multiple excellent SIRST detection networks ~\cite{li2022dense,yu2022infrared,yu2022pay,zhao2023gradient,dai2021attentional,zhao2025multi,dai2021asymmetric,wu2022uiu} into our framework. For fair comparison, we retrain all models with the same settings. Meanwhile, we also conduct detailed comparative experiments with various traditional methods~\cite{qin2019infrared,wei2016multiscale,dai2017non} and the latest LESPS framework~\cite{ying2023mapping}. More experiments such as MCLC~\cite{li2023monte} can be found in the Supplementary Materials.

\textbf{\textit{Evaluation on the SIRST3 dataset.}} From \cref{tab:tab01}, when the networks (UIU-Net, MSDA-Net) are embedded into the LESPS framework for single point supervision tasks, they fail to exert their performance advantages on full supervision. This is because LESPS directly uses all point labels for training and starts label evolution at fixed loss or epoch values, which can not accurately control the optimal timing for label evolution. In contrast, our PAL framework addresses this issue effectively. The performance trend of DLNs equipped with our PAL closely matches that of full supervision. At the same time, compared to LESPS, our PAL improves the IoU by 9.68\%-24.04\%, the nIoU by 14.53\%-31.06\%, and the $P_d$ by 1.99\%-12.03\% on the comprehensive SIRST3-Test. The improvement is obvious. Compared with the fully supervised task on the SIRST3-Test, our PAL framework can reach 79.33\%-86.94\% on IoU, 81.72\%-90.30\% on nIoU, and has comparable performance on $P_d$. Additionally, from the results on the three decomposed test subsets, our PAL has stable performance. Furthermore, from \cref{fig:fig05}, DLNs equipped with our PAL have target-level detection effects similar to those of full supervision, and they are significantly better than the LESPS.

\textbf{\textit{Evaluation on three individual datasets.}} To further explore the stability of our PAL framework under limited samples, we perform separate training and testing on the three individual datasets. From \cref{tab:tab02}, The PAL consistently outperforms the LESPS with notably higher stability. These show that our PAL framework can solve the single point supervised SIRST detection task with few training samples, whereas the LESPS framework suffers from a high risk of failure. Specifically, compared with the LESPS, using the PAL framework improves the IoU by 7.41\%-57.28\%, the nIoU by 8.36\%-62.37\%, and the $P_d$ by 0.76\%-70.72\%.

\subsection{Ablation Experiment}
\label{sec:ablation experiment}
To fully validate the proposed PAL framework, we perform detailed ablation experiments. More experiments can be found in the Supplementary Materials.

\textbf{\textit{1) Break-Down Ablation.}} To verify the effectiveness of model pre-start and fine dual-update strategy, \cref{tab:tab03} presents detailed ablation experiments. From Scheme 1 and Scheme 2, training directly with point labels will fail, but the network after applying the model pre-start will show significant performance improvement. This verifies the effectiveness of our model pre-start concept and EPG strategy. From Schemes 2-4, whether using COU or FIU alone can significantly improve the performance. In addition, from Schemes 2-5, when the COU and the FIU are used in combination, a significant performance improvement of ``$1+1>2$'' is achieved. This fully verifies the effectiveness of the proposed fine dual-update strategy.

\textbf{\textit{2) Decay Factor.}} As presented in \cref{tab:tab04}, compared with not using the decay factor ($ \lambda = 1 $), using decay factors can improve the segmentation accuracy. The IoU will increase by 0.04\%-1.48\% and the nIoU will increase by 1.10\%-2.92\%. At the same time, using a decay factor within a reasonable range will also lead to an increase in $P_d$. We finally set $\lambda$ to 0.97.

\begin{table}[]
\caption{Break-down ablation experiments on the SIRST3 dataset. \textit{MP} is the model pre-start. \textit{Coarse} denotes coarse point supervision. \textit{Centroid} denotes centroid point supervision.}
\vspace{-6pt}
\label{tab:tab03}
\setlength{\tabcolsep}{1mm}
\resizebox{\columnwidth}{!}{
\begin{tabular}{c|ccc|cccc|cccc}
\hline
                         & \multicolumn{3}{c|}{Variants}        & \multicolumn{4}{c|}{MSDA-Net Coarse + PAL}                                                                                                                   & \multicolumn{4}{c}{MSDA-Net Centroid + PAL}                                                                                                                                                                                          \\ \cline{2-12} 
\multirow{-2}{*}{Scheme} & MP         & COU        & FIU        & $IoU$                                   & $IoU$                                  & $P_d$                                    & $F_a$                                   & $IoU$                                                           & $nIoU$                                                          & $P_d$                                                            & $F_a$                                   \\ \hline
Scheme1                  & -          & -          & -          & 1.23                                  & 2.60                                  & 40.47                                 & {\color[HTML]{066F29} \textbf{0.26}} & 2.56                                                          & 4.94                                                          & 83.52                                                         & {\color[HTML]{066F29} \textbf{0.61}} \\ \hline
Scheme2                  & \ding{51} & \ding{55} & \ding{55} & 56.24                                 & 64.20                                 & 89.63                                 & 31.48                                & 57.25                                                         & 64.44                                                         & 91.83                                                         & 23.83                                \\
Scheme3                  & \ding{51} & \ding{51} & \ding{55} & 61.64                                 & 67.14                                 & 94.02                                 & 31.15                                & 60.19                                                         & 67.17                                                         & 95.75                                                         & 18.74                                \\
Scheme4                  & \ding{51} & \ding{55} & \ding{51} & 61.66                                 & 66.78                                 & 93.42                                 & 39.47                                & 59.81                                                         & 65.38                                                         & 92.89                                                         & 25.06                                \\ \hline
\rowcolor[HTML]{CFCECE} 
Scheme5                  & \ding{51} & \ding{51} & \ding{51} & {\color[HTML]{066F29} \textbf{69.38}} & {\color[HTML]{066F29} \textbf{71.55}} & {\color[HTML]{066F29} \textbf{97.41}} & 16.34                                & \cellcolor[HTML]{D7D7D7}{\color[HTML]{066F29} \textbf{69.21}} & \cellcolor[HTML]{D7D7D7}{\color[HTML]{066F29} \textbf{72.40}} & \cellcolor[HTML]{D7D7D7}{\color[HTML]{066F29} \textbf{97.01}} & \cellcolor[HTML]{D7D7D7}15.70        \\ \hline
\end{tabular}
}
\vspace{-2pt}
\end{table}

\begin{table}[]
\caption{Decay factor investigation on the SIRST3 dataset.}
\vspace{-6pt}
\label{tab:tab04}
\setlength{\tabcolsep}{1.5mm}
\resizebox{\columnwidth}{!}{
\begin{tabular}{c|cccc|c|cccc}
\hline
                             & \multicolumn{4}{c|}{MSDA-Net Coarse + PAL}                                                                                                                                                    &                     & \multicolumn{4}{c}{MSDA-Net Coarse + PAL}                                                     \\ \cline{2-5} \cline{7-10} 
\multirow{-2}{*}{$\lambda$}          & $IoU$                                                           & $nIoU$                          & $P_d$                                                            & $F_a$                            & \multirow{-2}{*}{$\lambda$} & $IoU$   & $nIoU$                                  & $P_d$    & $F_a$                                    \\ \hline
1.00                         & 67.90                                                         & 69.67                         & 96.68                                                         & 18.96                         & 0.95                & 68.96 & 71.40                                 & 97.41 & {\color[HTML]{066F29} \textbf{15.76}} \\
0.99                         & 68.02                                                         & 70.94                         & 96.74                                                         & 19.13                         & 0.90                & 68.92 & 72.00                                 & 96.48 & 16.35                                 \\
0.98                         & 68.92                                                         & 71.20                         & 96.88                                                         & 19.02                         & 0.85                & 68.73 & {\color[HTML]{066F29} \textbf{72.59}} & 96.41 & 18.66                                 \\
\cellcolor[HTML]{CFCECE}0.97 & \cellcolor[HTML]{CFCECE}{\color[HTML]{066F29} \textbf{69.38}} & \cellcolor[HTML]{CFCECE}71.55 & \cellcolor[HTML]{CFCECE}{\color[HTML]{066F29} \textbf{97.41}} & \cellcolor[HTML]{CFCECE}16.34 & 0.80                & 67.94 & 71.30                                 & 96.15 & 20.36                                 \\
0.96                         & 69.13                                                         & 70.77                         & 97.28                                                         & 16.52                         & 0.75                & 68.11 & 71.60                                 & 95.55 & 29.45                                 \\ \hline
\end{tabular}
}
\vspace{-2pt}
\end{table}

\begin{table}[]
\caption{Comparison of different losses on the SIRST3 dataset.}
\vspace{-6pt}
\label{tab:tab05}
\setlength{\tabcolsep}{1.5mm}
\resizebox{\columnwidth}{!}{
\begin{tabular}{c|cccc|cccc}
\hline
                                & \multicolumn{4}{c|}{MSDA-Net Coarse + PAL}                                                                                                                    & \multicolumn{4}{c}{MSDA-Net Centroid + PAL}                                                                                                                   \\ \cline{2-9} 
\multirow{-2}{*}{Loss Function} & $IoU$                                   & $nIoU$                                  & $P_d$                                    & $F_a$                                    & $IoU$                                   & $nIoU$                                  & $P_d$                                    & $F_a$                                    \\ \hline
BCE loss                        & 69.26                                 & 71.48                                 & 97.08                                 & {\color[HTML]{066F29} \textbf{15.98}} & 69.19                                 & 71.67                                 & 96.74                                 & {\color[HTML]{066F29} \textbf{14.25}} \\
Dice loss                       & 64.38                                 & 66.64                                 & 94.68                                 & 39.83                                 & 65.18                                 & 67.71                                 & 95.61                                 & 35.61                                 \\
Focal loss                      & 68.89                                 & {\color[HTML]{066F29} \textbf{72.21}} & 97.01                                 & 17.02                                 & 69.14                                 & 71.85                                 & 96.28                                 & 16.18                                 \\
\rowcolor[HTML]{CFCECE} 
EEDM loss                       & {\color[HTML]{066F29} \textbf{69.38}} & 71.55                                 & {\color[HTML]{066F29} \textbf{97.41}} & 16.34                                 & {\color[HTML]{066F29} \textbf{69.21}} & {\color[HTML]{066F29} \textbf{72.40}} & {\color[HTML]{066F29} \textbf{97.01}} & 15.70                                 \\ \hline
\end{tabular}
}
\vspace{-2pt}
\end{table}

\begin{table}[]
\caption{Comparison of the PAL framework with different settings on the SIRST3 dataset. \textit{MP} denotes the proposed model pre-start. \textit{MP-1} and \textit{MP-2} denote the model pre-start with all samples initially enter the training pool by setting the target recall threshold to 0 and directly carry the true point labels for remaining hard samples into the training pool. \textit{S} denotes scheme.}
\vspace{-6pt}
\label{tab:tab06}
\setlength{\tabcolsep}{0.8mm}
\resizebox{\columnwidth}{!}{
\begin{tabular}{c|ccccc|cccc|cccc}
\hline
                    & \multicolumn{5}{c|}{Variants}                                  & \multicolumn{4}{c|}{\begin{tabular}[c]{@{}c@{}}MSDA-Net \\ Coarse + PAL\end{tabular}}                                                                         & \multicolumn{4}{c}{\begin{tabular}[c]{@{}c@{}}MSDA-Net \\ Centroid + PAL\end{tabular}}                                                                        \\ \cline{2-14} 
\multirow{-2}{*}{S} & MP         & MP-1       & MP-2       & COU        & FIU        & $IoU$                                   & $nIoU$                                  & $P_d$                                    & $F_a$                                    & $IoU$                                   & $nIoU$                                  & $P_d$                                   & $F_a$                                    \\ \hline
S1                  & \ding{51} & \ding{55} & \ding{55} & \ding{55} & \ding{55} & 56.24                                 & 64.20                                 & 89.63                                 & 31.48                                 & 57.25                                 & 64.44                                 & 91.83                                 & 23.83                                 \\
S2                  & \ding{55} & \ding{51} & \ding{55} & \ding{55} & \ding{55} & 32.86                                 & 45.42                                 & 73.62                                 & 13.62                                 & 32.45                                 & 44.18                                 & 73.49                                 & 13.19                                 \\
S3                  & \ding{55} & \ding{55} & \ding{51} & \ding{55} & \ding{55} & 25.72                                 & 38.26                                 & 61.53                                 & 14.14                                 & 28.77                                 & 41.86                                 & 71.56                                 & 11.67                                 \\ \hline
S4                  & \ding{51} & \ding{55} & \ding{55} & \ding{55} & \ding{51} & 61.66                                 & 66.78                                 & 93.42                                 & 39.47                                 & 59.81                                 & 65.38                                 & 92.89                                 & 25.06                                 \\
S5                  & \ding{55} & \ding{51} & \ding{55} & \ding{55} & \ding{51} & 40.84                                 & 51.23                                 & 74.82                                 & 18.88                                 & 41.87                                 & 52.96                                 & 78.80                                 & 14.15                                 \\
S6                  & \ding{55} & \ding{55} & \ding{51} & \ding{55} & \ding{51} & 36.67                                 & 48.88                                 & 71.10                                 & {\color[HTML]{066F29} \textbf{12.10}} & 37.13                                 & 48.75                                 & 75.08                                 & {\color[HTML]{066F29} \textbf{11.37}} \\ \hline
\rowcolor[HTML]{CFCECE} 
S7                  & \ding{51} & \ding{55} & \ding{55} & \ding{51} & \ding{51} & {\color[HTML]{066F29} \textbf{69.38}} & {\color[HTML]{066F29} \textbf{71.55}} & {\color[HTML]{066F29} \textbf{97.41}} & 16.34                                 & {\color[HTML]{066F29} \textbf{69.21}} & {\color[HTML]{066F29} \textbf{72.40}} & {\color[HTML]{066F29} \textbf{97.01}} & 15.70                                 \\ \hline
\end{tabular}
}
\vspace{-2pt}
\end{table}

\begin{figure}[t]
  \captionsetup{skip=5pt}
  \centering
   \includegraphics[width=\columnwidth]{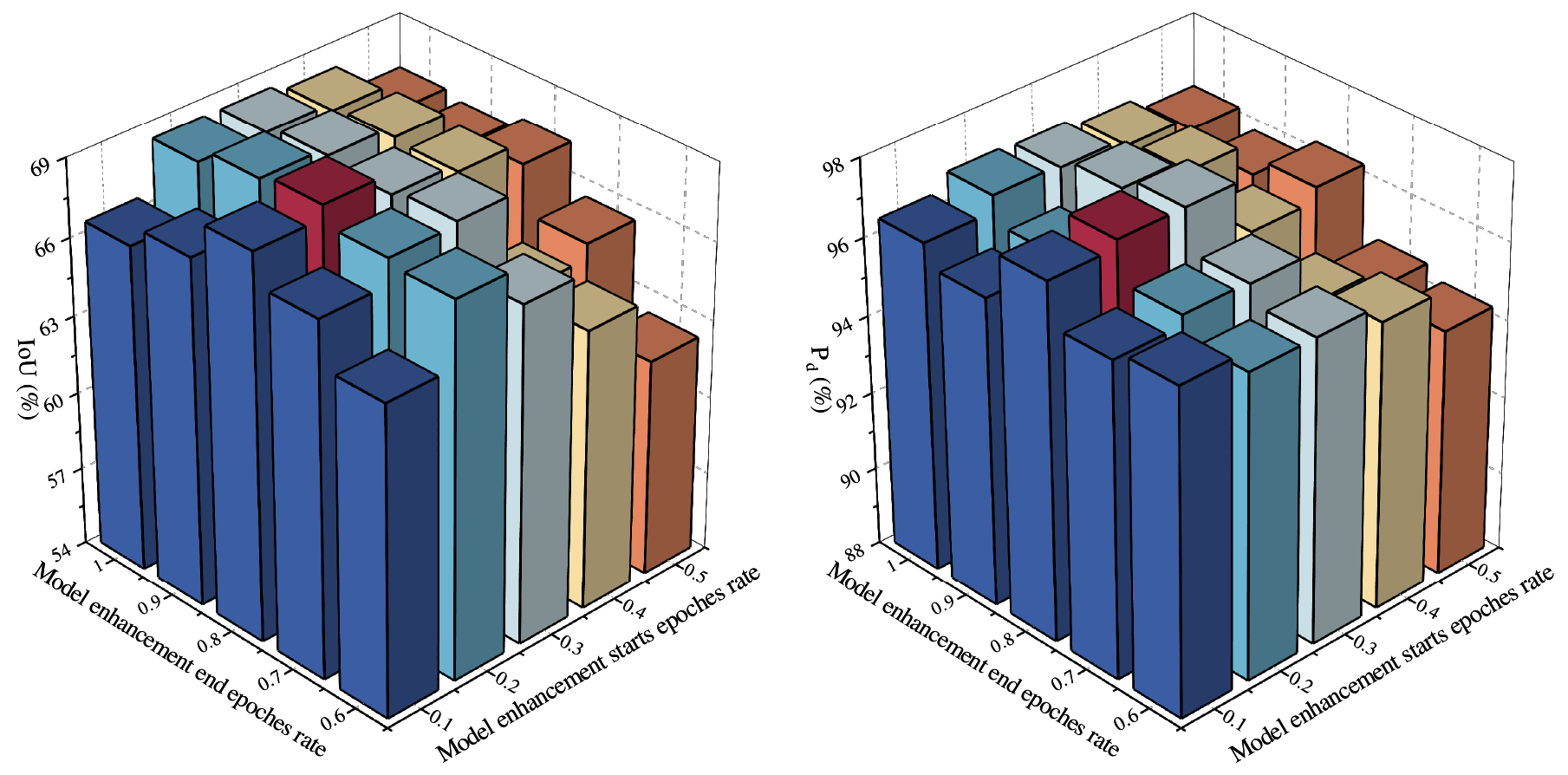}
   \caption{Epoch division investigation on the SIRST3 dataset. \textcolor{darkred}{\textit{Red}} denotes the optimal value.}
   \label{fig:fig06}
   \vspace{-8pt}
\end{figure}

\textbf{\textit{3) Epoches Division.}} As presented in \cref{fig:fig06}, when the model pre-start interval is greater than 0.1 and the model enhancement interval is from 0.1 to 0.4, the performance shows an upward trend. This shows that subdividing all the hard samples multiple times will have a more positive performance improvement. It is like a student model with basic abilities that refines the content to be learned in more detail and then learns from easy to hard. At the same time, when the model refinement phase is eliminated or the model pre-start interval is only 0.1, the performance of the network will degrade. This proves the superiority of the 3-phase design in our PAL framework. In addition, when the model pre-start interval is greater than 0.1 and the model enhancement interval is greater than 0.3, the performance is stable. It demonstrates the robustness of our PAL framework.

\textbf{\textit{4) Loss Function.}} We compare EEDM loss with the binary cross-entropy loss, Dice loss~\cite{milletari2016v} and focal loss~\cite{lin2017focal}. From \cref{tab:tab05}, compared with other losses, EEDM loss has the best results. Using EEDM loss can improve the IoU by 0.02\%-5.00\% and improve the $P_d$ by 0.33\%-2.73\%.

\subsection{Discussion of ``From easy to hard''}
\label{sec:discussion of from easy to hard}
To further discuss the proposed PAL that focuses on learning from easy samples to hard samples, we conduct detailed experiments. From S1-S6 in \cref{tab:tab06}, regardless of whether FIU are added or not, the final model generated by all samples entering the training pool in the initial stage is far worse than that of using only easy samples. This further verifies the effectiveness of our model pre-start concept of learning only easy samples in the initial phase. It's like babies should be given pronunciation exercises rather than forced to learn calculus. In addition, from S4 and S7 in \cref{tab:tab06}, compared to using only easy samples, gradually selecting hard samples into the training pool based on the current capabilities will again achieve significant performance improvements. This further verifies the effectiveness of the idea of learning from easy samples to hard samples. More visualization results can be found in the Supplementary Materials.

\section{Conclusion}
\label{sec:conclusion}
This paper proposes an innovative PAL framework for the fine-grained SIRST detection with single point supervision, which drives the existing SIRST detection networks progressively and actively recognizes and learns harder samples to achieve significant performance improvements. First, we propose a model pre-start concept, which can help models have basic task-specific learning capabilities in the initial training phase. Second, a refined dual-update strategy is proposed, which can promote reasonable learning of harder samples and continuous refinement of pseudo-labels. Finally, to alleviate the risk of region expansion without shrinkage, a decay factor is reasonably introduced to achieve a dynamic balance between the expansion and shrinkage of target annotations. Extensive comparison results demonstrate the superiority of our PAL framework. We hope that our study can draw attention to the research on weakly supervised SIRST detection.


\section*{Acknowledgements}
\vspace{-2pt}
This work was partially supported by the National Natural Science Foundation of China (Grant No.\ 62306261), the Shun Hing Institute of Advanced Engineering (SHIAE) Fund (Grant No.\ 8115074), the LiaoNing Revitalization Talents Program (Grant No.\ XLYC2201001), and the Centre for Perceptual and Interactive Intelligence (CPII) Ltd., a CUHK-led InnoCentre under the InnoHK initiative of the Innovation and Technology Commission of the Hong Kong Special Administrative Region Government.

{
    \small
    \bibliographystyle{ieeenat_fullname}
    \bibliography{main}
}

\appendix
\renewcommand{\thesection}{\Alph{section}}
\setcounter{section}{0}
\renewcommand{\thefigure}{S\arabic{figure}}
\setcounter{figure}{0}  
\renewcommand{\thetable}{S\arabic{table}}
\setcounter{table}{0}

\twocolumn[{
\renewcommand\twocolumn[1][]{#1}
\maketitlesupplementary
\vspace{-7pt}
\begin{center}
    \captionsetup{type=figure}
    \includegraphics[width=1.0\textwidth]{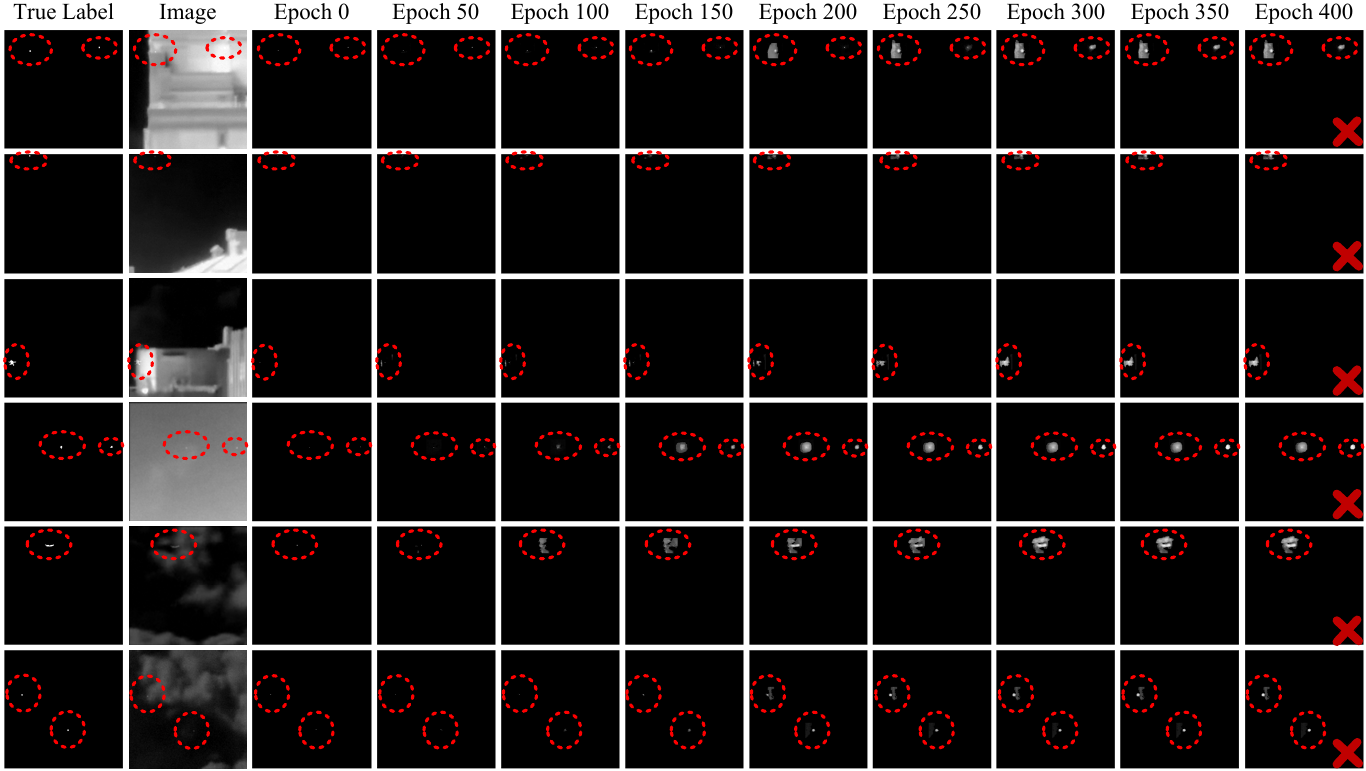}
    \captionof{figure}{Label evolution results of MSDA-Net equipped with the LESPS framework on the NUDT-SIRST dataset. As the number of epochs increases, the labelled area expands excessively and does not shrink, which will affect the final detection performance.}
    \label{fig:s-fig01}
\end{center}
}]
\vspace{3pt}


In this supplementary material, we offer extra details and additional results to complement the main paper. In \cref{sec:excessive label evolution in the lesps}, we provide a presentation and analysis of the risk of excessive label evolution in the LESPS framework. In \cref{sec:the eedm loss}, we provide a detailed introduction to the used edge-enhanced difficulty-mining (EEDM) loss. In \cref{sec:why}, we provide a more detailed explanation on why ``from easy to hard'' fits this task and more visualizations. In \cref{sec:compared with other methods}, we provide a detailed performance comparison with other methods (MCLC~\cite{li2023monte}, LELCM~\cite{yang2024label}). In \cref{sec:more ablation experiments}, we provide more ablation experiments to fully explore the performance of our proposed Progressive Active Learning (PAL) framework. In \cref{sec:more quantitative results}, we provide more quantitative comparative experiments on different datasets. In \cref{sec:more qualitative results}, we provide more qualitative results to further verify the superiority of the proposed PAL framework.

\section{Excessive Label Evolution in the LESPS}
\label{sec:excessive label evolution in the lesps}
In exploring the LESPS framework~\cite{ying2023mapping}, we find that it has the risk of excessive label evolution. From \cref{fig:s-fig01}, when using the LESPS framework, if the pseudo-label has an overly large annotation of the target area during evolution, the area will not shrink, but will either remain the same or expand further. The reason is that the label evolution rule does not consider the shrinkage problem of the target annotation in the pseudo-label after it is too large. The LESPS framework is designed to generate reasonable candidate regions using an adaptive threshold rule, which effectively avoids cumulative errors. However, it can only prevent the continued large-scale expansion and cannot prevent the initial occurrence of errors. At the same time, it ignores the fact that the calculation of the adaptive threshold is based on the target area in the pseudo-label rather than the prediction result of the current iteration. Therefore, when the target annotation of the pseudo-label is small and the annotation of the current prediction result is too large, there is a risk of over-expansion in the target annotation of the updated pseudo-label. Since there is no design for shrinking the annotation area in its pseudo-label update strategy, an overly expanded annotation area cannot be shrunk. Therefore, to reduce the risk, we introduce a decay factor, which helps achieve a dynamic balance between the expansion and contraction of target annotations.

\begin{table}[t]
\caption{Performance comparison of the PAL and MCLC on the SIRST3 dataset with coarse point.}
\vspace{-6pt}
\captionsetup{skip=4pt}
\label{tab:tab-mclc-coarse}
\centering
\renewcommand{\arraystretch}{1.1} 
\resizebox{1.0\columnwidth}{!}{
\setlength{\tabcolsep}{0.8mm}{
\begin{tabular}{c|c|cccc|c|c|cccc}
\hline
                                               &                          & \multicolumn{4}{c|}{SIRST3-Test}                                                                                                                                                                                                                                                                     &                                                &                          & \multicolumn{4}{c}{SIRST3-Test}                                                                                                                                                                                                                                                                      \\ \cline{3-6} \cline{9-12} 
\multirow{-2}{*}{Net}                          & \multirow{-2}{*}{Method} & \multicolumn{1}{c|}{$IoU$}                                                           & \multicolumn{1}{c|}{$nIoU$}                                                          & \multicolumn{1}{c|}{$P_d$}                                                            & $F_a$                                    & \multirow{-2}{*}{Net}                          & \multirow{-2}{*}{Method} & \multicolumn{1}{c|}{$IoU$}                                                           & \multicolumn{1}{c|}{$nIoU$}                                                          & \multicolumn{1}{c|}{$P_d$}                                                            & $F_a$                                    \\ \hline
\rowcolor[HTML]{FFFFFF} 
\cellcolor[HTML]{FFFFFF}                       & MCLC                     & \multicolumn{1}{c|}{\cellcolor[HTML]{FFFFFF}48.17}                                 & \multicolumn{1}{c|}{\cellcolor[HTML]{FFFFFF}49.94}                                 & \multicolumn{1}{c|}{\cellcolor[HTML]{FFFFFF}85.45}                                 & 110.30                                & \cellcolor[HTML]{FFFFFF}                       & MCLC                     & \multicolumn{1}{c|}{\cellcolor[HTML]{FFFFFF}54.57}                                 & \multicolumn{1}{c|}{\cellcolor[HTML]{FFFFFF}59.94}                                 & \multicolumn{1}{c|}{\cellcolor[HTML]{FFFFFF}87.04}                                 & 102.87                                \\ \cline{2-6} \cline{8-12} 
\rowcolor[HTML]{CFCECE} 
\multirow{-2}{*}{\cellcolor[HTML]{FFFFFF}ACM}  & PAL                      & \multicolumn{1}{c|}{\cellcolor[HTML]{CFCECE}{\color[HTML]{066F29} \textbf{51.51}}} & \multicolumn{1}{c|}{\cellcolor[HTML]{CFCECE}{\color[HTML]{066F29} \textbf{54.07}}} & \multicolumn{1}{c|}{\cellcolor[HTML]{CFCECE}{\color[HTML]{066F29} \textbf{92.89}}} & {\color[HTML]{066F29} \textbf{39.18}} & \multirow{-2}{*}{\cellcolor[HTML]{FFFFFF}DNA}  & PAL                      & \multicolumn{1}{c|}{\cellcolor[HTML]{CFCECE}{\color[HTML]{066F29} \textbf{67.20}}} & \multicolumn{1}{c|}{\cellcolor[HTML]{CFCECE}{\color[HTML]{066F29} \textbf{70.20}}} & \multicolumn{1}{c|}{\cellcolor[HTML]{CFCECE}{\color[HTML]{066F29} \textbf{96.15}}} & {\color[HTML]{066F29} \textbf{10.86}} \\ \hline
\rowcolor[HTML]{FFFFFF} 
\cellcolor[HTML]{FFFFFF}                       & MCLC                     & \multicolumn{1}{c|}{\cellcolor[HTML]{FFFFFF}51.05}                                 & \multicolumn{1}{c|}{\cellcolor[HTML]{FFFFFF}53.14}                                 & \multicolumn{1}{c|}{\cellcolor[HTML]{FFFFFF}82.99}                                 & 85.10                                 & \cellcolor[HTML]{FFFFFF}                       & MCLC                     & \multicolumn{1}{c|}{\cellcolor[HTML]{FFFFFF}55.54}                                 & \multicolumn{1}{c|}{\cellcolor[HTML]{FFFFFF}61.96}                                 & \multicolumn{1}{c|}{\cellcolor[HTML]{FFFFFF}88.24}                                 & 129.56                                \\ \cline{2-6} \cline{8-12} 
\rowcolor[HTML]{CFCECE} 
\multirow{-2}{*}{\cellcolor[HTML]{FFFFFF}ALC}  & PAL                      & \multicolumn{1}{c|}{\cellcolor[HTML]{CFCECE}{\color[HTML]{066F29} \textbf{57.11}}} & \multicolumn{1}{c|}{\cellcolor[HTML]{CFCECE}{\color[HTML]{066F29} \textbf{60.22}}} & \multicolumn{1}{c|}{\cellcolor[HTML]{CFCECE}{\color[HTML]{066F29} \textbf{93.95}}} & {\color[HTML]{066F29} \textbf{37.20}} & \multirow{-2}{*}{\cellcolor[HTML]{FFFFFF}GGL}  & PAL                      & \multicolumn{1}{c|}{\cellcolor[HTML]{CFCECE}{\color[HTML]{066F29} \textbf{68.52}}} & \multicolumn{1}{c|}{\cellcolor[HTML]{CFCECE}{\color[HTML]{066F29} \textbf{71.69}}} & \multicolumn{1}{c|}{\cellcolor[HTML]{CFCECE}{\color[HTML]{066F29} \textbf{97.14}}} & {\color[HTML]{066F29} \textbf{16.69}} \\ \hline
\rowcolor[HTML]{FFFFFF} 
\cellcolor[HTML]{FFFFFF}                       & MCLC                     & \multicolumn{1}{l|}{\cellcolor[HTML]{FFFFFF}52.26}                                 & \multicolumn{1}{c|}{\cellcolor[HTML]{FFFFFF}58.06}                                 & \multicolumn{1}{c|}{\cellcolor[HTML]{FFFFFF}89.57}                                 & 136.18                                & \cellcolor[HTML]{FFFFFF}                       & MCLC                     & \multicolumn{1}{c|}{\cellcolor[HTML]{FFFFFF}54.56}                                 & \multicolumn{1}{c|}{\cellcolor[HTML]{FFFFFF}62.21}                                 & \multicolumn{1}{c|}{\cellcolor[HTML]{FFFFFF}87.97}                                 & 164.06                                \\ \cline{2-6} \cline{8-12} 
\rowcolor[HTML]{CFCECE} 
\multirow{-2}{*}{\cellcolor[HTML]{FFFFFF}MLCL} & PAL                      & \multicolumn{1}{c|}{\cellcolor[HTML]{CFCECE}{\color[HTML]{066F29} \textbf{64.87}}} & \multicolumn{1}{c|}{\cellcolor[HTML]{CFCECE}{\color[HTML]{066F29} \textbf{69.40}}} & \multicolumn{1}{c|}{\cellcolor[HTML]{CFCECE}{\color[HTML]{066F29} \textbf{94.95}}} & {\color[HTML]{066F29} \textbf{24.43}} & \multirow{-2}{*}{\cellcolor[HTML]{FFFFFF}UIU}  & PAL                      & \multicolumn{1}{c|}{\cellcolor[HTML]{CFCECE}{\color[HTML]{066F29} \textbf{69.05}}} & \multicolumn{1}{c|}{\cellcolor[HTML]{CFCECE}{\color[HTML]{066F29} \textbf{71.53}}} & \multicolumn{1}{c|}{\cellcolor[HTML]{CFCECE}{\color[HTML]{066F29} \textbf{96.81}}} & {\color[HTML]{066F29} \textbf{15.45}} \\ \hline
\rowcolor[HTML]{FFFFFF} 
\cellcolor[HTML]{FFFFFF}                       & MCLC                     & \multicolumn{1}{c|}{\cellcolor[HTML]{FFFFFF}53.82}                                 & \multicolumn{1}{c|}{\cellcolor[HTML]{FFFFFF}58.69}                                 & \multicolumn{1}{c|}{\cellcolor[HTML]{FFFFFF}86.38}                                 & 109.44                                & \cellcolor[HTML]{FFFFFF}                       & MCLC                     & \multicolumn{1}{c|}{\cellcolor[HTML]{FFFFFF}54.71}                                 & \multicolumn{1}{c|}{\cellcolor[HTML]{FFFFFF}60.89}                                 & \multicolumn{1}{c|}{\cellcolor[HTML]{FFFFFF}88.90}                                 & 132.39                                \\ \cline{2-6} \cline{8-12} 
\rowcolor[HTML]{CFCECE} 
\multirow{-2}{*}{\cellcolor[HTML]{FFFFFF}ALCL} & PAL                      & \multicolumn{1}{c|}{\cellcolor[HTML]{CFCECE}{\color[HTML]{066F29} \textbf{66.29}}} & \multicolumn{1}{c|}{\cellcolor[HTML]{CFCECE}{\color[HTML]{066F29} \textbf{68.18}}} & \multicolumn{1}{c|}{\cellcolor[HTML]{CFCECE}{\color[HTML]{066F29} \textbf{94.75}}} & {\color[HTML]{066F29} \textbf{18.79}} & \multirow{-2}{*}{\cellcolor[HTML]{FFFFFF}MSDA} & PAL                      & \multicolumn{1}{c|}{\cellcolor[HTML]{CFCECE}{\color[HTML]{066F29} \textbf{69.38}}} & \multicolumn{1}{c|}{\cellcolor[HTML]{CFCECE}{\color[HTML]{066F29} \textbf{71.55}}} & \multicolumn{1}{c|}{\cellcolor[HTML]{CFCECE}{\color[HTML]{066F29} \textbf{97.41}}} & {\color[HTML]{066F29} \textbf{16.34}} \\ \hline
\end{tabular}
}}
\end{table}

\begin{table}[]
\caption{Performance comparison of the PAL and MCLC on the SIRST3 dataset with centroid point.}
\vspace{-6pt}
\captionsetup{skip=4pt}
\label{tab:tab-mclc-centroid}
\centering
\renewcommand{\arraystretch}{1.1} 
\resizebox{1.0\columnwidth}{!}{
\setlength{\tabcolsep}{0.8mm}{
\begin{tabular}{c|c|cccc|c|c|cccc}
\hline
                                               &                          & \multicolumn{4}{c|}{SIRST3-Test}                                                                                                                                                                                                                                                                     &                                                &                          & \multicolumn{4}{c}{SIRST3-Test}                                                                                                                                                                                                                                                                      \\ \cline{3-6} \cline{9-12} 
\multirow{-2}{*}{Net}                          & \multirow{-2}{*}{Method} & \multicolumn{1}{c|}{$IoU$}                                                           & \multicolumn{1}{c|}{$nIoU$}                                                          & \multicolumn{1}{c|}{$P_d$}                                                            & $F_a$                                   & \multirow{-2}{*}{Net}                          & \multirow{-2}{*}{Method} & \multicolumn{1}{c|}{$IoU$}                                                           & \multicolumn{1}{c|}{$nIoU$}                                                          & \multicolumn{1}{c|}{$P_d$}                                                            & $F_a$                                    \\ \hline
\rowcolor[HTML]{FFFFFF} 
\cellcolor[HTML]{FFFFFF}                       & MCLC                     & \multicolumn{1}{c|}{\cellcolor[HTML]{FFFFFF}47.87}                                 & \multicolumn{1}{c|}{\cellcolor[HTML]{FFFFFF}49.57}                                 & \multicolumn{1}{c|}{\cellcolor[HTML]{FFFFFF}85.51}                                 & 133.21                                & \cellcolor[HTML]{FFFFFF}                       & MCLC                     & \multicolumn{1}{c|}{\cellcolor[HTML]{FFFFFF}55.92}                                 & \multicolumn{1}{c|}{\cellcolor[HTML]{FFFFFF}62.53}                                 & \multicolumn{1}{c|}{\cellcolor[HTML]{FFFFFF}87.97}                                 & 110.28                                \\ \cline{2-6} \cline{8-12} 
\rowcolor[HTML]{CFCECE} 
\multirow{-2}{*}{\cellcolor[HTML]{FFFFFF}ACM}  & PAL                      & \multicolumn{1}{c|}{\cellcolor[HTML]{CFCECE}{\color[HTML]{066F29} \textbf{51.51}}} & \multicolumn{1}{c|}{\cellcolor[HTML]{CFCECE}{\color[HTML]{066F29} \textbf{53.73}}} & \multicolumn{1}{c|}{\cellcolor[HTML]{CFCECE}{\color[HTML]{066F29} \textbf{92.82}}} & {\color[HTML]{066F29} \textbf{35.98}} & \multirow{-2}{*}{\cellcolor[HTML]{FFFFFF}DNA}  & PAL                      & \multicolumn{1}{c|}{\cellcolor[HTML]{CFCECE}{\color[HTML]{066F29} \textbf{66.97}}} & \multicolumn{1}{c|}{\cellcolor[HTML]{CFCECE}{\color[HTML]{066F29} \textbf{70.63}}} & \multicolumn{1}{c|}{\cellcolor[HTML]{CFCECE}{\color[HTML]{066F29} \textbf{96.28}}} & {\color[HTML]{066F29} \textbf{14.66}} \\ \hline
\rowcolor[HTML]{FFFFFF} 
\cellcolor[HTML]{FFFFFF}                       & MCLC                     & \multicolumn{1}{c|}{\cellcolor[HTML]{FFFFFF}49.82}                                 & \multicolumn{1}{c|}{\cellcolor[HTML]{FFFFFF}52.85}                                 & \multicolumn{1}{c|}{\cellcolor[HTML]{FFFFFF}85.65}                                 & 105.89                                & \cellcolor[HTML]{FFFFFF}                       & MCLC                     & \multicolumn{1}{c|}{\cellcolor[HTML]{FFFFFF}55.82}                                 & \multicolumn{1}{c|}{\cellcolor[HTML]{FFFFFF}62.04}                                 & \multicolumn{1}{c|}{\cellcolor[HTML]{FFFFFF}87.91}                                 & 107.26                                \\ \cline{2-6} \cline{8-12} 
\rowcolor[HTML]{CFCECE} 
\multirow{-2}{*}{\cellcolor[HTML]{FFFFFF}ALC}  & PAL                      & \multicolumn{1}{c|}{\cellcolor[HTML]{CFCECE}{\color[HTML]{066F29} \textbf{55.01}}} & \multicolumn{1}{c|}{\cellcolor[HTML]{CFCECE}{\color[HTML]{066F29} \textbf{57.93}}} & \multicolumn{1}{c|}{\cellcolor[HTML]{CFCECE}{\color[HTML]{066F29} \textbf{93.94}}} & {\color[HTML]{066F29} \textbf{31.63}} & \multirow{-2}{*}{\cellcolor[HTML]{FFFFFF}GGL}  & PAL                      & \multicolumn{1}{c|}{\cellcolor[HTML]{CFCECE}{\color[HTML]{066F29} \textbf{67.83}}} & \multicolumn{1}{c|}{\cellcolor[HTML]{CFCECE}{\color[HTML]{066F29} \textbf{70.27}}} & \multicolumn{1}{c|}{\cellcolor[HTML]{CFCECE}{\color[HTML]{066F29} \textbf{95.68}}} & {\color[HTML]{066F29} \textbf{16.28}} \\ \hline
\rowcolor[HTML]{FFFFFF} 
\cellcolor[HTML]{FFFFFF}                       & MCLC                     & \multicolumn{1}{l|}{\cellcolor[HTML]{FFFFFF}53.83}                                 & \multicolumn{1}{c|}{\cellcolor[HTML]{FFFFFF}59.31}                                 & \multicolumn{1}{c|}{\cellcolor[HTML]{FFFFFF}88.04}                                 & 99.71                                 & \cellcolor[HTML]{FFFFFF}                       & MCLC                     & \multicolumn{1}{c|}{\cellcolor[HTML]{FFFFFF}55.97}                                 & \multicolumn{1}{c|}{\cellcolor[HTML]{FFFFFF}62.49}                                 & \multicolumn{1}{c|}{\cellcolor[HTML]{FFFFFF}87.71}                                 & 117.33                                \\ \cline{2-6} \cline{8-12} 
\rowcolor[HTML]{CFCECE} 
\multirow{-2}{*}{\cellcolor[HTML]{FFFFFF}MLCL} & PAL                      & \multicolumn{1}{c|}{\cellcolor[HTML]{CFCECE}{\color[HTML]{066F29} \textbf{66.38}}} & \multicolumn{1}{c|}{\cellcolor[HTML]{CFCECE}{\color[HTML]{066F29} \textbf{69.25}}} & \multicolumn{1}{c|}{\cellcolor[HTML]{CFCECE}{\color[HTML]{066F29} \textbf{95.08}}} & {\color[HTML]{066F29} \textbf{15.94}} & \multirow{-2}{*}{\cellcolor[HTML]{FFFFFF}UIU}  & PAL                      & \multicolumn{1}{c|}{\cellcolor[HTML]{CFCECE}{\color[HTML]{066F29} \textbf{69.05}}} & \multicolumn{1}{c|}{\cellcolor[HTML]{CFCECE}{\color[HTML]{066F29} \textbf{70.01}}} & \multicolumn{1}{c|}{\cellcolor[HTML]{CFCECE}{\color[HTML]{066F29} \textbf{95.68}}} & {\color[HTML]{066F29} \textbf{21.10}} \\ \hline
\rowcolor[HTML]{FFFFFF} 
\cellcolor[HTML]{FFFFFF}                       & MCLC                     & \multicolumn{1}{c|}{\cellcolor[HTML]{FFFFFF}54.31}                                 & \multicolumn{1}{c|}{\cellcolor[HTML]{FFFFFF}60.35}                                 & \multicolumn{1}{c|}{\cellcolor[HTML]{FFFFFF}88.24}                                 & 125.57                                & \cellcolor[HTML]{FFFFFF}                       & MCLC                     & \multicolumn{1}{c|}{\cellcolor[HTML]{FFFFFF}56.00}                                 & \multicolumn{1}{c|}{\cellcolor[HTML]{FFFFFF}62.09}                                 & \multicolumn{1}{c|}{\cellcolor[HTML]{FFFFFF}90.90}                                 & 93.61                                 \\ \cline{2-6} \cline{8-12} 
\rowcolor[HTML]{CFCECE} 
\multirow{-2}{*}{\cellcolor[HTML]{FFFFFF}ALCL} & PAL                      & \multicolumn{1}{c|}{\cellcolor[HTML]{CFCECE}{\color[HTML]{066F29} \textbf{65.99}}} & \multicolumn{1}{c|}{\cellcolor[HTML]{CFCECE}{\color[HTML]{066F29} \textbf{70.59}}} & \multicolumn{1}{c|}{\cellcolor[HTML]{CFCECE}{\color[HTML]{066F29} \textbf{95.22}}} & {\color[HTML]{066F29} \textbf{20.81}} & \multirow{-2}{*}{\cellcolor[HTML]{FFFFFF}MSDA} & PAL                      & \multicolumn{1}{c|}{\cellcolor[HTML]{CFCECE}{\color[HTML]{066F29} \textbf{69.21}}} & \multicolumn{1}{c|}{\cellcolor[HTML]{CFCECE}{\color[HTML]{066F29} \textbf{72.40}}} & \multicolumn{1}{c|}{\cellcolor[HTML]{CFCECE}{\color[HTML]{066F29} \textbf{97.01}}} & {\color[HTML]{066F29} \textbf{15.70}} \\ \hline
\end{tabular}
}}
\end{table}

\begin{table}[]
\caption{Performance comparison of the PAL and LELCM on three individual datasets with coarse point.}
\vspace{-6pt}
\captionsetup{skip=4pt}
\label{tab:tab-lelcm-coarse}
\centering
\renewcommand{\arraystretch}{1.1} 
\resizebox{1.0\columnwidth}{!}{
\setlength{\tabcolsep}{1mm}{
\begin{tabular}{c|c|ccc|ccc|ccc}
\hline
                      &                             & \multicolumn{3}{c|}{NUAA-SIRST}                                                                                                                                                                                                         & \multicolumn{3}{c|}{NUDT-SIRST}                                                                                                                                                                                                         & \multicolumn{3}{c}{IRSTD-1K}                                                                                                                                                                                    \\ \cline{3-11} 
\multirow{-2}{*}{Net} & \multirow{-2}{*}{Method}    & \multicolumn{1}{c|}{$IoU$}                                                           & \multicolumn{1}{c|}{$P_d$}                                                            & $F_a$                                                            & \multicolumn{1}{c|}{$IoU$}                                                           & \multicolumn{1}{c|}{$P_d$}                                                            & $F_a$                                                            & \multicolumn{1}{c|}{$IoU$}                                                           & \multicolumn{1}{c|}{$P_d$}                                                            & $F_a$                                    \\ \hline
                      & LELCM                       & \multicolumn{1}{c|}{52.11}                                                         & \multicolumn{1}{c|}{{\color[HTML]{066F29} \textbf{91.61}}}                         & 51.36                                                         & \multicolumn{1}{c|}{50.35}                                                         & \multicolumn{1}{c|}{89.27}                                                         & 46.64                                                         & \multicolumn{1}{c|}{50.02}                                                         & \multicolumn{1}{c|}{87.33}                                                         & {\color[HTML]{066F29} \textbf{19.96}} \\ \cline{2-11} 
\multirow{-2}{*}{UIU} & \cellcolor[HTML]{CFCECE}PAL & \multicolumn{1}{c|}{\cellcolor[HTML]{CFCECE}{\color[HTML]{066F29} \textbf{62.25}}} & \multicolumn{1}{c|}{\cellcolor[HTML]{CFCECE}91.25}                                 & \cellcolor[HTML]{CFCECE}{\color[HTML]{066F29} \textbf{34.99}} & \multicolumn{1}{c|}{\cellcolor[HTML]{CFCECE}{\color[HTML]{066F29} \textbf{74.89}}} & \multicolumn{1}{c|}{\cellcolor[HTML]{CFCECE}{\color[HTML]{066F29} \textbf{98.62}}} & \cellcolor[HTML]{CFCECE}{\color[HTML]{066F29} \textbf{7.10}}  & \multicolumn{1}{c|}{\cellcolor[HTML]{CFCECE}{\color[HTML]{066F29} \textbf{61.70}}} & \multicolumn{1}{c|}{\cellcolor[HTML]{CFCECE}{\color[HTML]{066F29} \textbf{92.57}}} & \cellcolor[HTML]{CFCECE}21.60         \\ \hline
                      & LELCM                       & \multicolumn{1}{c|}{53.98}                                                         & \multicolumn{1}{c|}{88.39}                                                         & 36.67                                                         & \multicolumn{1}{c|}{56.56}                                                         & \multicolumn{1}{c|}{91.96}                                                         & 23.31                                                         & \multicolumn{1}{c|}{58.04}                                                         & \multicolumn{1}{c|}{87.78}                                                         & {\color[HTML]{066F29} \textbf{24.36}} \\ \cline{2-11} 
\multirow{-2}{*}{DNA} & \cellcolor[HTML]{CFCECE}PAL & \multicolumn{1}{c|}{\cellcolor[HTML]{CFCECE}{\color[HTML]{066F29} \textbf{66.57}}} & \multicolumn{1}{c|}{\cellcolor[HTML]{CFCECE}{\color[HTML]{066F29} \textbf{90.49}}} & \cellcolor[HTML]{CFCECE}{\color[HTML]{066F29} \textbf{27.44}} & \multicolumn{1}{c|}{\cellcolor[HTML]{CFCECE}{\color[HTML]{066F29} \textbf{73.29}}} & \multicolumn{1}{c|}{\cellcolor[HTML]{CFCECE}{\color[HTML]{066F29} \textbf{98.10}}} & \cellcolor[HTML]{CFCECE}{\color[HTML]{066F29} \textbf{22.08}} & \multicolumn{1}{c|}{\cellcolor[HTML]{CFCECE}{\color[HTML]{066F29} \textbf{58.40}}} & \multicolumn{1}{c|}{\cellcolor[HTML]{CFCECE}{\color[HTML]{066F29} \textbf{91.25}}} & \cellcolor[HTML]{CFCECE}24.52         \\ \hline
\end{tabular}
}}
\end{table}

\begin{table}[]
\caption{Performance comparison of the PAL and LELCM on three individual datasets with centroid point.}
\vspace{-6pt}
\captionsetup{skip=4pt}
\label{tab:tab-lelcm-centroid}
\centering
\renewcommand{\arraystretch}{1.1} 
\resizebox{1.0\columnwidth}{!}{
\setlength{\tabcolsep}{1mm}{
\begin{tabular}{c|c|ccc|ccc|ccc}
\hline
                      &                             & \multicolumn{3}{c|}{NUAA-SIRST}                                                                                                                                                                                                         & \multicolumn{3}{c|}{NUDT-SIRST}                                                                                                                                                                                                        & \multicolumn{3}{c}{IRSTD-1K}                                                                                                                                                                                                            \\ \cline{3-11} 
\multirow{-2}{*}{Net} & \multirow{-2}{*}{Method}    & \multicolumn{1}{c|}{$IoU$}                                                           & \multicolumn{1}{c|}{$P_d$}                                                            & $F_a$                                                            & \multicolumn{1}{c|}{$IoU$}                                                           & \multicolumn{1}{c|}{$P_d$}                                                            & $F_a$                                                           & \multicolumn{1}{c|}{$IoU$}                                                           & \multicolumn{1}{c|}{$P_d$}                                                            & $F_a$                                                            \\ \hline
                      & LELCM                       & \multicolumn{1}{c|}{53.21}                                                         & \multicolumn{1}{c|} {89.24}                         & 59.68                                                         & \multicolumn{1}{c|}{54.84}                                                         & \multicolumn{1}{c|}{90.27}                                                         & 59.21                                                        & \multicolumn{1}{c|}{50.68}                                                         & \multicolumn{1}{c|}{90.47}                                                         & {43.92}                         \\ \cline{2-11} 
\multirow{-2}{*}{UIU} & \cellcolor[HTML]{CFCECE}PAL & \multicolumn{1}{c|}{\cellcolor[HTML]{CFCECE}{\color[HTML]{066F29} \textbf{66.43}}} & \multicolumn{1}{c|}{\cellcolor[HTML]{CFCECE}{\color[HTML]{066F29} \textbf{96.20}}} & \cellcolor[HTML]{CFCECE}{\color[HTML]{066F29} \textbf{14.54}} & \multicolumn{1}{c|}{\cellcolor[HTML]{CFCECE}{\color[HTML]{066F29} \textbf{74.58}}} & \multicolumn{1}{c|}{\cellcolor[HTML]{CFCECE}{\color[HTML]{066F29} \textbf{98.20}}} & \cellcolor[HTML]{CFCECE}{\color[HTML]{066F29} \textbf{4.67}} & \multicolumn{1}{c|}{\cellcolor[HTML]{CFCECE}{\color[HTML]{066F29} \textbf{61.30}}} & \multicolumn{1}{c|}{\cellcolor[HTML]{CFCECE}{\color[HTML]{066F29} \textbf{91.25}}} & \cellcolor[HTML]{CFCECE}{\color[HTML]{066F29} \textbf{36.61}} \\ \hline
                      & LELCM                       & \multicolumn{1}{c|}{58.71}                                                         & \multicolumn{1}{c|}{{\color[HTML]{066F29} \textbf{92.26}}}                         & 38.49                                                         & \multicolumn{1}{c|}{58.30}                                                         & \multicolumn{1}{c|}{89.46}                                                         & 22.43                                                        & \multicolumn{1}{c|}{55.23}                                                         & \multicolumn{1}{c|}{87.15}                                                         & {22.67}                         \\ \cline{2-11} 
\multirow{-2}{*}{DNA} & \cellcolor[HTML]{CFCECE}PAL & \multicolumn{1}{c|}{\cellcolor[HTML]{CFCECE}{\color[HTML]{066F29} \textbf{66.16}}} & \multicolumn{1}{c|}{\cellcolor[HTML]{CFCECE}91.63} & \cellcolor[HTML]{CFCECE}{\color[HTML]{066F29} \textbf{24.42}} & \multicolumn{1}{c|}{\cellcolor[HTML]{CFCECE}{\color[HTML]{066F29} \textbf{73.24}}} & \multicolumn{1}{c|}{\cellcolor[HTML]{CFCECE}{\color[HTML]{066F29} \textbf{97.99}}} & \cellcolor[HTML]{CFCECE}{\color[HTML]{066F29} \textbf{9.10}} & \multicolumn{1}{c|}{\cellcolor[HTML]{CFCECE}{\color[HTML]{066F29} \textbf{59.53}}} & \multicolumn{1}{c|}{\cellcolor[HTML]{CFCECE}{\color[HTML]{066F29} \textbf{88.89}}} & \cellcolor[HTML]{CFCECE}{\color[HTML]{066F29} \textbf{20.14}} \\ \hline
\end{tabular}
}}
\end{table}

\section{The EEDM Loss}
\label{sec:the eedm loss}
For the SIRST detection task, the lack of intrinsic features makes it difficult to accurately locate the target area~\cite{yu2022pay,zhao2025multi}. Therefore, we introduce an edge-enhanced difficulty-mining (EEDM) loss~\cite{zhao2025towards} to constrain the network optimization. The EEDM loss consists of two parts: edge pixel enhancement and difficult pixel mining. Taking a single image as an example, firstly, we obtain the target edge contour in the binary pseudo-label. Secondly, the binary cross-entropy loss is used to obtain the loss value of each pixel and form a loss matrix. Finally, we weight the edge contours extracted from the labels and apply this weighting matrix to the calculated loss matrix to obtain the loss value of each point after edge weighting. The expressions are
\begin{gather}
{L_{ij}} = {W_{ij}}(-({T_{ij}}\log({P_{ij}})+(1 - {T_{ij}})\log(1 - {P_{ij}}))) \\
{W_{ij}} = \alpha  \cdot {E_{ij}} + (1 - {E_{ij}})
\end{gather}
where $E_{ij}$ denotes the edge extracted from the binary pseudo-label, edge pixels are marked as 1 and non-edge pixels are marked as 0. $P_{ij}$ denotes the prediction result. $T$ denotes the pseudo-label after binarization. $\alpha$ is the edge weighting coefficient, which is set to 4~\cite{zhao2025towards}.

Subsequently, difficult pixel mining is performed. Firstly, the loss values of each point are sorted. Secondly, the set of loss values that are greater than or equal to the median is obtained. Finally, the final loss is obtained by calculating the mean loss of difficult pixels. The expressions are as follows:
\begin{gather}
{L_{EEDM}} = \frac{1}{{\left| S \right|}}\sum\limits_{(i,j \in S)} {{L_{ij}}}  \\
S = \{ (i,j)|{L_{ij}} \ge median({L_{ij}})\}
\end{gather}
where $L_{EEDM}$ is the output, $S$ is the difficult pixel set.

On the one hand, EEDM loss promotes the network to increase its sensitivity to target edges by using edge information as an additional constraint. On the other hand, it uses difficult pixel mining  to help the model better focus on difficult-to-detect target areas, thereby preventing small targets from being submerged by the background.

\begin{figure}[t]
  \captionsetup{skip=2pt}
  \centering
   \includegraphics[width=\columnwidth]{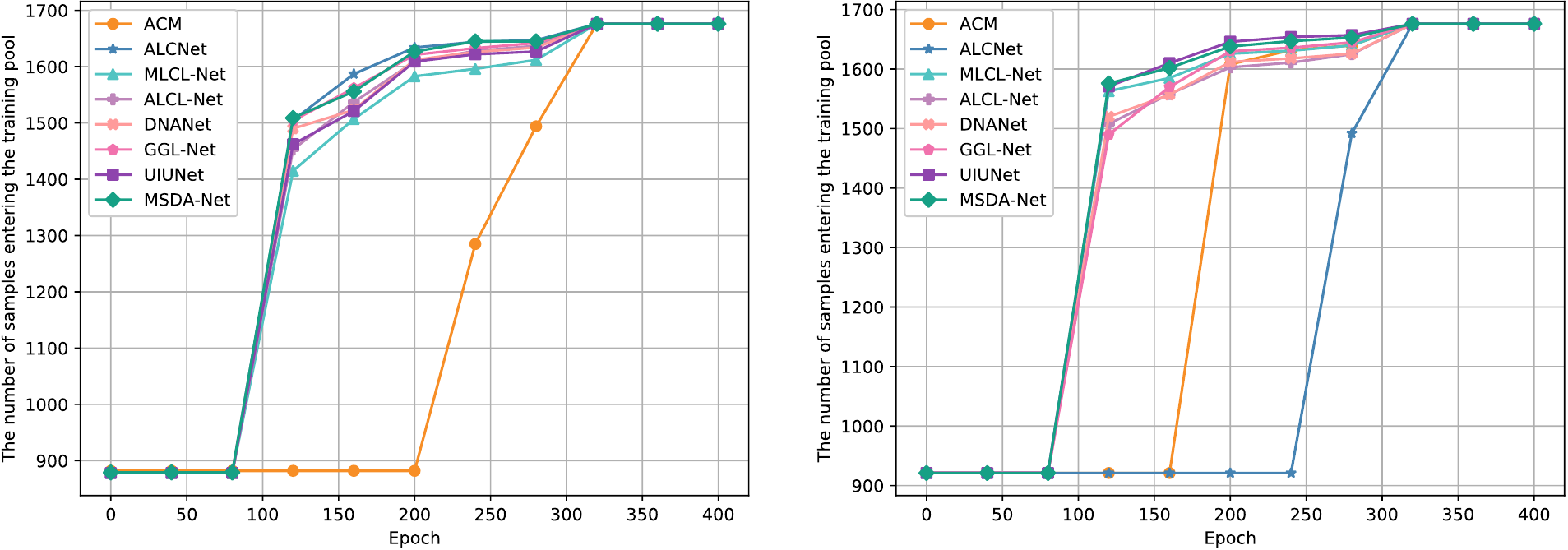}
   \caption{Training pool sample count on the SIRST3 dataset every 40 epochs. Left: coarse point. Right: centroid point.}
   \label{fig:fig-curve}
   \vspace{-6pt}
\end{figure}

\begin{figure}[t]
  \captionsetup{skip=2pt}
  \centering
   \includegraphics[width=\columnwidth]{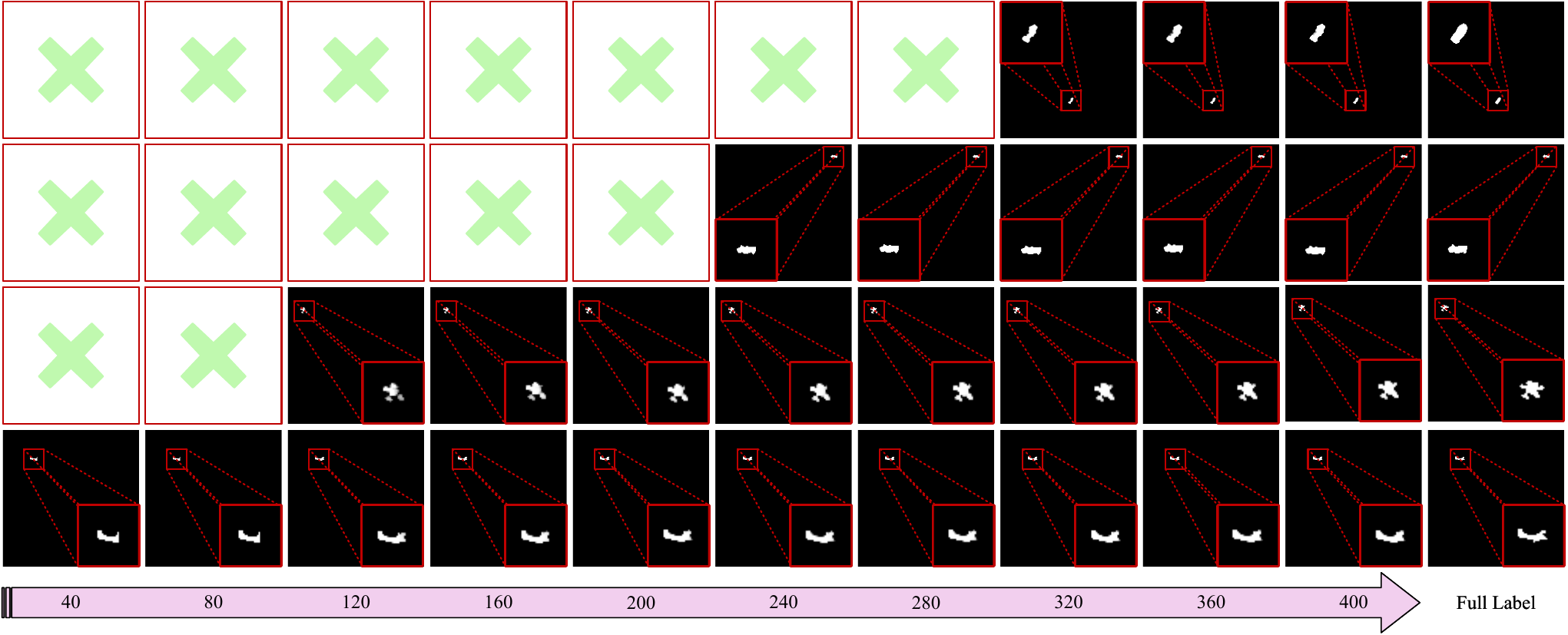}
   \caption{Pseudo-label evolution results of MSDA-Net equipped with the PAL framework on the SIRST3 dataset.}
   \label{fig:fig-evolution}
   \vspace{-6pt}
\end{figure}

\begin{table}[]
\caption{Batch size investigation on the SIRST3 dataset. \textit{Coarse} denotes coarse point supervision. \textit{Centroid} denotes centroid point supervision.}
\vspace{-3pt}
\label{tab:s-tab01}
\setlength{\tabcolsep}{1mm}
\resizebox{\columnwidth}{!}{
\begin{tabular}{c|cccc|cccc}
\hline
& \multicolumn{4}{c|}{MSDA-Net Coarse + PAL}                                                                                                                    & \multicolumn{4}{c}{MSDA-Net Centroid + PAL}                                                                                                                  \\ \cline{2-9} 
\multirow{-2}{*}{\begin{tabular}[c]{@{}c@{}}Batch \\ size\end{tabular}} & $IoU$                                   & $nIoU$                                  & $P_d$                                    & $F_a$                                    & $IoU$                                   & $nIoU$                                  & $P_d$                                    & $F_a$                                   \\ \hline
40                                                                      & 67.33                                 & 70.90                                 & 96.41                                 & 20.27                                 & 67.31                                 & 70.18                                 & 96.48                                 & 28.77                                \\
32                                                                      & 68.03                                 & 71.30                                 & 97.28                                 & 18.67                                 & 68.43                                 & 71.60                                 & 96.08                                 & 14.89                                \\
24                                                                      & {\color[HTML]{066F29} \textbf{70.31}} & 71.53                                 & 95.15                                 & {\color[HTML]{066F29} \textbf{13.78}} & 69.16                                 & 71.98                                 & 95.55                                 & {\color[HTML]{066F29} \textbf{7.82}} \\
\rowcolor[HTML]{CFCECE} 
16                                                                      & 69.38                                 & {\color[HTML]{066F29} \textbf{71.55}} & {\color[HTML]{066F29} \textbf{97.41}} & 16.34                                 & {\color[HTML]{066F29} \textbf{69.21}} & {\color[HTML]{066F29} \textbf{72.40}} & {\color[HTML]{066F29} \textbf{97.01}} & 15.70                                \\
8                                                                       & 68.60                                 & 71.12                                 & 96.61                                 & 24.20                                 & 68.83                                 & 71.84                                 & 96.48                                 & 14.88                                \\ \hline
\end{tabular}
}
\end{table}

\begin{table}[]
\caption{Update period investigation on the SIRST3 dataset. \textit{Coarse} denotes coarse point supervision. \textit{Centroid} denotes centroid point supervision.}
\vspace{-3pt}
\label{tab:s-tab02}
\setlength{\tabcolsep}{1mm}
\resizebox{\columnwidth}{!}{
\begin{tabular}{c|cccc|cccc}
\hline
 & \multicolumn{4}{c|}{MSDA-Net Coarse + PAL}                                                                                                                    & \multicolumn{4}{c}{MSDA-Net Centroid + PAL}                                                                                                                   \\ \cline{2-9} 
\multirow{-2}{*}{\begin{tabular}[c]{@{}c@{}}Update \\ period\end{tabular}} & $IoU$                                   & $nIoU$                                  & $P_d$                                    & $F_a$                                    & $IoU$                                   & $nIoU$                                  & $P_d$                                    & $F_a$                                    \\ \hline
1                                                                          & 66.35                                 & 68.21                                 & 95.88                                 & 29.37                                 & 64.61                                 & 47.27                                 & 82.66                                 & 41.45                                 \\
3                                                                          & {\color[HTML]{066F29} \textbf{70.01}} & 70.65                                 & 96.68                                 & 20.18                                 & {\color[HTML]{066F29} \textbf{69.65}} & 71.51                                 & {\color[HTML]{066F29} \textbf{97.41}} & {\color[HTML]{066F29} \textbf{13.62}} \\
\rowcolor[HTML]{CFCECE} 
5                                                                          & 69.38                                 & 71.55                                 & {\color[HTML]{066F29} \textbf{97.41}} & 16.34                                 & 69.21                                 & {\color[HTML]{066F29} \textbf{72.40}} & 97.01                                 & 15.70                                 \\
7                                                                          & 69.04                                 & 71.79                                 & 96.88                                 & 15.06                                 & 68.53                                 & 72.11                                 & 97.01                                 & 22.07                                 \\
10                                                                         & 68.57                                 & {\color[HTML]{066F29} \textbf{71.80}} & 97.08                                 & {\color[HTML]{066F29} \textbf{14.38}} & 68.33                                 & 71.16                                 & 96.94                                 & 26.89                                 \\ \hline
\end{tabular}
}
\end{table}

\begin{table}[]
\caption{Learning rate investigation on the SIRST3 dataset. \textit{Coarse} denotes coarse point supervision. \textit{Centroid} denotes centroid point supervision.}
\vspace{-3pt}
\label{tab:s-tab03}
\setlength{\tabcolsep}{1mm}
\resizebox{\columnwidth}{!}{
\begin{tabular}{c|cccc|cccc}
\hline
& \multicolumn{4}{c|}{MSDA-Net Coarse + PAL}                                                                                                                    & \multicolumn{4}{c}{MSDA-Net Centroid + PAL}                                                                                                                   \\ \cline{2-9} 
\multirow{-2}{*}{\begin{tabular}[c]{@{}c@{}}Learning \\ rate\end{tabular}} & $IoU$                                   & $nIoU$                                  & $P_d$                                    & $F_a$                                    & $IoU$                                   & $nIoU$                                  & $P_d$                                    & $F_a$                                    \\ \hline
$1e^{\text{\scriptsize -2}}$                                                                       & -                                     & -                                     & -                                     & -                                     & -                                     & -                                     & -                                     & -                                     \\
$5e^{\text{\scriptsize -3}}$                                                                       & {\color[HTML]{066F29} \textbf{69.53}} & {\color[HTML]{066F29} \textbf{71.69}} & {\color[HTML]{066F29} \textbf{97.81}} & 17.13                                 & 68.61                                 & 70.21                                 & 96.88                                 & 19.81                                 \\
\rowcolor[HTML]{CFCECE} 
$1e^{\text{\scriptsize -3}}$                                                                       & 69.38                                 & 71.55                                 & 97.41                                 & 16.34                                 & 69.21                                 & {\color[HTML]{066F29} \textbf{72.40}} & {\color[HTML]{066F29} \textbf{97.01}} & {\color[HTML]{066F29} \textbf{15.70}} \\
$5e^{\text{\scriptsize -4}}$                                                                       & 68.83                                 & 70.71                                 & 95.88                                 & {\color[HTML]{066F29} \textbf{16.23}} & {\color[HTML]{066F29} \textbf{69.24}} & 72.34                                 & 96.35                                 & 23.05                                 \\
$1e^{\text{\scriptsize -4}}$                                                                       & 65.31                                 & 70.87                                 & 94.88                                 & 16.94                                 & 65.41                                 & 70.09                                 & 95.02                                 & 18.07                                 \\
$5e^{\text{\scriptsize -5}}$                                                                       & 59.68                                 & 65.70                                 & 92.09                                 & 19.38                                 & 61.64                                 & 67.33                                 & 94.49                                 & 19.14                                 \\ \hline
\end{tabular}
}
\end{table}

\begin{table}[]
\caption{Missed detection rate threshold investigation on the SIRST3 dataset. \textit{Coarse} denotes coarse point supervision. \textit{Centroid} denotes centroid point supervision.}
\vspace{-3pt}
\label{tab:s-tab04}
\setlength{\tabcolsep}{1mm}
\resizebox{\columnwidth}{!}{
\begin{tabular}{c|cccc|cccc}
\hline
& \multicolumn{4}{c|}{MSDA-Net Coarse + PAL}                                                                                                                                        & \multicolumn{4}{c}{MSDA-Net Centroid + PAL}                                                                                                                   \\ \cline{2-9} 
\multirow{-2}{*}{\textit{$T_{miss}$}} & $IoU$                                   & $nIoU$                                                      & $P_d$                                    & $F_a$                                    & $IoU$                                   & $nIoU$                                  & $P_d$                                    & $F_a$                                    \\ \hline
\rowcolor[HTML]{CFCECE} 
Change                           & {\color[HTML]{066F29} \textbf{69.38}} & 71.55                                                     & {\color[HTML]{066F29} \textbf{97.41}} & 16.34                                 & {\color[HTML]{066F29} \textbf{69.21}} & {\color[HTML]{066F29} \textbf{72.40}} & 97.01                                 & 15.70                                 \\
0.2                              & 69.30                                 & \multicolumn{1}{l}{{\color[HTML]{066F29} \textbf{71.69}}} & 96.88                                 & 19.05                                 & 69.07                                 & 70.89                                 & {\color[HTML]{066F29} \textbf{97.54}} & 16.32                                 \\
0.4                              & 69.14                                 & 71.14                                                     & 96.21                                 & {\color[HTML]{066F29} \textbf{14.00}} & 69.06                                 & 71.38                                 & 96.94                                 & 16.52                                 \\
\rowcolor[HTML]{FFFFFF} 
0.6                              & 68.73                                 & 71.61                                                     & 96.61                                 & 15.82                                 & 68.91                                 & 71.41                                 & 96.41                                 & 17.69                                 \\
0.8                              & 67.41                                 & 71.06                                                     & 96.08                                 & 15.61                                 & 68.42                                 & 71.14                                 & 96.08                                 & {\color[HTML]{066F29} \textbf{15.64}} \\ \hline
\end{tabular}
}
\end{table}

\section{Why ``From Easy to Hard'' Fits this Task?}
\label{sec:why}
For this task, the target regions are usually very small and low-contrast, which makes them highly sensitive to pseudo-label noise. This challenge is further exacerbated with single point supervision, as sparse annotations provide little spatial guidance. Training directly on unreliable pseudo-labels not only leads to degraded performance but also causes semantic drift due to misleading supervision. The “from easy to hard” idea can alleviate this problem, so we systematically introduce it into the SIRST detection with single point supervision for the first time and build a Progressive Active Learning (PAL) framework. \cref{fig:fig-curve} and \cref{fig:fig-evolution} show that our PAL can gradually introduce harder samples and generate more refined pseudo-labels. The adaptive modifications to the characteristics of this task are mainly reflected in the easy-sample pseudo-label generation (EPG) strategy and fine dual-update strategy. The former automatically screens ``easy samples'' based on target characteristics through local brightness and edge information. The latter combines point labels with model feedback to regulate sample introduction and label evolution.

\section{Compared with Other Methods}
\label{sec:compared with other methods}

To further verify the performance of the proposed PAL framework, this section further presents the performance comparison with other methods. On the one hand, we compare with the MCLC (Monte Carlo Linear Clustering) method~\cite{li2023monte}, which is a static pseudo-label generation method. On the other hand, we compare with the LELCM (Label Evolution framework based on Local Contrast Measure) method~\cite{yang2024label}, which is a dynamic pseudo-label evolution method. Notably, the MCLC will additionally use the area information of the true target area to classify the target type (“Point”, “Spot”, “Extended”) when generating pseudo-labels. However, the task setting of this study is that the training set only has point labels. To make a fair comparison and ensure that MCLC can cover all targets in the selected area, all targets in the MCLC experiment are assigned the “Extended” type. In addition, since the code of LELCM is not available, we directly use the results in the paper for comparison.

\textbf{\textit{Comparison with the the MCLC method.}} To fully compare the performance of PAL and MCLC, we conduct comparative experiments on the SIRST3 dataset with different point labels and multiple networks. From \cref{tab:tab-mclc-coarse}, compared with MCLC, using our proposed PAL improves the IoU by 3.34\%-14.67\% and the $P_d$ by 5.38\%-10.96\% on the SIRST3 dataset with coarse point. From \cref{tab:tab-mclc-centroid}, compared with MCLC, using our proposed PAL improves the IoU by 3.64\%-13.21\% and the $P_d$ by 6.11\%-8.31\% on the SIRST3 dataset with centroid point. In summary, compared with MCLC, our PAL framework has significantly better performance.

\begin{table*}[]
\caption{$IoU$ (\%), $nIoU$ (\%), $P_d$ (\%) and $F_a$ ($\times 10^{\text{\scriptsize -6}}$)  values of different methods achieved on the SIRST3 dataset with centroid point labels. \textit{NUAA-SIRST-Test}, \textit{NUDT-SIRST-Test} and \textit{IRSTD-1K-Test} denote the decompositions of \textit{SIRST3-Test} to verify the robustness of the model. \textit{DLN Centroid} denotes DLN-based methods under centroid point supervision.}
\label{tab:s-tab05}
\setlength{\tabcolsep}{1mm}
\renewcommand{\arraystretch}{1.1}
\resizebox{\textwidth}{!}{
\begin{tabular}{c|c|cccc|cccc|cccc|cccc}
\hline
                                    &                                                   & \multicolumn{4}{c|}{SIRST3-Test}                                                                                                                                                                                                                              & \multicolumn{4}{c|}{NUAA-SIRST-Test}                                                                                                                                                                                                                          & \multicolumn{4}{c|}{NUDT-SIRST-Test}                                                                                                                                                                                                                          & \multicolumn{4}{c}{IRSTD-1K-Test}                                                                                                                                                                                                                             \\
\hline
\end{tabular}
}
\end{table*}

\textbf{\textit{Comparison with the the LELCM method.}} We further compare the performance of PAL and LELCM on three individual datasets with different point labels. The experimental results are shown in \cref{tab:tab-lelcm-coarse} and \cref{tab:tab-lelcm-centroid}. From the results on three individual datasets with coarse point, compared with LELCM, using the proposed PAL improves the IoU by an average of 12.67\% and the $P_d$ by an average of 4.32\%. From the results on three individual datasets with centroid point, compared with LELCM, using the proposed PAL improves the IoU by an average of 11.71\% and the $P_d$ by an average of 4.22\%. In summary, compared with LELCM, our PAL framework has significantly better performance.

\section{More Ablation Experiments}
\label{sec:more ablation experiments}
In this section, we study more influencing factors in detail, including the batch size, update period of the refined dual-update strategy, learning rate, and missed detection rate threshold.

\textbf{\textit{1) Batch size.}} To explore the impact of batch size on the performance of the final generated model, we explore the PAL framework with different batch size settings. From \cref{tab:s-tab01}, when the batch size is set too large or too small, the final generated model will experience a slight performance degradation. Specifically, when the batch size is set too large, the number of model updates will be reduced and each update will be based on a large number of samples, making the gradient update smoother and reducing the randomness of the model, which makes the model more likely to overfit on the training data. When the batch size is set too small, each gradient update is based only on a small number of data samples, resulting in a large variance in the gradient estimate and large fluctuations in the gradient direction, which makes it easy to fall into a local optimum. On the whole, the final model with different batch size settings has relatively stable results, which verifies the stability of the proposed PAL framework. Based on the results in \cref{tab:s-tab01}, the batch size is uniformly set to 16 in the experiments.

\begin{table*}[]
\caption{$IoU$ (\%), $nIoU$ (\%), $P_d$ (\%) and $F_a$ ($\times 10^{\text{\scriptsize -6}}$)  values of different methods achieved on the separate NUAA-SIRST, NUDT-SIRST, and IRSTD-1k datasets with centroid point labels. \textit{(213:214)}, \textit{(663:664)} and \textit{(800:201)} denote the division of training samples and test samples. \textit{DLN Centroid} denotes DLN-based methods under centroid point supervision.}
\label{tab:s-tab06}
\setlength{\tabcolsep}{2mm}
\renewcommand{\arraystretch}{1.1}
\resizebox{\textwidth}{!}{
\begin{tabular}{c|c|cccc|cccc|cccc}
\hline
                                    &                                                   & \multicolumn{4}{c|}{NUAA-SIRST (213:214)}                                                                                                                                                                                                                     & \multicolumn{4}{c|}{NUDT-SIRST (663:664)}                                                                                                                                                                                                                     & \multicolumn{4}{c}{IRSTD-1K (800:201)}                                                                                                                                                                                                                        \\ \cline{3-14} 
\multirow{-2}{*}{Scheme}            & \multirow{-2}{*}{Description}                     & $IoU$                                                           & $nIoU$                                                          & $P_d$                                                            & $F_a$                                                            & $IoU$                                                           & $nIoU$                                                          & $P_d$                                                            & $F_a$                                                           & $IoU$                                                           & $nIoU$                                                          & $P_d$                                                            & $F_a$                                                            \\ \hline
ACM                        & DLN Full                                          & 65.67                                                         & 63.74                                                         & 90.11                                                         & 24.01                                                         & 65.33                                                         & 65.12                                                         & 95.87                                                         & 12.92                                                         & 60.45                                                         & 53.70                                                         & 92.26                                                         & 46.06                                                         \\ \hline
ALCNet                     & DLN Full                                          & 66.41                                                         & 65.18                                                         & 91.63                                                         & 35.26                                                         & 69.74                                                         & 70.67                                                         & 97.46                                                         & 11.15                                                         & 62.47                                                         & 55.25                                                         & 88.89                                                         & 36.79                                                         \\ \hline
                                    & DLN Full                                          & 74.68                                                         & 76.50                                                         & 95.82                                                         & 28.74                                                         & 94.03                                                         & 93.97                                                         & 98.73                                                         & 7.72                                                          & 64.86                                                         & 63.35                                                         & 91.25                                                         & 23.23                                                         \\
                                    & \cellcolor[HTML]{CFCECE}DLN Centroid + LESPS      & \cellcolor[HTML]{CFCECE}32.69                                 & \cellcolor[HTML]{CFCECE}32.59                                 & \cellcolor[HTML]{CFCECE}82.89                                 & \cellcolor[HTML]{CFCECE}{\color[HTML]{066F29} \textbf{20.85}} & \cellcolor[HTML]{CFCECE}34.11                                 & \cellcolor[HTML]{CFCECE}32.00                                 & \cellcolor[HTML]{CFCECE}89.52                                 & \cellcolor[HTML]{CFCECE}46.17                                 & \cellcolor[HTML]{CFCECE}46.59                                 & \cellcolor[HTML]{CFCECE}45.16                                 & \cellcolor[HTML]{CFCECE}86.87                                 & \cellcolor[HTML]{CFCECE}29.42                                 \\
\multirow{-3}{*}{MLCL-Net} & \cellcolor[HTML]{CFCECE}DLN Centroid + PAL (Ours) & \cellcolor[HTML]{CFCECE}{\color[HTML]{066F29} \textbf{67.64}} & \cellcolor[HTML]{CFCECE}{\color[HTML]{066F29} \textbf{70.28}} & \cellcolor[HTML]{CFCECE}{\color[HTML]{066F29} \textbf{93.92}} & \cellcolor[HTML]{CFCECE}45.00                                 & \cellcolor[HTML]{CFCECE}{\color[HTML]{066F29} \textbf{72.67}} & \cellcolor[HTML]{CFCECE}{\color[HTML]{066F29} \textbf{73.87}} & \cellcolor[HTML]{CFCECE}{\color[HTML]{066F29} \textbf{98.31}} & \cellcolor[HTML]{CFCECE}{\color[HTML]{066F29} \textbf{7.79}}  & \cellcolor[HTML]{CFCECE}{\color[HTML]{066F29} \textbf{59.14}} & \cellcolor[HTML]{CFCECE}{\color[HTML]{066F29} \textbf{59.21}} & \cellcolor[HTML]{CFCECE}{\color[HTML]{066F29} \textbf{91.25}} & \cellcolor[HTML]{CFCECE}{\color[HTML]{066F29} \textbf{25.79}} \\ \hline
                                    & DLN Full                                          & 72.22                                                         & 72.64                                                         & 94.68                                                         & 35.06                                                         & 92.80                                                         & 93.01                                                         & 99.05                                                         & 2.21                                                          & 65.56                                                         & 65.03                                                         & 91.58                                                         & 9.68                                                          \\
                                    & \cellcolor[HTML]{CFCECE}DLN Centroid + LESPS      & \cellcolor[HTML]{CFCECE}35.56                                 & \cellcolor[HTML]{CFCECE}32.99                                 & \cellcolor[HTML]{CFCECE}92.02                                 & \cellcolor[HTML]{CFCECE}{\color[HTML]{066F29} \textbf{32.45}} & \cellcolor[HTML]{CFCECE}46.08                                 & \cellcolor[HTML]{CFCECE}43.19                                 & \cellcolor[HTML]{CFCECE}86.88                                 & \cellcolor[HTML]{CFCECE}41.00                                 & \cellcolor[HTML]{CFCECE}45.77                                 & \cellcolor[HTML]{CFCECE}42.80                                 & \cellcolor[HTML]{CFCECE}86.53                                 & \cellcolor[HTML]{CFCECE}{\color[HTML]{066F29} \textbf{20.46}} \\
\multirow{-3}{*}{ALCL-Net}  & \cellcolor[HTML]{CFCECE}DLN Centroid + PAL (Ours) & \cellcolor[HTML]{CFCECE}{\color[HTML]{066F29} \textbf{65.10}} & \cellcolor[HTML]{CFCECE}{\color[HTML]{066F29} \textbf{66.41}} & \cellcolor[HTML]{CFCECE}{\color[HTML]{066F29} \textbf{93.16}} & \cellcolor[HTML]{CFCECE}44.11                                 & \cellcolor[HTML]{CFCECE}{\color[HTML]{066F29} \textbf{71.55}} & \cellcolor[HTML]{CFCECE}{\color[HTML]{066F29} \textbf{72.55}} & \cellcolor[HTML]{CFCECE}{\color[HTML]{066F29} \textbf{97.25}} & \cellcolor[HTML]{CFCECE}{\color[HTML]{066F29} \textbf{9.81}}  & \cellcolor[HTML]{CFCECE}{\color[HTML]{066F29} \textbf{59.82}} & \cellcolor[HTML]{CFCECE}{\color[HTML]{066F29} \textbf{53.92}} & \cellcolor[HTML]{CFCECE}{\color[HTML]{066F29} \textbf{87.54}} & \cellcolor[HTML]{CFCECE}22.38                                 \\ \hline
                                    & DLN Full                                          & 76.40                                                         & 78.32                                                         & 96.20                                                         & 20.72                                                         & 95.17                                                         & 95.19                                                         & 98.94                                                         & 2.00                                                          & 69.06                                                         & 65.22                                                         & 91.58                                                         & 11.56                                                         \\
                                    & \cellcolor[HTML]{CFCECE}DLN Centroid + LESPS      & \cellcolor[HTML]{CFCECE}16.89                                 & \cellcolor[HTML]{CFCECE}19.50                                 & \cellcolor[HTML]{CFCECE}61.98                                 & \cellcolor[HTML]{CFCECE}45.14                                 & \cellcolor[HTML]{CFCECE}39.39                                 & \cellcolor[HTML]{CFCECE}45.53                                 & \cellcolor[HTML]{CFCECE}86.14                                 & \cellcolor[HTML]{CFCECE}{\color[HTML]{C00000} 291.02}         & \cellcolor[HTML]{CFCECE}50.14                                 & \cellcolor[HTML]{CFCECE}49.95                                 & \cellcolor[HTML]{CFCECE}87.54                                 & \cellcolor[HTML]{CFCECE}{\color[HTML]{066F29} \textbf{16.13}} \\
\multirow{-3}{*}{DNANet}    & \cellcolor[HTML]{CFCECE}DLN Centroid + PAL (Ours) & \cellcolor[HTML]{CFCECE}{\color[HTML]{066F29} \textbf{66.16}} & \cellcolor[HTML]{CFCECE}{\color[HTML]{066F29} \textbf{66.68}} & \cellcolor[HTML]{CFCECE}{\color[HTML]{066F29} \textbf{91.63}} & \cellcolor[HTML]{CFCECE}{\color[HTML]{066F29} \textbf{24.42}} & \cellcolor[HTML]{CFCECE}{\color[HTML]{066F29} \textbf{73.24}} & \cellcolor[HTML]{CFCECE}{\color[HTML]{066F29} \textbf{74.41}} & \cellcolor[HTML]{CFCECE}{\color[HTML]{066F29} \textbf{97.99}} & \cellcolor[HTML]{CFCECE}{\color[HTML]{066F29} \textbf{9.10}}  & \cellcolor[HTML]{CFCECE}{\color[HTML]{066F29} \textbf{59.53}} & \cellcolor[HTML]{CFCECE}{\color[HTML]{066F29} \textbf{57.62}} & \cellcolor[HTML]{CFCECE}{\color[HTML]{066F29} \textbf{88.89}} & \cellcolor[HTML]{CFCECE}20.14                                 \\ \hline
                                    & DLN Full                                          & 75.47                                                         & 75.93                                                         & 97.34                                                         & 20.65                                                         & 94.86                                                         & 94.93                                                         & 99.47                                                         & 1.03                                                          & 69.09                                                         & 65.52                                                         & 92.59                                                         & 13.68                                                         \\
                                    & \cellcolor[HTML]{CFCECE}DLN Centroid + LESPS      & \cellcolor[HTML]{CFCECE}54.85                                 & \cellcolor[HTML]{CFCECE}53.82                                 & \cellcolor[HTML]{CFCECE}90.49                                 & \cellcolor[HTML]{CFCECE}21.20                                 & \cellcolor[HTML]{CFCECE}48.23                                 & \cellcolor[HTML]{CFCECE}46.86                                 & \cellcolor[HTML]{CFCECE}89.95                                 & \cellcolor[HTML]{CFCECE}50.23                                 & \cellcolor[HTML]{CFCECE}48.82                                 & \cellcolor[HTML]{CFCECE}42.67                                 & \cellcolor[HTML]{CFCECE}81.48                                 & \cellcolor[HTML]{CFCECE}{\color[HTML]{066F29} \textbf{19.66}} \\
\multirow{-3}{*}{GGL-Net}  & \cellcolor[HTML]{CFCECE}DLN Centroid + PAL (Ours) & \cellcolor[HTML]{CFCECE}{\color[HTML]{066F29} \textbf{64.79}} & \cellcolor[HTML]{CFCECE}{\color[HTML]{066F29} \textbf{65.05}} & \cellcolor[HTML]{CFCECE}{\color[HTML]{066F29} \textbf{94.68}} & \cellcolor[HTML]{CFCECE}{\color[HTML]{066F29} \textbf{20.51}} & \cellcolor[HTML]{CFCECE}{\color[HTML]{066F29} \textbf{73.42}} & \cellcolor[HTML]{CFCECE}{\color[HTML]{066F29} \textbf{74.71}} & \cellcolor[HTML]{CFCECE}{\color[HTML]{066F29} \textbf{98.41}} & \cellcolor[HTML]{CFCECE}{\color[HTML]{066F29} \textbf{5.68}}  & \cellcolor[HTML]{CFCECE}{\color[HTML]{066F29} \textbf{61.73}} & \cellcolor[HTML]{CFCECE}{\color[HTML]{066F29} \textbf{54.95}} & \cellcolor[HTML]{CFCECE}{\color[HTML]{066F29} \textbf{82.49}} & \cellcolor[HTML]{CFCECE}21.22                                 \\ \hline
                                    & DLN Full                                          & 78.02                                                         & 76.85                                                         & 96.58                                                         & 14.68                                                         & 95.07                                                         & 95.10                                                         & 98.73                                                         & 0.21                                                          & 70.94                                                         & 64.32                                                         & 91.25                                                         & 10.57                                                         \\
                                    & \cellcolor[HTML]{CFCECE}DLN Centroid + LESPS      & \cellcolor[HTML]{CFCECE}26.05                                 & \cellcolor[HTML]{CFCECE}25.27                                 & \cellcolor[HTML]{CFCECE}54.75                                 & \cellcolor[HTML]{CFCECE}37.52                                 & \cellcolor[HTML]{CFCECE}41.60                                 & \cellcolor[HTML]{CFCECE}39.43                                 & \cellcolor[HTML]{CFCECE}85.93                                 & \cellcolor[HTML]{CFCECE}68.23                                 & \cellcolor[HTML]{CFCECE}41.40                                 & \cellcolor[HTML]{CFCECE}40.42                                 & \cellcolor[HTML]{CFCECE}87.88                                 & \cellcolor[HTML]{CFCECE}84.93                                 \\
\multirow{-3}{*}{UIUNet}    & \cellcolor[HTML]{CFCECE}DLN Centroid + PAL (Ours) & \cellcolor[HTML]{CFCECE}{\color[HTML]{066F29} \textbf{66.43}} & \cellcolor[HTML]{CFCECE}{\color[HTML]{066F29} \textbf{69.49}} & \cellcolor[HTML]{CFCECE}{\color[HTML]{066F29} \textbf{96.20}} & \cellcolor[HTML]{CFCECE}{\color[HTML]{066F29} \textbf{14.54}} & \cellcolor[HTML]{CFCECE}{\color[HTML]{066F29} \textbf{74.58}} & \cellcolor[HTML]{CFCECE}{\color[HTML]{066F29} \textbf{75.43}} & \cellcolor[HTML]{CFCECE}{\color[HTML]{066F29} \textbf{98.20}} & \cellcolor[HTML]{CFCECE}{\color[HTML]{066F29} \textbf{4.67}}  & \cellcolor[HTML]{CFCECE}{\color[HTML]{066F29} \textbf{61.30}} & \cellcolor[HTML]{CFCECE}{\color[HTML]{066F29} \textbf{55.37}} & \cellcolor[HTML]{CFCECE}{\color[HTML]{066F29} \textbf{91.25}} & \cellcolor[HTML]{CFCECE}{\color[HTML]{066F29} \textbf{36.61}} \\ \hline
                                    & DLN Full                                          & 76.73                                                         & 77.78                                                         & 96.20                                                         & 21.75                                                         & 95.27                                                         & 95.18                                                         & 99.15                                                         & 1.72                                                          & 70.98                                                         & 65.70                                                         & 93.94                                                         & 33.95                                                         \\
                                    & \cellcolor[HTML]{CFCECE}DLN Centroid + LESPS      & \cellcolor[HTML]{CFCECE}48.63                                 & \cellcolor[HTML]{CFCECE}48.70                                 & \cellcolor[HTML]{CFCECE}87.45                                 & \cellcolor[HTML]{CFCECE}37.59                                 & \cellcolor[HTML]{CFCECE}32.90                                 & \cellcolor[HTML]{CFCECE}30.59                                 & \cellcolor[HTML]{CFCECE}83.92                                 & \cellcolor[HTML]{CFCECE}{\color[HTML]{C00000} 112.97}         & \cellcolor[HTML]{CFCECE}48.67                                 & \cellcolor[HTML]{CFCECE}46.41                                 & \cellcolor[HTML]{CFCECE}85.86                                 & \cellcolor[HTML]{CFCECE}26.72                                 \\
\multirow{-3}{*}{MSDA-Net}  & \cellcolor[HTML]{CFCECE}DLN Centroid + PAL (Ours) & \cellcolor[HTML]{CFCECE}{\color[HTML]{066F29} \textbf{64.56}} & \cellcolor[HTML]{CFCECE}{\color[HTML]{066F29} \textbf{66.42}} & \cellcolor[HTML]{CFCECE}{\color[HTML]{066F29} \textbf{91.25}} & \cellcolor[HTML]{CFCECE}{\color[HTML]{066F29} \textbf{32.04}} & \cellcolor[HTML]{CFCECE}{\color[HTML]{066F29} \textbf{73.45}} & \cellcolor[HTML]{CFCECE}{\color[HTML]{066F29} \textbf{74.65}} & \cellcolor[HTML]{CFCECE}{\color[HTML]{066F29} \textbf{98.52}} & \cellcolor[HTML]{CFCECE}{\color[HTML]{066F29} \textbf{11.84}} & \cellcolor[HTML]{CFCECE}{\color[HTML]{066F29} \textbf{65.33}} & \cellcolor[HTML]{CFCECE}{\color[HTML]{066F29} \textbf{60.18}} & \cellcolor[HTML]{CFCECE}{\color[HTML]{066F29} \textbf{92.93}} & \cellcolor[HTML]{CFCECE}{\color[HTML]{066F29} \textbf{23.50}} \\ \hline
\end{tabular}
}
\end{table*}

\textbf{\textit{2) Update period of the refined dual-update strategy.}} To explore the impact of the update period of the refined dual-update strategy on the performance of the final generated model, we explore the PAL framework with different update period settings. The experimental results are shown in \cref{tab:s-tab02}. Except that the update period is set to 1, the other settings have relatively stable results, which illustrates the robustness of the proposed PAL framework. The significant decrease in performance when the update period is set to 1 is because hard samples need to be trained for appropriate epochs after entering the training pool so that the model can fully learn the newly entered hard samples. It is just like when students face difficult content, they need to spend a certain amount of time to recognize, understand and apply it flexibly. If the time given is too short, students will not be able to deeply understand the knowledge, which will lead to a certain degree of knowledge confusion. In addition, the smaller the update period is set, the more time it takes to train. In the experiments, the update period is set to 5.

\textbf{\textit{3) Learning rate.}} To explore the impact of the learning rate on the performance of the final generative model, we explore the PAL framework with different learning rate settings. As shown in \cref{tab:s-tab03}, the results are consistent with the relationship between the learning rate settings and performance changes in general deep learning networks. When the learning rate is set too high ($1e^{\text{\scriptsize -2}}$), the update step size of the model will become larger, resulting in unstable parameter updates during training, which in turn leads to training collapse. When the learning rate is set too low ($5e^{\text{\scriptsize -5}}$), the update step size of the model becomes smaller, which leads to slow network optimization and easy to fall into local optimality. In addition, when the learning rate is set to $5e^{\text{\scriptsize -5}}-5e^{\text{\scriptsize -3}}$, the results of the final generated model are stable. This verifies the robustness of the proposed PAL framework. In the experiment, we set the learning rate uniformly to $1e^{\text{\scriptsize -3}}$.

\definecolor{yellow}{RGB}{231,111,13}  
\definecolor{red}{RGB}{255,0,0}   
\definecolor{blue}{RGB}{0,112,192} 

\begin{figure*}[t]
    \centering
    \includegraphics[width=\textwidth]{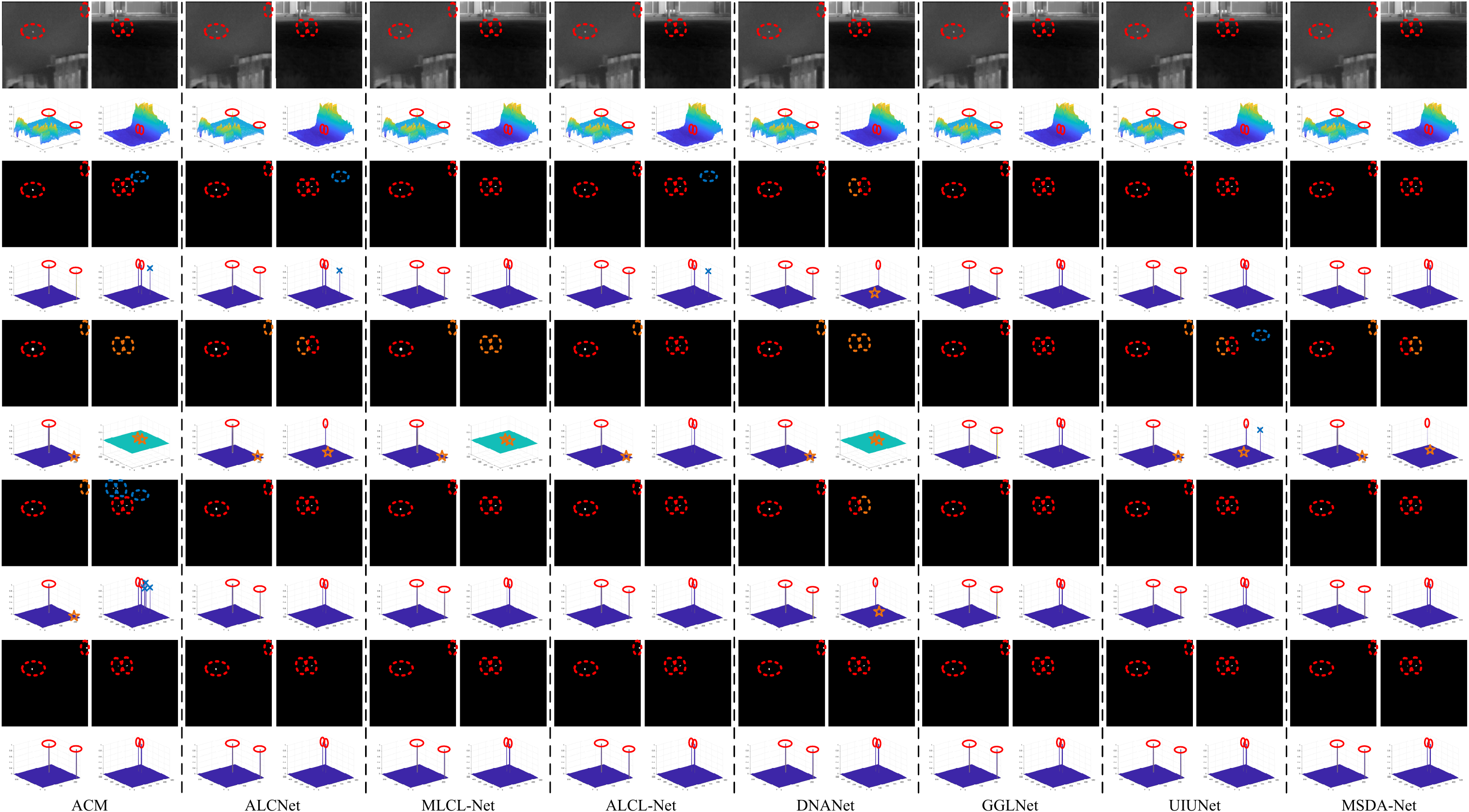}
    \caption{Visualization of several excellent methods on the SIRST3 dataset with coarse point labels. \textcolor{red}{\textit{Red}} denotes the correct detections, \textcolor{blue}{\textit{blue}} denotes the false detections, and \textcolor{yellow}{\textit{yellow}} denotes the missed detections. Every two rows from top to bottom: \textit{Image}, \textit{DLN Full}, \textit{DLN Coarse + LESPS}, \textit{DLN Coarse + PAL}, \textit{True label}.}
    \label{fig:s-fig02}
    
\end{figure*}

\begin{figure*}[t]
    \centering
    \includegraphics[width=\textwidth]{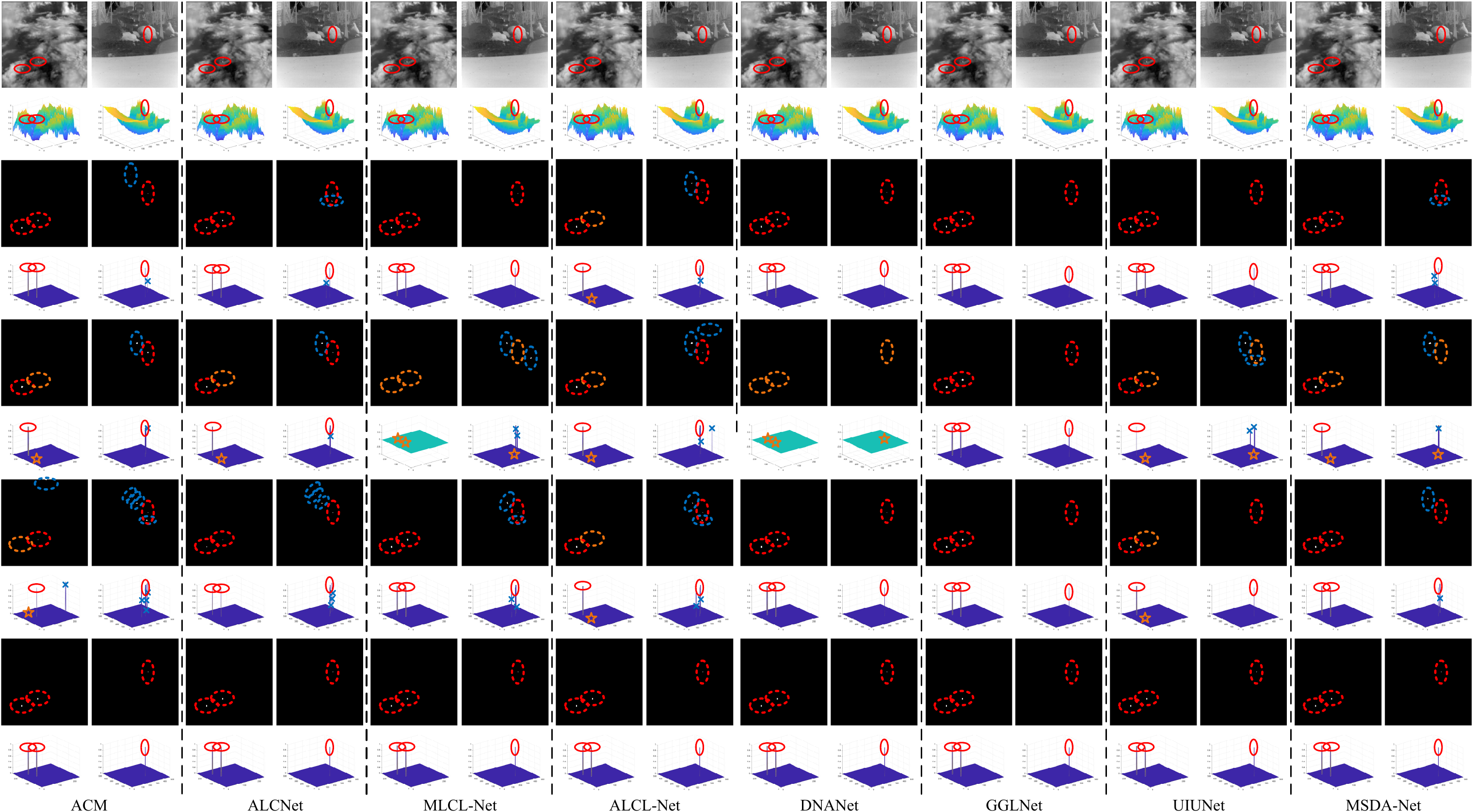}
    \caption{Visualization of several excellent methods on the SIRST3 dataset with centroid point labels. \textcolor{red}{\textit{Red}} denotes the correct detections, \textcolor{blue}{\textit{blue}} denotes the false detections, and \textcolor{yellow}{\textit{yellow}} denotes the missed detections. Every two rows from top to bottom: \textit{Image}, \textit{DLN Full}, \textit{DLN Centroid + LESPS}, \textit{DLN Centroid + PAL}, \textit{True label}.}
    \label{fig:s-fig03}
\end{figure*}

\begin{figure*}[!t]
    \centering
    \includegraphics[width=\textwidth]{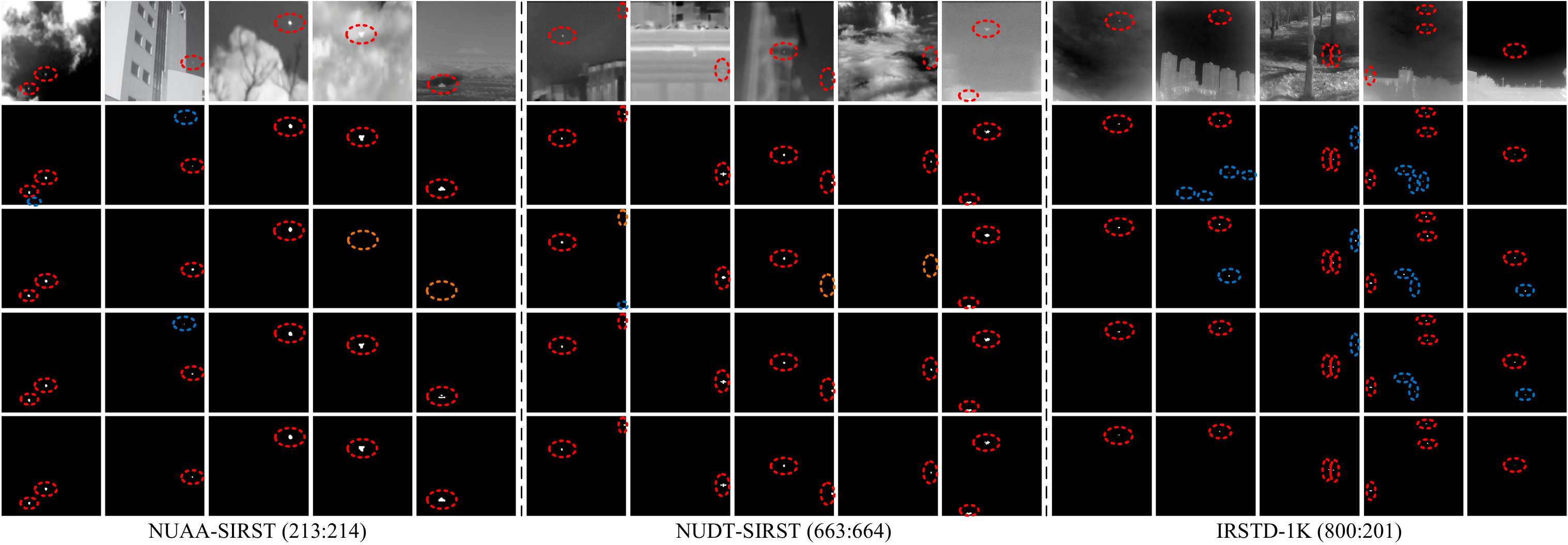}
    \vspace{-15pt}
    \caption{Visualization of MLCL-Net on the SIRST3 dataset with coarse point labels. \textcolor{red}{\textit{Red}}, \textcolor{blue}{\textit{blue}}, and \textcolor{yellow}{\textit{yellow}} denotes correct detections, false detections, and missed detections. From top to bottom: \textit{Image}, \textit{DLN Full}, \textit{DLN Coarse + LESPS}, \textit{DLN Coarse + PAL}, \textit{True label}.}
    \label{fig:s-fig04}
\end{figure*}

\begin{figure*}[!t]
    \centering
    \includegraphics[width=\textwidth]{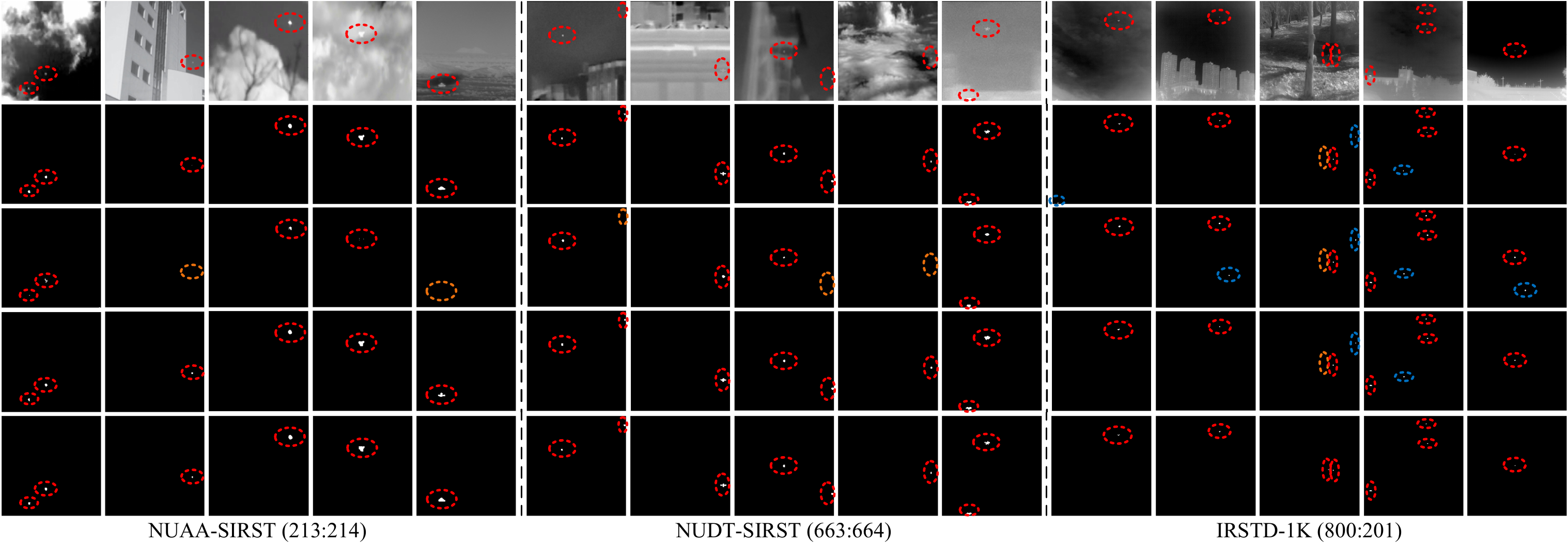}
    \vspace{-15pt}
    \caption{Visualization of ALCL-Net on the SIRST3 dataset with coarse point labels. \textcolor{red}{\textit{Red}}, \textcolor{blue}{\textit{blue}}, and \textcolor{yellow}{\textit{yellow}} denotes correct detections, false detections, and missed detections. From top to bottom: \textit{Image}, \textit{DLN Full}, \textit{DLN Coarse + LESPS}, \textit{DLN Coarse + PAL}, \textit{True label}.}
    \label{fig:s-fig05}
\end{figure*}

\begin{figure*}[!t]
    \centering
    \includegraphics[width=\textwidth]{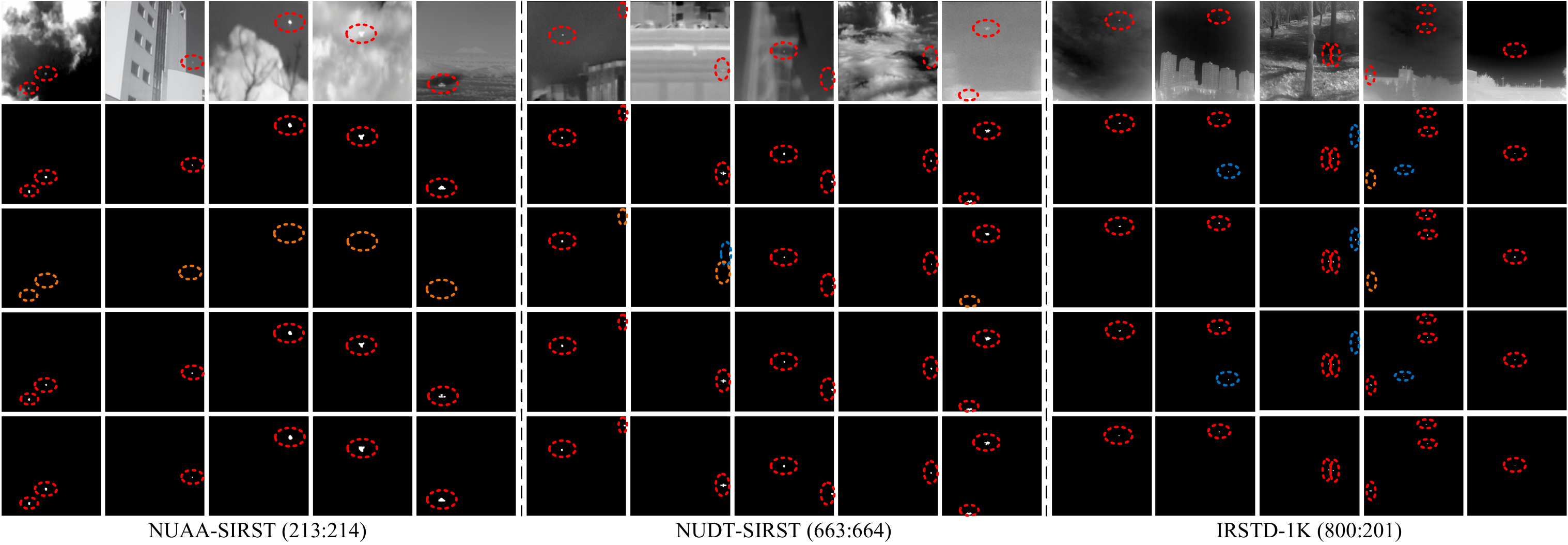}
    \vspace{-15pt}
    \caption{Visualization of DNANet on the SIRST3 dataset with coarse point labels. \textcolor{red}{\textit{Red}}, \textcolor{blue}{\textit{blue}}, and \textcolor{yellow}{\textit{yellow}} denotes correct detections, false detections, and missed detections. From top to bottom: \textit{Image}, \textit{DLN Full}, \textit{DLN Coarse + LESPS}, \textit{DLN Coarse + PAL}, \textit{True label}.}
    \label{fig:s-fig06}
\end{figure*}

\begin{figure*}[!t]
    \centering
    \includegraphics[width=\textwidth]{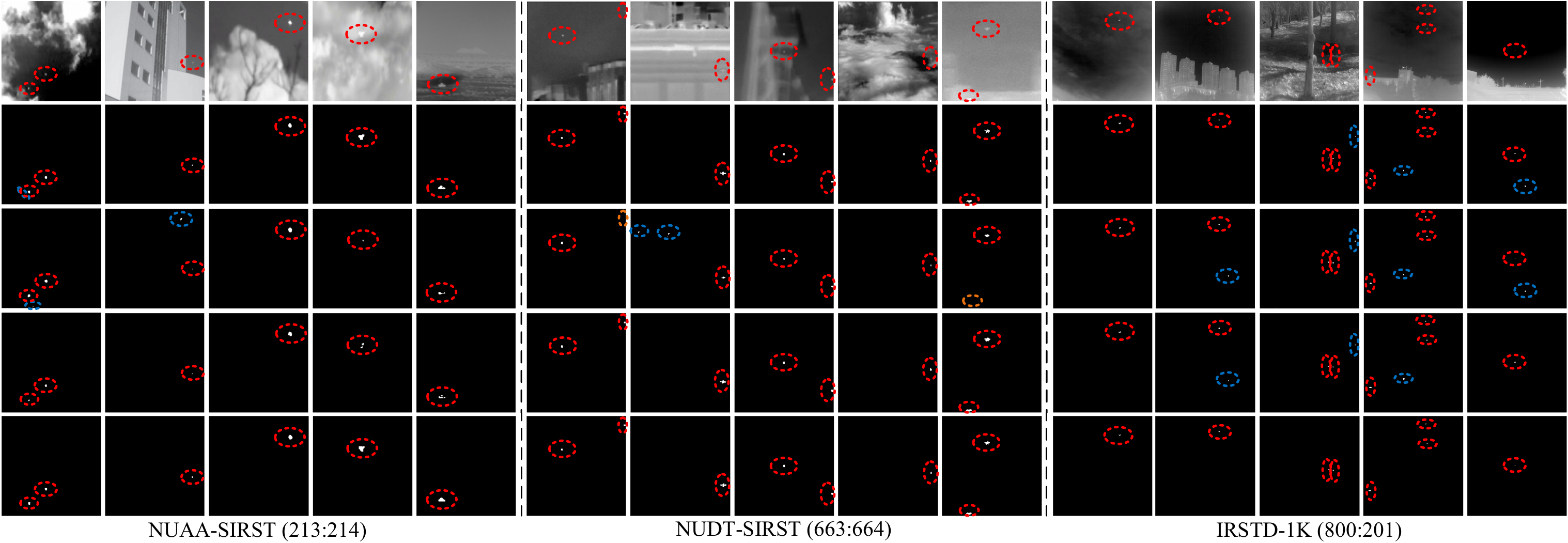}
    \vspace{-15pt}
    \caption{Visualization of GGL-Net on the SIRST3 dataset with coarse point labels. \textcolor{red}{\textit{Red}}, \textcolor{blue}{\textit{blue}}, and \textcolor{yellow}{\textit{yellow}} denotes correct detections, false detections, and missed detections. From top to bottom: \textit{Image}, \textit{DLN Full}, \textit{DLN Coarse + LESPS}, \textit{DLN Coarse + PAL}, \textit{True label}.}
    \label{fig:s-fig07}
\end{figure*}

\begin{figure*}[!t]
    \centering
    \includegraphics[width=\textwidth]{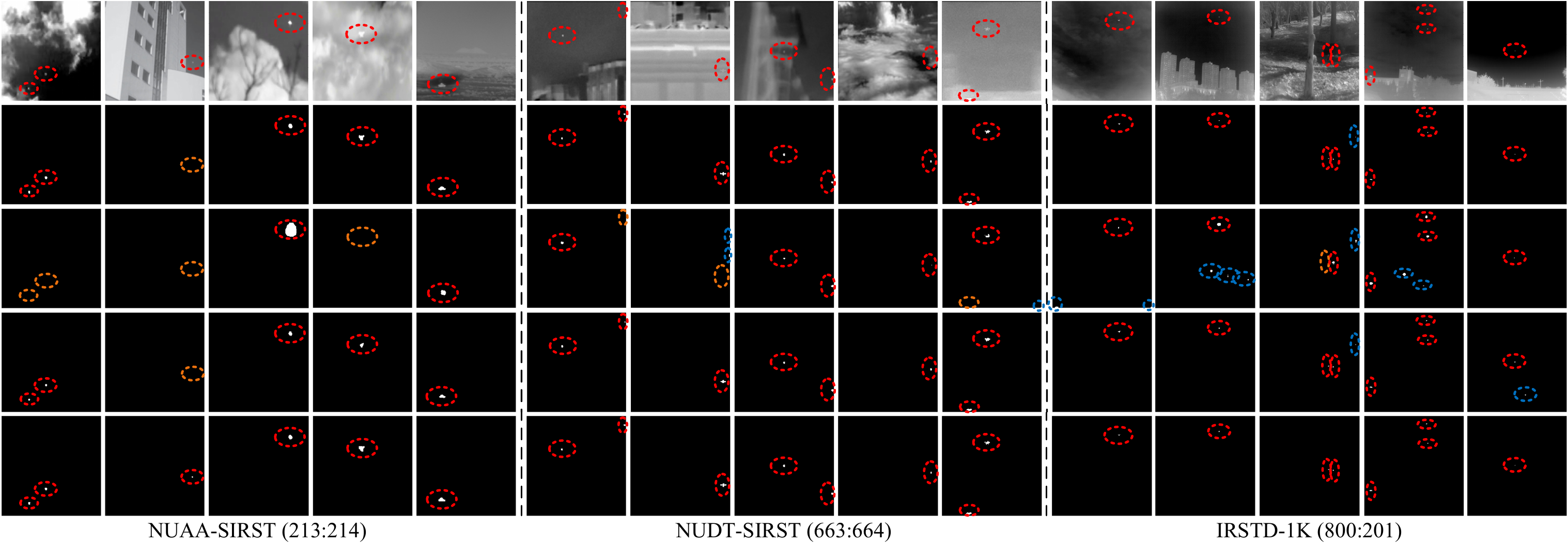}
    \vspace{-15pt}
    \caption{Visualization of UIUNet on the SIRST3 dataset with coarse point labels. \textcolor{red}{\textit{Red}}, \textcolor{blue}{\textit{blue}}, and \textcolor{yellow}{\textit{yellow}} denotes correct detections, false detections, and missed detections. From top to bottom: \textit{Image}, \textit{DLN Full}, \textit{DLN Coarse + LESPS}, \textit{DLN Coarse + PAL}, \textit{True label}.}
    \label{fig:s-fig08}
\end{figure*}

\begin{figure*}[!t]
    \centering
    \includegraphics[width=\textwidth]{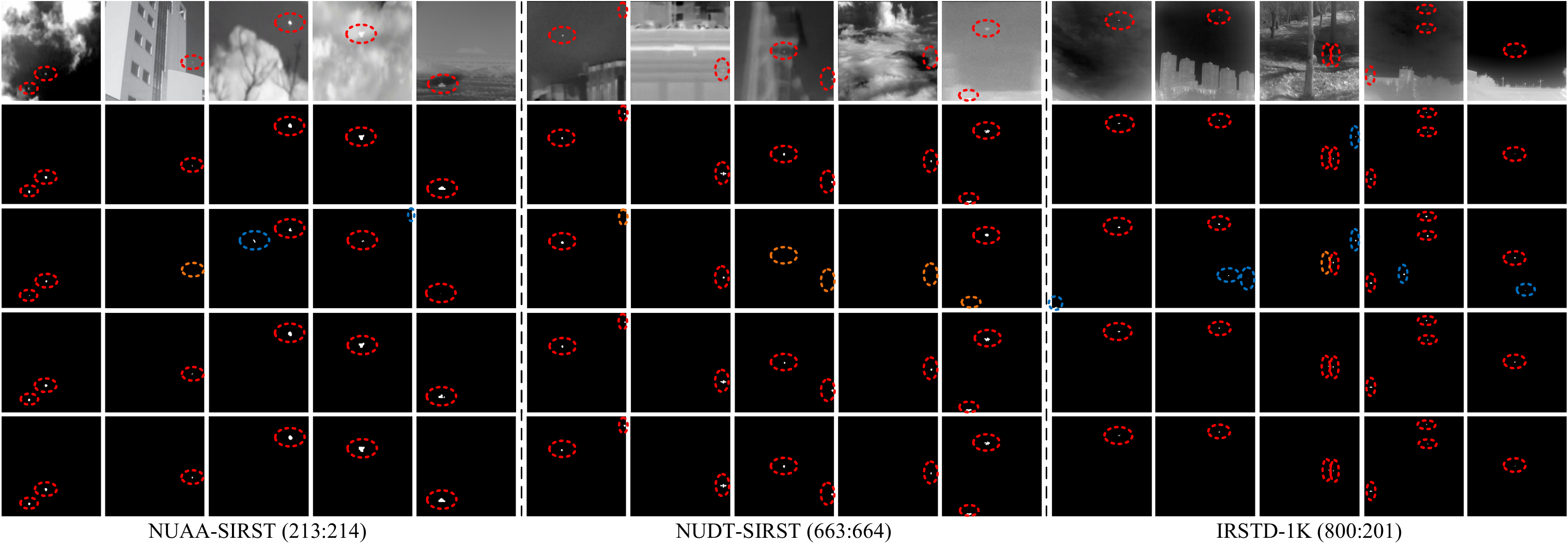}
    \vspace{-15pt}
    \caption{Visualization of MSDA-Net on the SIRST3 dataset with coarse point labels. \textcolor{red}{\textit{Red}}, \textcolor{blue}{\textit{blue}}, and \textcolor{yellow}{\textit{yellow}} denotes correct detections, false detections, and missed detections. From top to bottom: \textit{Image}, \textit{DLN Full}, \textit{DLN Coarse + LESPS}, \textit{DLN Coarse + PAL}, \textit{True label}.}
    \label{fig:s-fig09}
\end{figure*}

\begin{figure*}[!t]
    \centering
    \includegraphics[width=\textwidth]{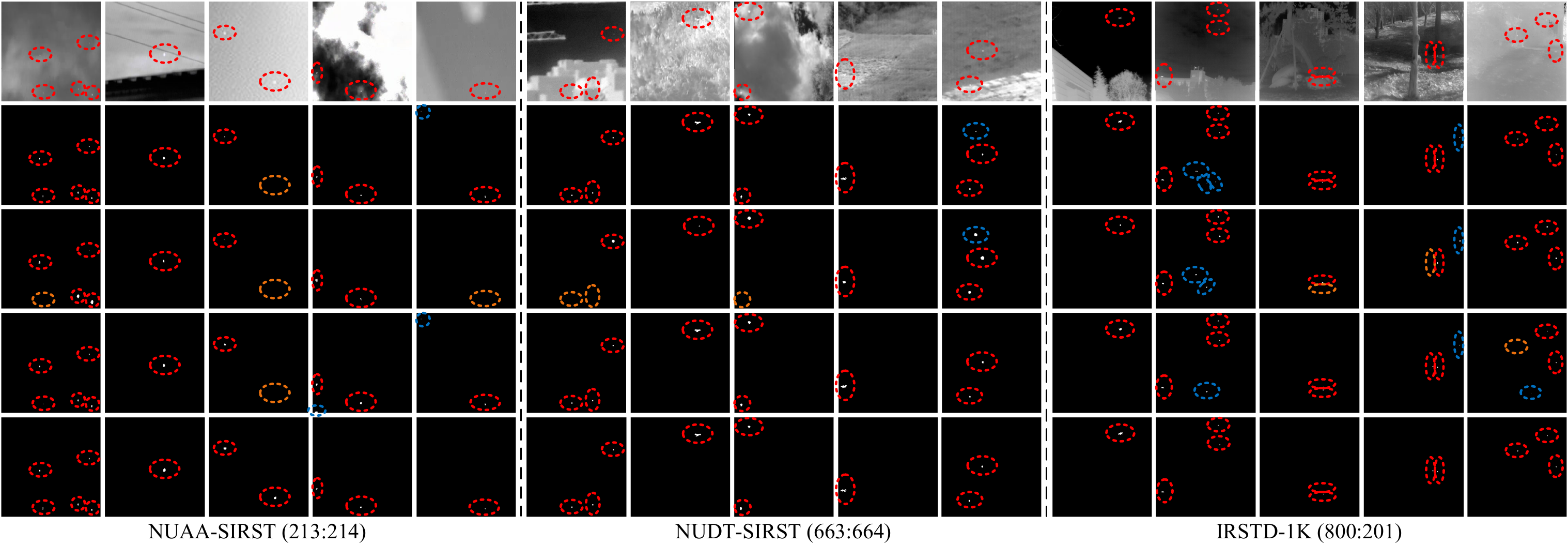}
    \vspace{-15pt}
    \caption{Visualization of MLCL-Net on the SIRST3 dataset with centroid point labels. \textcolor{red}{\textit{Red}}, \textcolor{blue}{\textit{blue}}, and \textcolor{yellow}{\textit{yellow}} denotes correct detections, false detections, and missed detections. From top to bottom: \textit{Image}, \textit{DLN Full}, \textit{DLN Centroid + LESPS}, \textit{DLN Centroid + PAL}, \textit{True label}.}
    \label{fig:s-fig10}
\end{figure*}

\begin{figure*}[!t]
    \centering
    \includegraphics[width=\textwidth]{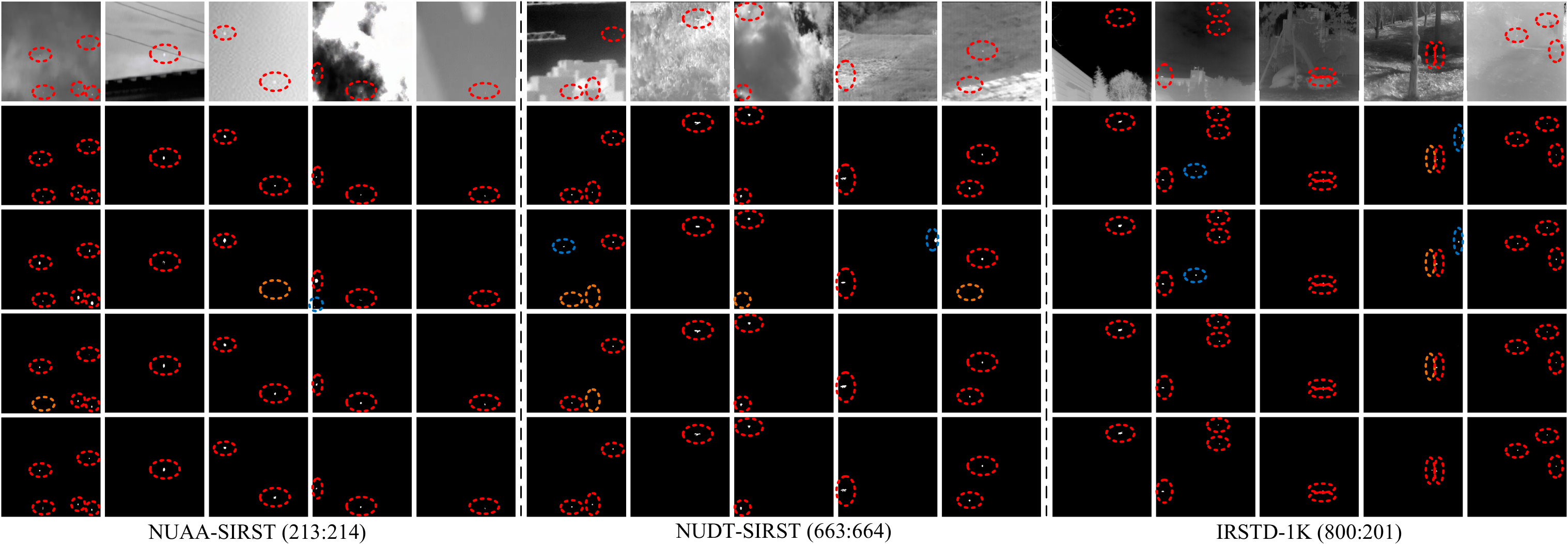}
    \vspace{-15pt}
    \caption{Visualization of ALCL-Net on the SIRST3 dataset with centroid point labels. \textcolor{red}{\textit{Red}}, \textcolor{blue}{\textit{blue}}, and \textcolor{yellow}{\textit{yellow}} denotes correct detections, false detections, and missed detections. From top to bottom: \textit{Image}, \textit{DLN Full}, \textit{DLN Centroid + LESPS}, \textit{DLN Centroid + PAL}, \textit{True label}.}
    \label{fig:s-fig11}
\end{figure*}

\begin{figure*}[!t]
    \centering
    \includegraphics[width=\textwidth]{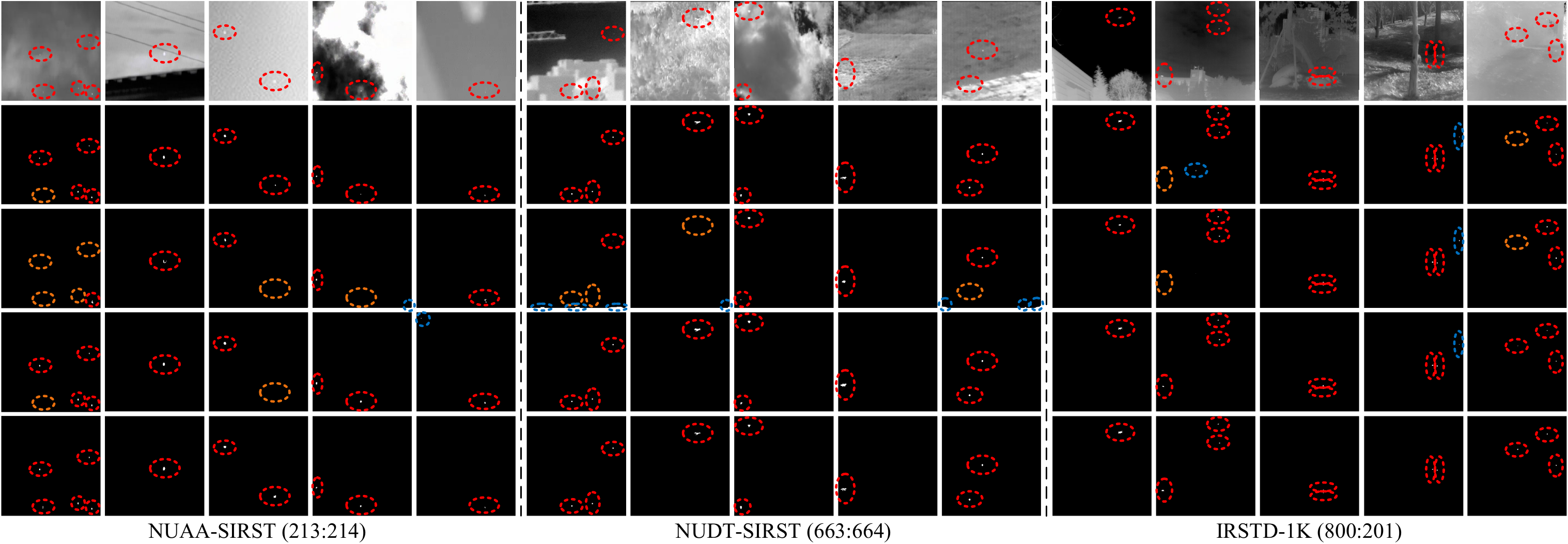}
    \vspace{-15pt}
    \caption{Visualization of DNANet on the SIRST3 dataset with centroid point labels. \textcolor{red}{\textit{Red}}, \textcolor{blue}{\textit{blue}}, and \textcolor{yellow}{\textit{yellow}} denotes correct detections, false detections, and missed detections. From top to bottom: \textit{Image}, \textit{DLN Full}, \textit{DLN Centroid + LESPS}, \textit{DLN Centroid + PAL}, \textit{True label}.}
    \label{fig:s-fig12}
\end{figure*}

\begin{figure*}[!t]
    \centering
    \includegraphics[width=\textwidth]{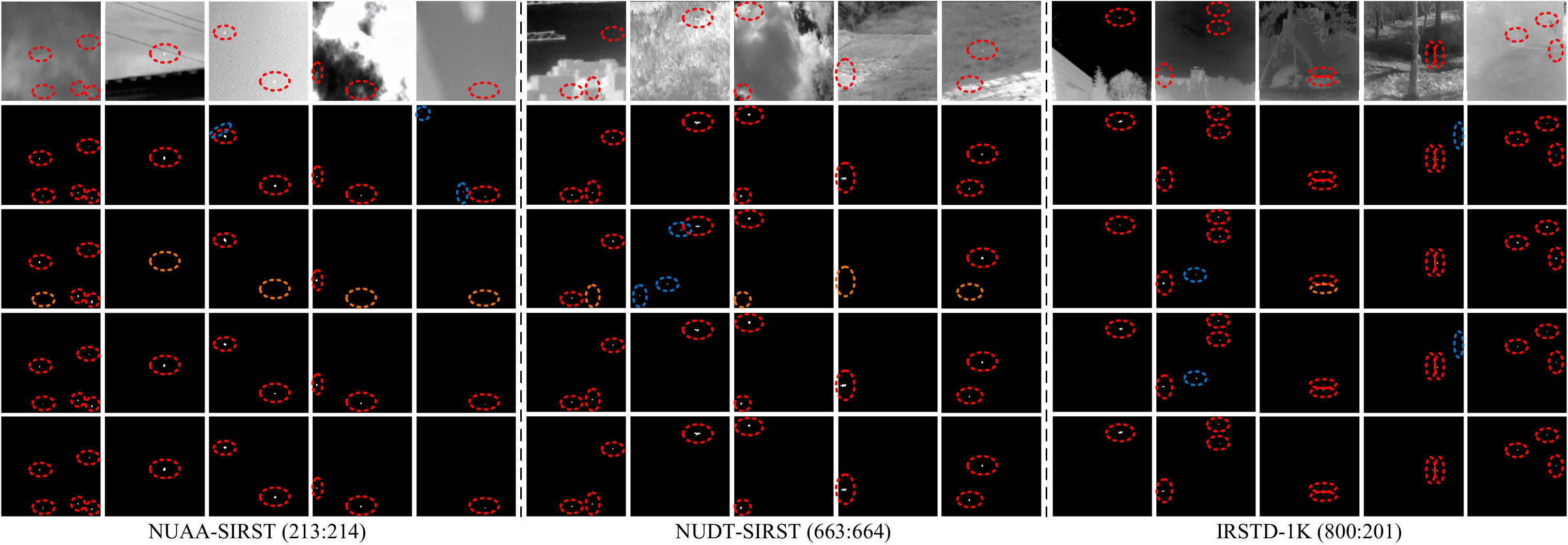}
    \vspace{-15pt}
    \caption{Visualization of GGL-Net on the SIRST3 dataset with centroid point labels. \textcolor{red}{\textit{Red}}, \textcolor{blue}{\textit{blue}}, and \textcolor{yellow}{\textit{yellow}} denotes correct detections, false detections, and missed detections. From top to bottom: \textit{Image}, \textit{DLN Full}, \textit{DLN Centroid + LESPS}, \textit{DLN Centroid + PAL}, \textit{True label}.}
    \label{fig:s-fig13}
\end{figure*}

\begin{figure*}[!t]
    \centering
    \includegraphics[width=\textwidth]{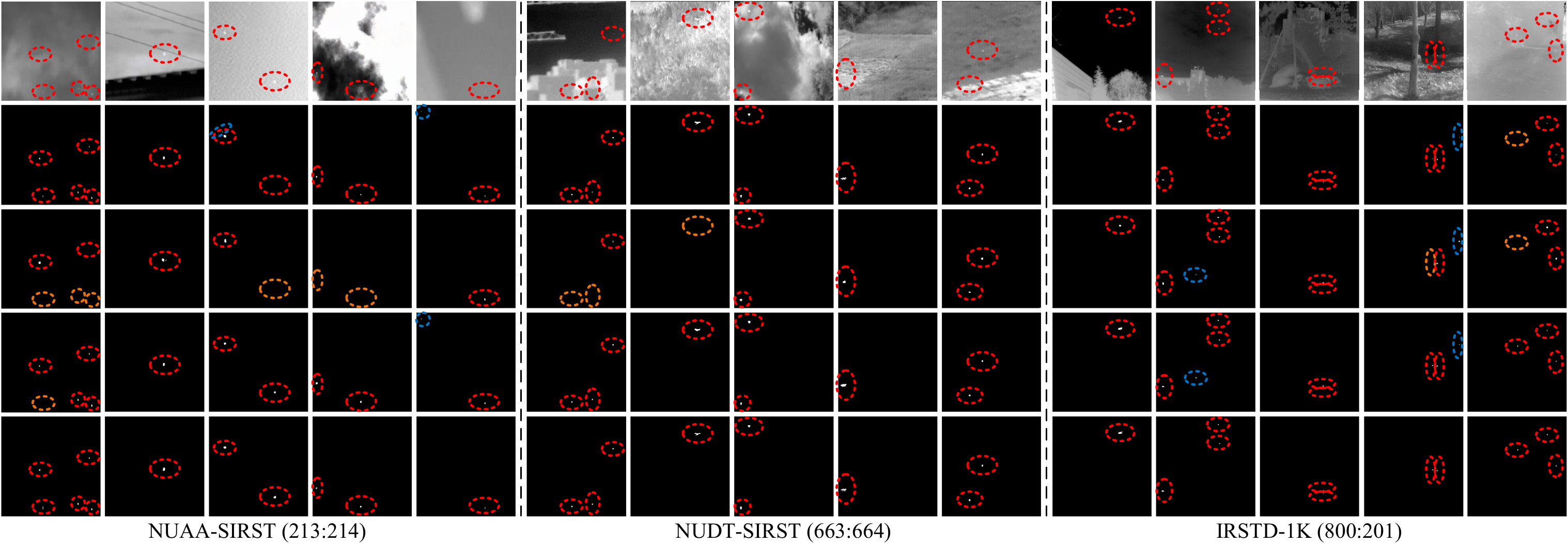}
    \vspace{-15pt}
    \caption{Visualization of UIUNet on the SIRST3 dataset with centroid point labels. \textcolor{red}{\textit{Red}}, \textcolor{blue}{\textit{blue}}, and \textcolor{yellow}{\textit{yellow}} denotes correct detections, false detections, and missed detections. From top to bottom: \textit{Image}, \textit{DLN Full}, \textit{DLN Centroid + LESPS}, \textit{DLN Centroid + PAL}, \textit{True label}.}
    \label{fig:s-fig14}
\end{figure*}

\begin{figure*}[!t]
    \centering
    \includegraphics[width=\textwidth]{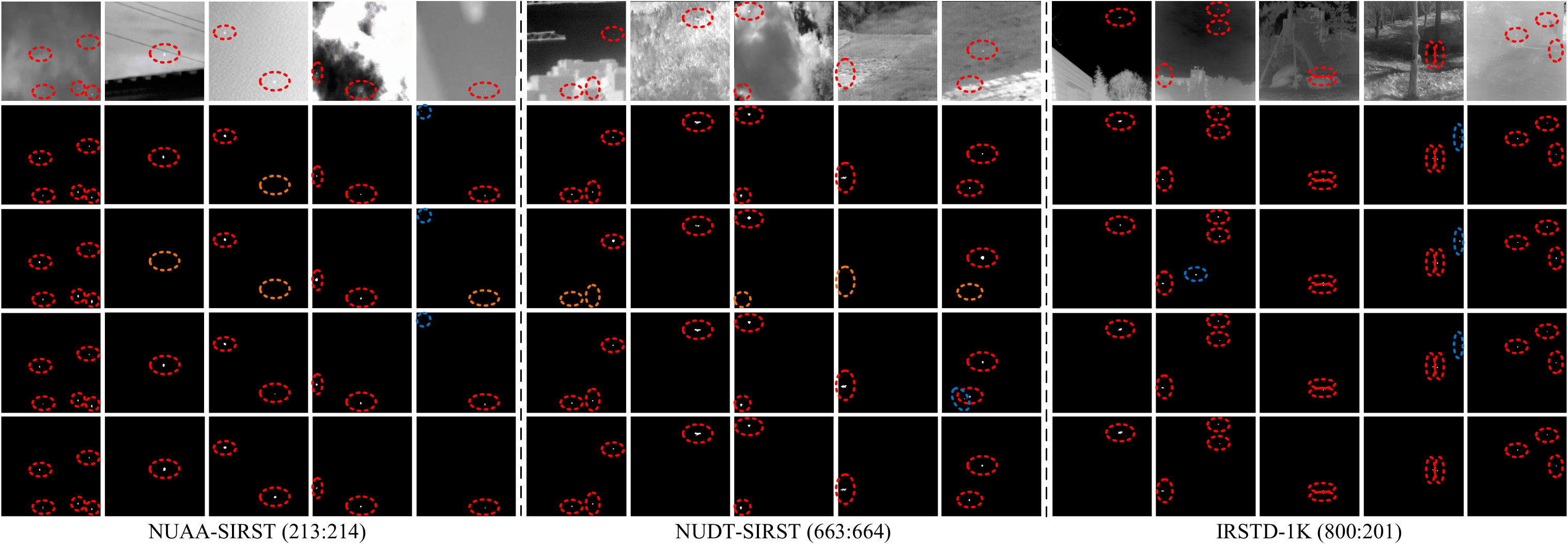}
    \vspace{-15pt}
    \caption{Visualization of MSDA-Net on the SIRST3 dataset with centroid point labels. \textcolor{red}{\textit{Red}}, \textcolor{blue}{\textit{blue}}, and \textcolor{yellow}{\textit{yellow}} denotes correct detections, false detections, and missed detections. From top to bottom: \textit{Image}, \textit{DLN Full}, \textit{DLN Centroid + LESPS}, \textit{DLN Centroid + PAL}, \textit{True label}.}
    \label{fig:s-fig15}
\end{figure*}

\textbf{\textit{4) Missed detection rate threshold.}} During this research, we discover an interesting phenomenon: there are few false detections in the detection results of single-frame infrared small target (SIRST) based on DLNs. This can also be found from the order of magnitude ($1e^{\text{\scriptsize -6}}$) of the $F_a$. At the same time, the falsely detected areas will be eliminated in the coarse outer updates. Therefore, we focus on exploring the threshold of the missed detection rate in the model enhancement phase. To further explore the impact of the missed detection rate threshold setting in the coarse outer updates on the performance of the final generated model, we explore the PAL framework with different missed detection rate threshold settings. The experimental results are shown in \cref{tab:s-tab04}. Compared with using a fixed missed detection rate threshold, using a variable value has relatively better detection results. At the same time, when the missed detection rate threshold is set larger and larger, the final performance gradually decrease. A larger threshold setting means that more harder samples will enter the training pool for training in the early of the model enhancement phase. Combined with the final results, it shows that hard samples should be input reasonably and gradually from simple to difficult in the model enhancement phase. This further verifies the effectiveness of our proposed progressive active learning idea. For the missed detection rate threshold, we set its initial threshold to 0.2 and gradually increase it to 1 as the number of epochs increases in the experiment.

\section{More Quantitative Results}
\label{sec:more quantitative results}
Considering that the main papers only conduct experiments on various datasets with coarse point labels, to further verify the effectiveness of our PAL framework, we conduct additional experiments on the SIRST3, NUAA-SIRST, NUDT-SIRST and IRSTD-1K datasets with centroid point labels in this section.

\textbf{\textit{Evaluation on the SIRST3 dataset with centroid point labels.}} As shown in \cref{tab:s-tab05}, consistent with the use of coarse point labels, when the networks (UIUNet, MSDA-Net) with obvious performance advantages under full supervision are embedded into the LESPS framework for single point supervision tasks, the potential performance advantages of these networks cannot be effectively exploited. However, the performance change trend of each SIRST detection network equipped with our PAL framework under single-point supervision is basically consistent with that of the network under full supervision. Our PAL framework can build an efficient and stable bridge between full supervision and single point supervision tasks. Meanwhile, compared with the LESPS framework, using our PAL framework improves the IoU by 8.53\%-29.10\%, the nIoU by 9.33\%-32.30\%, and the $P_d$ by 1.07\%-6.71\% on the comprehensive SIRST3-Test. The improvement is very obvious. Compared with the fully supervised task on the SIRST3-Test, our PAL framework can reach 79.33\%-84.63\% on IoU, 82.08\%-87.05\% on nIoU, and has comparable performance on $P_d$. In addition, by observing the results of the three decomposed test subsets, our PAL framework has a stable performance that is better than that of the LESPS framework and is in line with the full supervision performance trend. These verify that the proposed PAL framework has excellent generalizability and robustness for SIRST detection tasks in multiple scenes and multiple target types.

\textbf{\textit{Evaluation on three individual datasets with centroid point labels.}} To further explore the stability of the PAL framework when there are few training samples and centroid point labels are used, we conduct separate experiments on the NUAA-SIRST, NUDT-SIRST, and IRSTD-1K datasets. From \cref{tab:s-tab06}, consistent with the results using coarse point labels, the performance of DLNs equipped with the PAL framework is significantly better than that with the LESPS framework. At the same time, some DLNs equipped with the LESPS framework will experience the phenomenon of ``model invalidity'' where the final generated model does not meet the $F_a$ requirements. Specifically, compared with the LESPS framework, using the PAL framework improves the IoU by 9.39\%-49.27\%, the nIoU by 7.67\%-47.18\%, and the $P_d$ by 1.01\%-41.45\%. The improvement is very significant. In addition, except for some minor differences, the performance change trend of each SIRST detection network equipped with our PAL framework under single point supervision is basically consistent with that of the network under full supervision. These results fully verify that our PAL framework still has excellent robustness in the single point supervised SIRST detection task with a small number of training samples.

\section{More Qualitative Results}
\label{sec:more qualitative results}
To further qualitatively compare and analyze the performance of the proposed PAL framework, in this section, we present detailed visualizations of the detection results of multiple methods on multiple datasets using either coarse point labels or centroid point labels.

\textbf{\textit{Visualization on the SIRST3 dataset.}} As shown in \cref{fig:s-fig02} and \cref{fig:s-fig03}, we can find that whether using coarse point labels or centroid point labels, DLNs equipped with our PAL framework are significantly better than those equipped with the LESPS framework. From the 3D results, the LESPS framework easily leads to a large number of missed detections in difficult scenarios, whereas the PAL framework can solve this problem. In addition, for some images, the target-level detection effect of DLNs equipped with the PAL framework under single point supervision is even better than that under full supervision. From the 2D results, DLNs equipped with the PAL framework are significantly better than the LESPS framework in pixel-level segmentation. These results fully demonstrate the effectiveness of our proposed PAL framework for the SIRST detection task with single-point supervision.

\textbf{\textit{Visualization on three individual datasets.}} As shown in \cref{fig:s-fig04,fig:s-fig05,fig:s-fig06,fig:s-fig07,fig:s-fig08,fig:s-fig09,fig:s-fig10,fig:s-fig11,fig:s-fig12,fig:s-fig13,fig:s-fig14,fig:s-fig15}, we provide a detailed visualization of various methods, different point labels and different training frameworks. First, when the LESPS framework is used, there is a significant decrease in detection performance in some networks, such as \cref{fig:s-fig08}, \cref{fig:s-fig09}, \cref{fig:s-fig14}, and \cref{fig:s-fig15}. This shows that the LESPS framework is prone to unstable performance when facing a dataset with few samples. Secondly, DLNs equipped with the PAL framework generally have more refined segmentation effects than LESPS. Finally, compared with the detection results with full supervision, DLNs equipped with the PAL can achieve similar results with single point supervision. These results fully demonstrate the robustness of our proposed PAL framework on the single point supervised SIRST detection task with a small number of samples.

\end{document}